\PassOptionsToPackage{table}{xcolor}
\documentclass[]{paper}

\RequirePackage{tocloft}

\usepackage[utf8]{inputenc}
\usepackage[T1]{fontenc}
\usepackage{hyperref}
\usepackage{url}
\usepackage{booktabs}
\usepackage{amsfonts}
\usepackage{nicefrac}
\usepackage{microtype}
\usepackage{caption}
\usepackage{multirow}
\usepackage{makecell}
\usepackage{siunitx}

\usepackage{graphicx}
\usepackage{amsmath}

\usepackage{subcaption}
\usepackage{wrapfig}
\usepackage{listings}
\usepackage{csquotes}

\usepackage{tikz}
\usepackage[edges]{forest}

\title{A Survey of Vibe Coding with Large Language Models}

\author{\textbf{Yuyao Ge}$^{1*}$\quad \textbf{Lingrui Mei}$^{1*}$\quad \textbf{Zenghao Duan}$^{1}$\quad  \textbf{Tianhao Li}$^{2}$ \quad
\textbf{Yujia Zheng}$^{2}$ \\[0.7em] \textbf{Yiwei Wang}$^{3}$\quad \textbf{Lexin Wang}$^{1}$\quad \textbf{Jiayu Yao}$^{1*}$\quad  \textbf{Tianyu Liu}$^{4}$\quad \textbf{Yujun Cai}$^{5}$ \\[0.7em] 
 \textbf{Baolong Bi}$^{1*}$\quad \textbf{Fangda Guo}$^{1}$\quad \textbf{Jiafeng Guo}$^{1*}$\quad \textbf{Shenghua Liu}$^{1\dagger *}$ \quad \textbf{Xueqi Cheng}$^{1*}$ \\[1.5em] 
$^1$Institute of Computing Technology, Chinese Academy of Sciences\quad $^2$Duke University\\[0.5em] 
$^3$University of California, Merced\quad $^4$Peking University\quad $^5$University of Queensland}

\abstract{
  The advancement of large language models (LLMs) has catalyzed a paradigm shift from code generation assistance to autonomous coding agents, enabling a novel development methodology termed "Vibe Coding" where developers validate AI-generated implementations through outcome observation rather than line-by-line code comprehension.
  Despite its transformative potential, the effectiveness of this emergent paradigm remains under-explored, with empirical evidence revealing unexpected productivity losses and fundamental challenges in human-AI collaboration.
  To address this gap, this survey provides the first comprehensive and systematic review of Vibe Coding with large language models, establishing both theoretical foundations and practical frameworks for this transformative development approach.
  Drawing from systematic analysis of over 1000 research papers, we survey the entire vibe coding ecosystem, examining critical infrastructure components including LLMs for coding, LLM-based coding agent, development environment of coding agent, and feedback mechanisms.
  We first introduce Vibe Coding as a formal discipline by formalizing it through a Constrained Markov Decision Process that captures the dynamic triadic relationship among human developers, software projects, and coding agents.
  Building upon this theoretical foundation, we then synthesize existing practices into five distinct development models: Unconstrained Automation, Iterative Conversational Collaboration, Planning-Driven, Test-Driven, and Context-Enhanced Models, thus providing the first comprehensive taxonomy in this domain.
  Critically, our analysis reveals that successful Vibe Coding depends not merely on agent capabilities but on systematic context engineering, well-established development environments, and human-agent collaborative development models.
  Based on these findings, we identify key challenges spanning technical infrastructure optimization, security mechanisms, and human-centered design considerations.
  Ultimately, this survey serves as both a conceptual foundation for AI-augmented software engineering and a technical roadmap for researchers and practitioners navigating this rapidly evolving field.
}

\lab{Also affiliated with: (1) Key Laboratory of Network Data Science and Technology, ICT, CAS; (2) State Key Laboratory of AI Safety; (3) University of Chinese Academy of Sciences}
\correspondence{Corresponding Author}


\date{\today}

\Contact{\href{mailto:geyuyao24z@ict.ac.cn}{geyuyao24z@ict.ac.cn}, \href{mailto:liushenghua@ict.ac.cn}{liushenghua@ict.ac.cn}}

\repo{\href{https://github.com/YuyaoGe/Awesome-Vibe-Coding}{https://github.com/YuyaoGe/Awesome-Vibe-Coding}}

\begin{document}

\maketitle

\newpage
\tableofcontents
\newpage

\section{Introduction}

Large language models (LLMs) have significantly advanced artificial intelligence through conversational systems capable of fluent natural language understanding and generation \citep{brown2020language, ouyang2022training}. Early adoption in software development positioned LLMs as supplementary assistants—developers employed natural language prompts to generate code snippets, but significant accuracy limitations necessitated manual review and iterative debugging throughout the software development lifecycle \citep{zhang2023repocoder, nijkamp2023codegen, luo2023wizardcoder, li2023starcoder, roziere2023code}.

The emergence of advanced architectures like GPT-4 \citep{achiam2023gpt} and Claude Sonnet 4 \citep{anthropic2025claude45} enabled qualitative improvements, leading to \textbf{Coding Agents} capable of autonomously completing programming tasks through dynamic environmental interaction via shell commands, file operations, and test execution \citep{qian2023chatdev}. 
These agents have demonstrated rapid progress on real-world programming tasks. 
Taking SWE-bench as an example \citep{jimenez2023swe}, SWE-agent reached 12.5\% with custom interfaces \citep{yang2024swe}, AutoCodeRover achieved 19.0\% resolution through code search and fault localization \citep{zhang2024autocoderover}, Agentless attained 27.3\% \citep{xia2024agentless}, OpenHands achieved 53\% on SWE-bench Verified \citep{wang2024openhands}, and self-improving agents demonstrated 17--53\% performance gains on SWE-bench Verified \citep{robeyns2025self}.

With the advancement of large language models such as GPT-5 Pro \citep{openai2025gpt5} and Claude Sonnet 4.5 \citep{anthropic2025claude45}, llm-based coding agent capabilities have achieved significant breakthroughs, giving rise to ``\textbf{Vibe Coding}''—a paradigm where developers rely on AI-generated code without line-by-line inspection, engaging instead in iterative cycles of natural language requirement articulation, execution observation, and feedback \citep{karpathy2025, ray2025review, sapkota2025vibe, horvat2025vibe}.
Coding Agents go beyond code generation—they autonomously configure environments, execute programs, self-diagnose errors, and update implementations. This represents a substantial elevation in human trust and a departure from traditional comprehension mandates toward outcome-oriented validation \citep{novikov2025alphaevolve, ho2025verilogcoder, zhao2025mage, zeng2025glm}.

However, possessing powerful agents proves insufficient. Task complexity exposes fundamental limitations in unstructured natural language instructions, which fail to convey nuanced requirements and architectural constraints \citep{jimenez2023swe, yang2023intercode}. Empirical evidence reveals that experienced developers using Cursor with Claude experienced 19\% increased completion time rather than anticipated productivity gains \citep{becker2025measuring}. Effective human-AI collaboration demands systematic prompt engineering and context engineering \citep{mei2025survey}, structured instructions \citep{nashid2023retrieval, zhang2023repocoder, zhou2023large, white2023prompt}, and balanced agency distribution across various distinct interaction types \citep{treude2025developers, fragiadakis2024evaluating}.

To address this critical gap, this survey provides the first comprehensive and systematic review of Vibe Coding with Large Language Models.
As shown in Figure~\ref{fig:vibe_coding}, \textbf{we introduce Vibe Coding as a dynamic triadic relationship among human developers, software projects, and coding agents, providing its first formal definition as an engineering discipline through a Constrained Markov Decision Process~\citep{altman2021constrained}. 
Building upon this theoretical foundation, we distill Vibe Coding workflows into five development models—Unconstrained Automation Model, Iterative Conversational Collaboration Model, Planning-Driven Model, Test-Driven Model, and Context-Enhanced Model—representing the first comprehensive synthesis of existing practices.}
Through this framework, we: (1) establish rigorous theoretical foundations for understanding human-agent collaboration in software development; (2) provide developers with actionable guidance for selecting and implementing appropriate development strategies; and (3) identify critical challenges and future directions spanning technical infrastructure, security mechanisms, and human factors. 
This work serves as both a conceptual foundation for the emerging field of AI-augmented software engineering and a technical roadmap for advancing research and practice in coding agent systems.

The remainder of this paper is organized as follows. Section~\ref{sec:related} reviews related surveys and foundational technologies relevant to Vibe Coding. Section~\ref{sec:vibe_coding} formally defines Vibe Coding and establishes its theoretical foundations through a Constrained Markov Decision Process formalization. Section~\ref{sec:llm_coding} surveys large language models for coding, covering data foundations, pre-training techniques, and post-training methods. Section~\ref{sec:coding_agent} examines LLM-based coding agents, analyzing planning capabilities, memory mechanisms, action execution, and collaborative architectures. Section~\ref{sec:dev_environment} explores critical infrastructure components including isolated execution environments, interactive development interfaces, and distributed orchestration platforms. Section~\ref{sec:feedback} investigates feedback mechanisms spanning compiler feedback, execution feedback, human feedback, and self-refinement. Section~\ref{sec:dev_models} presents our proposed taxonomy of five development models with analysis of their characteristics and applications. Finally, Section~\ref{sec:future} discusses future impacts and open challenges encompassing technical advancement, security considerations, and human-centered design.

\definecolor{softblue}{RGB}{220,230,242}    
\definecolor{softgreen}{RGB}{226,239,218}   
\definecolor{softpurple}{RGB}{229,224,236}  
\definecolor{softyellow}{RGB}{255,242,204}  
\definecolor{softred}{RGB}{242,220,219}     
\definecolor{softgray}{RGB}{240,240,240}     
\definecolor{softgold}{RGB}{235,190,115}     

\tikzstyle{leaf}=[draw=black,
    rounded corners,minimum height=1em,
    text width=26em,
    text opacity=1, 
    align=left,
    fill opacity=.3,  text=black,font=\scriptsize,
    inner xsep=5pt, inner ysep=3pt,
    ]
\tikzstyle{leaf1}=[draw=black,
    rounded corners,minimum height=1em,
    text width=7.0em,
    text opacity=1, align=center,
    fill opacity=.5,  text=black, font=\scriptsize,
    inner xsep=3pt, inner ysep=3pt,
    ]
\tikzstyle{leaf2}=[draw=black,
    rounded corners,minimum height=1em,
    text width=10.5em,
    text opacity=1, align=center,
    fill opacity=.8,  text=black,font=\scriptsize,
    inner xsep=3pt, inner ysep=3pt,
    ]
\tikzstyle{leaf3}=[draw=black,
    rounded corners,minimum height=1em,
    text width=7.0em,
    text opacity=1, align=center,
    fill opacity=1.0,  text=black,font=\scriptsize,
    inner xsep=3pt, inner ysep=3pt,
]

\begin{figure*}[!t]
\centering
\begin{forest}
  for tree={
  forked edges,
  calign=center,  
  grow=east,
  reversed=true,
  anchor=center,
  parent anchor=east,  
  child anchor=west,   
  base=center,
  font=\small,
  rectangle,
  draw=black,
  edge=black!50, 
  rounded corners,
  minimum width=3em,
  minimum height=2.5em,
  s sep=2pt,
  inner xsep=5pt,
  inner ysep=2pt,
  }
  [Vibe Coding,rotate=90, anchor=north, inner xsep=4pt,inner ysep=2pt,edge=black!50,draw=black,minimum height=2em,
    [Large Language Models for Coding (\S~\ref{sec:llm_coding}),rotate=90, parent anchor=south, child anchor=north, inner xsep=8pt,inner ysep=3pt, edge=black!50, leaf3, fill=softblue, 
      [Data Foundation of Code LLMs (\S~\ref{sec:llm_coding_data}), leaf2, fill=softblue,
        [Codex~\cite{chen2021evaluating}{,}
           MathPile~\cite{wang2023mathpile}{,}
           Arctic-SnowCoder~\cite{wei2024arctic}{,}
           OpenCoder~\cite{huang2024opencoder}{,}
           SwallowCode~\cite{fujii2025rewriting}{,}
           WizardCoder~\cite{luo2023wizardcoder}{,}
           OctoPack~\cite{muennighoff2023octopack}{,}
           OpenCodeInstruct~\cite{ahmad2025opencodeinstruct}{,}
           CodeArena~\cite{yang2024evaluating}{,}
           CodeUltraFeedback~\cite{weyssow2024codeultrafeedback}{,}
           Self-Instruct~\cite{wang2022self}{,}
           etc{.}
            ,leaf,fill=softblue]
      ]
      [Pre-training Techniques (\S~\ref{sec:llm_coding_pre}), leaf2, fill=softblue,
        [BERT~\cite{devlin2019bert}{,}
           T5~\cite{raffel2019exploring}{,}
           CodeBPE~\cite{chirkova2023codebpe}{,}
           GraphCodeBERT~\cite{guo2020graphcodebert}{,}
           CodeT5~\cite{wang2021identifier}{,}
           Code Llama~\cite{roziere2023code}{,}
           Birdie~\cite{blouir2024birdie}{,}
           Agent-Q~\cite{jern2025agent}{,}
           GPT-3~\cite{brown2020language}{,}
           CodeBERT~\cite{feng2020codebert}{,}
           UniXcoder~\cite{guo2022unixcoder}{,}
           CoDist~\cite{huang2023program}{,}
           NatGen~\cite{chakraborty2022natgen}{,}
           PLBART~\cite{ahmad2021unified}{,}
           CodeT5+~\cite{wang2023open}{,}
           etc{.}
            ,leaf,fill=softblue]
      ]
      [Post-training Techniques (\S~\ref{sec:llm_coding_post}), leaf2, fill=softblue,
        [VeriCoder~\cite{wei2025vericoder}{,}
           Flan-PaLM~\cite{chung2022scaling}{,}
           FLAN~\cite{wei2021finetuned}{,}
           SparkRA~\cite{wu2024sparkra}{,}
           PanGu-Coder2~\cite{shen2023pangu}{,}
           LoRA~\cite{hu2022lora}{,}
           Adapter Modules~\cite{houlsby2019parameter}{,}
           LIMA~\cite{zhou2023lima}{,}
           AlpaGasus~\cite{chen2023alpagasus}{,}
           FinDPO~\cite{iacovides2025findpo}{,}
           DPO~\cite{rafailov2023direct}{,}
           RAG-Gym~\cite{xiong2025rag}{,}
           RLHF~\cite{stiennon2020learning}{,}
           DUMP~\cite{wang2025dump}{,}
           etc{.}
            ,leaf,fill=softblue]
      ]
    ]
    [LLM-based \\ Coding Agent (\S~\ref{sec:coding_agent}),rotate=90,parent anchor=south, child anchor=north,inner xsep=8pt,inner ysep=3pt, edge=black!50, leaf3, fill=softgreen,
      [Planning and Decomposition Capability (\S~\ref{sec:coding_agent_plan}), leaf2, fill=softgreen,
        [Chain-of-Thought~\cite{wei2022chain}{,}
           AgentGen~\cite{hu2024agentgen}{,}
           ReAct~\cite{yao2022react}{,}
           DPPM~\cite{lu2025decompose}{,}
           HuggingGPT~\cite{shen2023hugginggpt}{,}
           Zero-shot-CoT~\cite{kojima2022large}{,}
           Auto-CoT~\cite{zhang2022automatic}{,}
           Tree of Thoughts~\cite{yao2023tree}{,}
           HyperTree Planning~\cite{gui2025hypertree}{,}
           CodePlan~\cite{wen2024codeplan}{,}
           SQLucid~\cite{tian2024sqlucid}{,}
           HQD~\cite{patel2022question}{,}
           etc{.}
            ,leaf,fill=softgreen]
      ]
      [Memory Mechanism (\S~\ref{sec:coding_agent_mem}), leaf2, fill=softgreen,
        [Transformer~\cite{vaswani2017attention}{,}
           MMS~\cite{zhang2025multiple}{,}
           A-MEM~\cite{xu2025mem}{,}
           MemEngine~\cite{zhang2025memengine}{,}
           TaSL~\cite{feng2024tasl}{,}
           MemoryBank~\cite{zhong2023memorybank}{,}
           Generative Agents~\cite{park2023generative}{,}
           Zero-Shot Planner~\cite{huang2022language}{,}
           RET-LLM~\cite{modarressi2023ret}{,}
           etc{.}
            ,leaf,fill=softgreen]
      ]
      [Action Execution (\S~\ref{sec:coding_agent_act}), leaf2, fill=softgreen,
        [Toolformer~\cite{schick2023toolformer}{,}
           Large Knowledge Model~\cite{chen2023large}{,}
           Alpha-UMi~\cite{shen2024small}{,}
           MCP~\cite{hou2025model}{,}
           MCP-Zero~\cite{fei2025mcp}{,}
           AutoTools~\cite{shi2024tool}{,}
           OpenHands~\cite{wang2024openhands}{,}
           CodeAgent~\cite{zhang2024codeagent}{,}
           RAIT~\cite{srinivas2024retrieval}{,}
           ScaleMCP~\cite{lumer2025scalemcp}{,}
           Doc2Agent~\cite{ni2025doc2agent}{,}
           SE-Agent~\cite{lin2025agent}{,}
           AgentCoder~\cite{huang2023agentcoder}{,}
           etc{.}
            ,leaf,fill=softgreen]
      ]
      [Reflection (\S~\ref{sec:coding_agent_ref}), leaf2, fill=softgreen,
        [Self-Refine~\cite{madaan2023self}{,}
           Saarthi~\cite{kumar2025saarthi}{,}
           Repeton~\cite{vinh2025repeton}{,}
           Self-Planning~\cite{jiang2023self}{,}
           PairCoder~\cite{zhang2024pair}{,}
           Self-Debugging~\cite{chen2023teaching}{,}
           Chained~\cite{ashrafi2025enhancing}{,}
           TiCoder~\cite{fakhoury2024llm}{,}
           ITDCG~\cite{lahiri2022interactive}{,}
           Self-critique~\cite{ho2025self}{,}
           N2M-RSI~\cite{ando2025when}{,}      ReVeal~\cite{jin2025reveal}{,}
           ProCoder~\cite{bi2024iterative}{,}
           RGD~\cite{jin2024rgd}{,}
           etc{.}
            ,leaf,fill=softgreen]
      ]
      [Agent Collaboration (\S~\ref{sec:coding_agent_agent}), leaf2, fill=softgreen,
        [CoMAL~\cite{yao2024comal}{,}
           DRF~\cite{lou2025drf}{,}
           TeamMedAgents~\cite{mishra2025teammedagents}{,}
           AgentMesh~\cite{khanzadeh2025agentmesh}{,}
           Hybrid~\cite{williams2025multi}{,}
           LMA~\cite{he2025llm}{,}
           MASs~\cite{tran2025multi}{,}
           Agent Forest~\cite{li2024more}{,}
           Chatdev~\cite{qian2023chatdev}{,}
           MapCoder~\cite{islam2024mapcoder}{,}
           etc{.}
            ,leaf,fill=softgreen]
      ]
    ]
    [Development Environment (\S~\ref{sec:dev_environment}),rotate=90,parent anchor=south, child anchor=north,inner xsep=8pt,inner ysep=3pt, edge=black!50, leaf3, fill=softyellow,
      [Isolated Execution \\ Runtime (\S~\ref{sec:dev_environment_iso}), leaf2, fill=softyellow,
        [Virtual Earth Cloud~\cite{santoro2023virtual}{,}
           KUNERVA~\cite{lee2023kunerva}{,}
           FunDa~\cite{lounissi2025funda}{,}
           COCOS~\cite{baresi2020cocos}{,}
           SCHEMA lab~\cite{adamidi2025virtual}{,}
           Singularity~\cite{kurtzer2017singularity}{,}
           DRIVE~\cite{zhou2022drive}{,}
           MultiPL-E~\cite{cassano2023multipl}{,}
           TableGPT2~\cite{su2024large}{,}
           SWE-bench~\cite{jimenez2023swe}{,}
           SandboxEval~\cite{rabin2025sandboxeval}{,}
           AutoSafeCoder~\cite{nunez2024autosafecoder}{,}
           Secure SDLC~\cite{gajbhiye2024secure}{,}
           etc{.}
            ,leaf,fill=softyellow]
      ]
      [Interactive Development Interface (\S~\ref{sec:dev_environment_int}), leaf2, fill=softyellow,
        [Designing PairBuddy~\cite{robe2022designing}{,}
           MultiMind~\cite{donato2025multimind}{,}
           Language server protocol~\cite{bork2023language}{,}
           Git Context Controller~\cite{wu2025git}{,}
           StackSpot AI~\cite{pinto2023developer}{,}
           LSP~\cite{gunasinghe2021language}{,}
           Opik~\cite{koc2025mind}{,}
           etc{.}
            ,leaf,fill=softyellow]
      ]
      [Distributed Orchestration Platform (\S~\ref{sec:dev_environment_dis}), leaf2, fill=softyellow,
        [TosKer~\cite{brogi2018tosker}{,}
           TOSCAdata~\cite{dehury2021toscadata}{,}
           TORCH~\cite{tomarchio2021torch}{,}
           AutoGen~\cite{wu2023autogen}{,}
           Autogen~\cite{zhu2023autogen}{,}
           AUTOGEN~\cite{porsdam2023autogen}{,}
           CrewAI~\cite{taulli2025crewai}{,}
           MetaGPT~\cite{zhou2024metagpt}{,}
           Microservices~\cite{singh2023microservices}{,}
           XP~\cite{beck2000extreme}{,}
           FogArm~\cite{bisicchia2023continuous}{,}
           Agentic~\cite{joshi2025architecting}{,}
           LLMOps~\cite{choi2025intelligent}{,}
           PharmaSwarm~\cite{song2025llm}{,}
           EGI~\cite{zeng2025edge}{,}
           etc{.}
            ,leaf,fill=softyellow]
      ]
    ]
    [Feedback \\ Mechanisms (\S~\ref{sec:feedback}),rotate=90,parent anchor=south, child anchor=north,inner xsep=8pt,inner ysep=3pt, edge=black!50, leaf3, fill=softred,
      [Compiler Feedback (\S~\ref{sec:feedback_com}), leaf2, fill=softred,
        [RLCF~\cite{jain2023coarse}{,}
           CYCLE~\cite{ding2024cycle}{,}
           AlphaTrans~\cite{ibrahimzada2024alphatrans}{,}
           RLEF~\cite{gehring2024rlef}{,}
           AceCoder~\cite{zeng2025acecoder}{,}
           VisCoder~\cite{ni2025viscoder}{,}
           InterCode~\cite{yang2023intercode}{,}
           The Art of Repair~\cite{ruiz2025art}{,}
           Rtlfixer~\cite{tsai2023rtlfixer}{,}
           etc{.}
            ,leaf,fill=softred]
      ]
      [Execution Feedback (\S~\ref{sec:feedback_exe}), leaf2, fill=softred,
        [PerfCodeGen~\cite{peng2024perfcodegen}{,}
           ExecutionAgent~\cite{bouzenia2024you}{,}
           TypeTest~\cite{liu2025llm}{,}
           ProjectTest~\cite{wang2025projecttest}{,}
           PyCapsule~\cite{adnan2025large}{,}
           ASTER~\cite{pan2025aster}{,}
           CodeT~\cite{chen2022codet}{,}
           STROT~\cite{rath2025structured}{,}
           KPC~\cite{ren2023from}{,}
           APT~\cite{zhang2024llm}{,}
           Seeker~\cite{zhang2024seeker}{,}
           TestGen-LLM~\cite{alshahwan2024automated}{,}
           Healer~\cite{sun2024llm}{,}
           Agentic workflow~\cite{schultz2025potential}{,}
           etc{.}
            ,leaf,fill=softred]
      ]
      [Human Feedback (\S~\ref{sec:feedback_hum}), leaf2, fill=softred,
        [Grounded Copilot~\cite{barke2023grounded}{,}
           Cutting-Search~\cite{marozzo2025iterative}{,}
           Deep Sets~\cite{zaheer2017deep}{,}
           MA-RLHF~\cite{chai2024ma}{,}
           ClarifyGPT~\cite{mu2024clarifygpt}{,}
           PPO~\cite{schulman2017proximal}{,}
           RLSF~\cite{niekerk2025post}{,}
           CAGSR–vLLM–MTC~\cite{kiruluta2025history}{,}
           etc{.}
            ,leaf,fill=softred]
      ]
      [Self-Refinement Feedback (\S~\ref{sec:feedback_self}), leaf2, fill=softred,
        [CRITIC~\cite{gou2023critic}{,}
           BioAgents~\cite{mehandru2025bioagents}{,}
           Self-Organized Agents~\cite{ishibashi2024self}{,}
           Multi-Agent Collaboration~\cite{talebirad2023multi}{,}
           LLM-Agent-UMF~\cite{hassouna2024llm}{,}
           SWE-Search~\cite{antoniades2024swe}{,}
           N-Critics~\cite{mousavi2023critics}{,}
           Self-Review Framework~\cite{park2025self}{,}
           SAMULE~\cite{ge2025samule}{,}
           Reflexion~\cite{shinn2023reflexion}{,}
           etc{.}
            ,leaf,fill=softred]
      ]
    ]
  ]
\end{forest}

\caption{The taxonomy of Vibe Coding is categorized into large language model foundations, coding agent architectures, development environments, and feedback mechanisms. Each area encompasses specific techniques and frameworks that collectively advance the systematic integration of LLMs and agents into intelligent and collaborative software development workflows.}
\label{fig:tree}
\end{figure*}
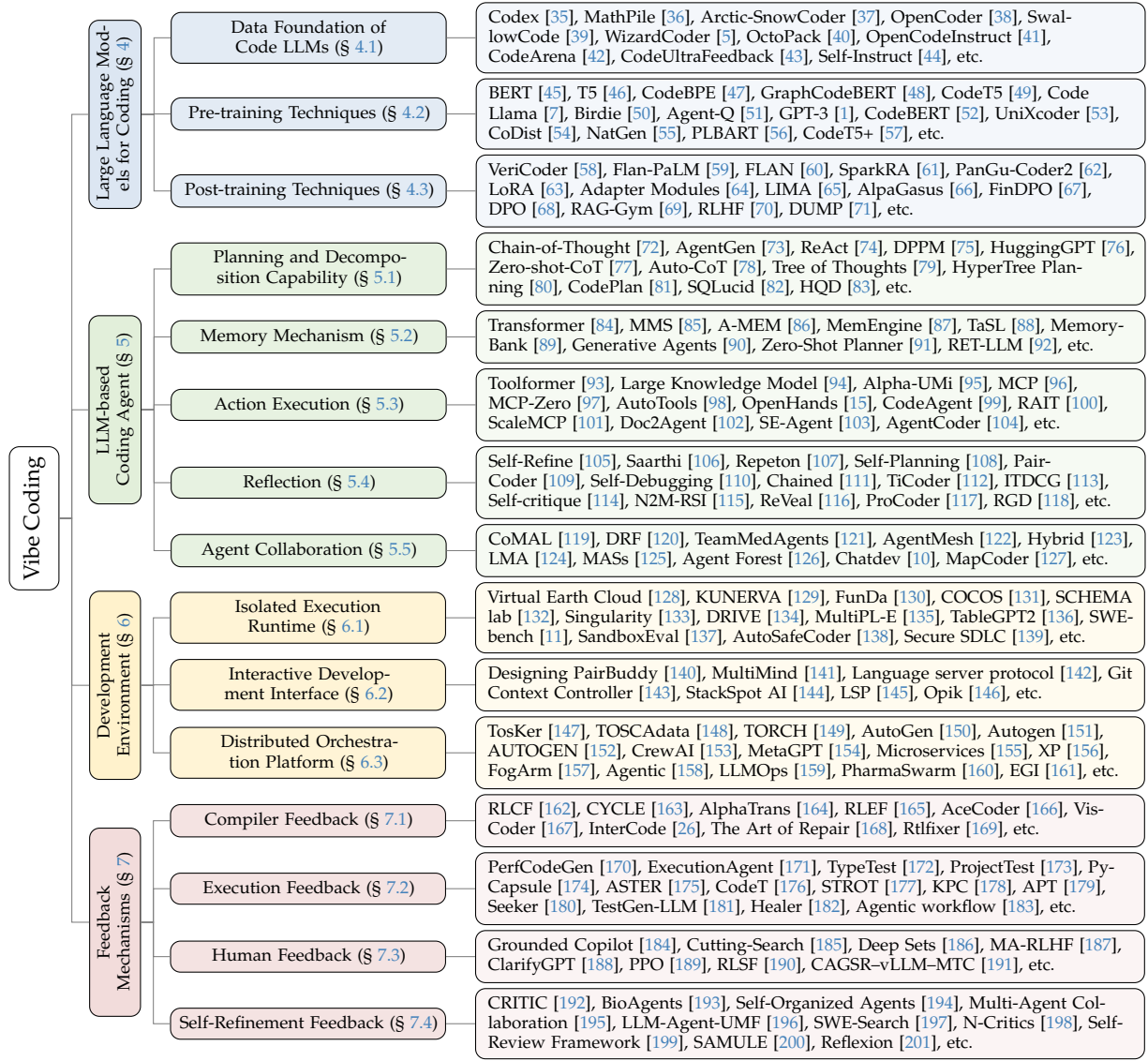

\section{Related Work}
\label{sec:related}

\subsection{Related Surveys}

\paragraph{Foundational LLM.}
Several surveys have documented the evolution of large language models, covering LLM architectures, training paradigms, and capabilities~\cite{minaee2024large, zheng2023survey}. These works examine Transformer variants with a focus on efficient architectures and long-context capabilities~\cite{tay2022efficient, wan2023efficient, liu2025comprehensive}. The historical trajectory from BERT to ChatGPT traces the development of foundation models, examining both opportunities and risks~\cite{zhou2024comprehensive}. Specialized surveys address evaluation methodologies, efficiency from model-centric and data-centric perspectives, and specific capabilities including text generation and knowledge-enhanced PLMs~\cite{chang2024survey}.

\paragraph{In-Context Learning.}
Building on these architectural foundations, research has turned to techniques for utilizing pre-trained models without additional training. Prompt engineering and in-context learning have emerged as fundamental techniques, with taxonomies covering extensive prompting methods, applications, security considerations, and performance across Natural Language Processing (NLP) tasks~\cite{schulhoff2024prompt, sahoo2024systematic, vatsal2024survey}. In-context learning mechanisms are explored with context engineering emerging as a formal discipline~\cite{dong2022survey, mei2025survey}. Chain-of-Thought (CoT) reasoning has proven particularly effective, with taxonomies examining Chain-of-X paradigms and investigating long CoT and multimodal CoT reasoning~\cite{chu2023navigate, xia2024beyond, wang2025multimodal}. Multimodal large language models represent a rapidly advancing frontier, with surveys examining architectures, training methods, and vision-language integration across multiple data modalities~\cite{yin2024survey, zhang2024mm, liang2024comprehensive, wu2023multimodal}.

\paragraph{Post-Training.}
When in-context learning proves insufficient, post-training methods offer pathways to align models with specific requirements and enhance reasoning capabilities. Reinforcement learning approaches, including Proximal Policy Optimization (PPO), Q-Learning, and Actor-Critic methods, have been surveyed with particular emphasis on Reinforcement Learning from Human Feedback (RLHF), Reinforcement Learning from AI Feedback (RLAIF), and Direct Preference Optimization (DPO)~\cite{srivastava2025technical, zhang2025survey, cao2024survey, kaufmann2024survey, ma2023training}. Instruction tuning and supervised fine-tuning methodologies are reviewed covering dataset construction and training strategies, with examinations of data selection approaches for enhanced instruction-following capabilities~\cite{shengyu2023instruction}. Alignment research categorizes methods into outer and inner alignment with adversarial considerations, while exploring training-free alignment and personalized alignment techniques~\cite{tie2025survey}. DPO emerges as an reinforcement learning RL-free alternative to RLHF, with taxonomies categorizing data strategies, learning frameworks, and constraint mechanisms~\cite{xiao2024comprehensive}. Post-training paradigms are explored covering fine-tuning, alignment, reasoning, efficiency, and domain adaptation, with parameter-efficient methods including Low-Rank Adaptation (LoRA) and adapters providing experimental comparisons of computational overhead~\cite{xu2023parameter}.

\paragraph{Agent Systems.}
The integration of tool use and planning capabilities transforms LLMs from passive models into active agents. Foundational surveys establish frameworks covering agent construction, brain-perception-action architectures, and autonomous decision-making capabilities, while providing unified taxonomies across benchmarks spanning reasoning and code generation~\cite{su2024language, zhao2023depth, ferrag2025llm}. Multi-agent systems are examined covering agent profiling, communication protocols, and collaborative workflows across complex task-solving scenarios~\cite{guo2024large, chen2024survey}. Agent capabilities are addressed through specialized surveys: tool use with Retrieval-Augmented Generation (RAG) and feedback learning~\cite{huang2024understanding, masterman2024landscape}, planning mechanisms including task decomposition and memory~\cite{zhang2025survey}, single and multi-agent architectures for reasoning and tool execution, and memory mechanisms with short-term and long-term analysis~\cite{xu2025llm}. Evaluation methodologies cover planning, tool use, self-reflection, and application-specific benchmarks~\cite{yehudai2025survey, mohammadi2025evaluation}. Domain-specific applications span web automation~\cite{ning2025survey}, scientific discovery across life sciences and materials science~\cite{wei2025ai}, Operating System agents with GUI interaction~\cite{hu2025agents}, and self-evolving agents with feedback loops and lifelong learning capabilities~\cite{fang2025comprehensive}. Of particular relevance to this work, recent surveys examine coding agents with single and multi-agent architectures across the software development lifecycle, covering planning, context management, and tool integration with benchmarking frameworks~\cite{dong2025survey, wang2025ai}.

\subsection{Preliminary}

\paragraph{Reinforcement Learning for Code Generation.}
Applying reinforcement learning to code generation requires executable feedback signals, evolving from basic mechanisms to sophisticated training paradigms. Early approaches combine pretrained language models with deep RL using unit test feedback and critical sampling, achieving strong performance on competitive benchmarks~\cite{le2022coderl}. Execution-based methods leverage PPO with compiler feedback for real-time refinement~\cite{shojaee2023execution}. Advanced RL frameworks employ multi-granularity unit test feedback~\cite{liu2023rltf}, break generation into curriculum subtasks with fine-grained optimization~\cite{dou2024stepcoder}, use ranking-based alignment mechanisms~\cite{shen2023pangu}, and leverage Group Relative Policy Optimization with compiler feedback to achieve competitive performance~\cite{deepseekai2024deepseek}.

\paragraph{Autonomous Coding Agent Systems.}
Beyond supervised generation, autonomous agents tackle complete software engineering tasks through specialized architectures and multi-agent collaboration. Single-agent systems introduce custom agent-computer interfaces achieving strong benchmark performance~\cite{yang2024swe}, combine structure-aware code search with spectrum-based fault localization for cost-effective issue resolution~\cite{zhang2024autocoderover}, and demonstrate that simple hierarchical localization can outperform complex agents~\cite{xia2024agentless, ma2025alibaba}. Repository-level code generation integrates programming tools with agent strategies showing substantial improvements~\cite{zhang2024codeagent, pan2024enhancing}, while iterative retrieval-generation approaches improve baselines on repository evaluation benchmarks~\cite{pan2024enhancing}. Multi-agent frameworks employ specialized roles across the development process. These assign distinct programmer, test designer, and test executor agents achieving high pass rates with reduced token consumption~\cite{huang2023agentcoder}, encode standardized operating procedures into assembly-line paradigms with role-based collaboration achieving near-perfect task completion~\cite{hong2024metagpt}, and implement chat chain architectures with dual-agent communication covering the full software development lifecycle at minimal cost~\cite{qian2023chatdev}. Specialized frameworks replicate human programming cycles~\cite{islam2024mapcoder}, investigate test-driven development principles showing consistent improvements with upfront test provision~\cite{mathews2024test}, and analyze self-repair mechanisms revealing modest performance gains bottlenecked by self-feedback abilities~\cite{chen2023teaching}.

\paragraph{Function Calling.}
Effective agent systems require mechanisms for tool use, function calling, and execution infrastructure to interact with external systems and APIs. Function calling frameworks teach language models to self-supervise tool use with simple APIs requiring minimal demonstrations~\cite{schick2023toolformer}, use teacher models for tool-related instruction data generation~\cite{yang2023gpt4tools}, and employ multi-task learning on granular tasks with large-scale API datasets~\cite{abdelaziz2024granite}. Execution and interaction environments provide lightweight RL frameworks with safe execution for multiple programming languages~\cite{yang2023intercode}, use executable code as unified action space outperforming alternative formats with specialized instruction datasets and fine-tuned models~\cite{wang2024executable}, and integrate Monte Carlo Tree Search with external feedback for deliberate problem-solving~\cite{zhou2023language}. Optimization and deployment advances automatically identify parallelizable function calls achieving significant latency speedup and cost savings~\cite{kim2024llm}, enable edge deployment with compact models matching large model capabilities locally~\cite{erdogan2024tinyagent}, and coordinate specialized agents for tool selection, execution, and calibration achieving higher success rates~\cite{wu2023autogen}. Retrieval, evaluation, and infrastructure contributions provide large-scale benchmarks showing retrieval from diverse datastores significantly improves performance~\cite{wang2024coderag}, decompose tool use capability into evaluation aspects for API manipulation~\cite{liu2024evaluation}, develop multi-agent systems for code review automation achieving strong results on critical review tasks~\cite{wu2023autogen}, enable customizable conversable agents programmed in natural or code language~\cite{wu2023autogen}, and provide massive instruction-API pair datasets demonstrating cross-language transfer enabling open-source models to outperform proprietary alternatives on new API calls~\cite{guo2024api}.

\paragraph{Supervised Fine-Tuning.}
An alternative to reinforcement learning lies in supervised fine-tuning and instruction tuning methods, which have become foundational for code model training. Instruction evolution approaches iteratively evolve code instructions achieving competitive performance~\cite{luo2023wizardcoder}, while self-instruction methods bootstrap instruction-following capabilities~\cite{wang2022self}. Fully transparent self-alignment pipelines enable training without proprietary distillation~\cite{wei2024selfcodealign}, and large-scale instruction datasets combine multiple instruction generation approaches~\cite{ahmad2025opencodeinstruct}. Specialized tuning addresses security-centric generation improving code security substantially~\cite{he2024instruction}, optimization code generation~\cite{ma2024llamoco}, and code editing, debugging, and refinement through self-debugging approaches and iterative refinement without additional training~\cite{chen2023teaching, madaan2023self, ma2024lingma}. Foundation models for code employ repository-level pretraining with extended context windows~\cite{guo2024deepseek}, train on diverse languages with fill-in-the-middle objectives and PII redaction~\cite{li2023starcoder}, incorporate GitHub issues and documentation with repository-level training~\cite{lozhkov2024starcoder}, and achieve competition-level performance using massive sampling and filtering strategies~\cite{li2022competition}.

\begin{figure}[t]
    \centering
    \includegraphics[width=\textwidth]{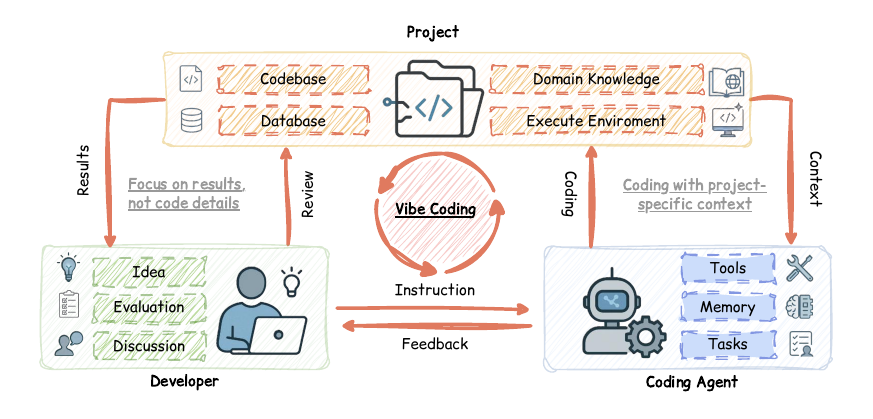}
    \caption{\textbf{Vibe Coding Overview}. A triadic collaboration among developer, coding agent, and project, where iterative instruction–feedback loops enable context-aware coding and result-oriented review.}
    \label{fig:vibe_coding}
\end{figure}
\section{Vibe Coding: The Engineering of Managing Coding Agents}
\label{sec:vibe_coding}

\subsection{Definition of Vibe Coding}
\label{sec:definition}

In this paper, we define Vibe Coding as an engineering methodology for software development grounded in large language models \citep{karpathy2025, ray2025review, sapkota2025vibe, horvat2025vibe, moore2025vibe}. At its core, as shown in Figure~\ref{fig:vibe_coding}, it represents \textbf{a dynamic triadic relationship among human developers, software projects, and Coding Agents}. In this paradigm, humans evolve from direct code authors into intent articulators, context curators, and quality arbiters. Projects extend beyond static code repositories to become multifaceted information spaces encompassing codebases, databases, and domain knowledge. Coding Agents, as intelligent executors, perform code generation, modification, and debugging under the dual guidance of human intent and project constraints.

\paragraph{Formalization of the Triadic Relationship.}
We model Vibe Coding as a dynamic interactive system defined by the triple $\mathcal{V} = \langle \mathcal{H}, \mathcal{P}, \mathcal{A}_\theta \rangle$, where:

\begin{itemize}
    \item $\mathcal{H}$: Human developer, equipped with requirement cognition capability $\mathcal{H}_{\text{req}}: \mathcal{D} \rightarrow \mathcal{I}$ (translating domain requirements $\mathcal{D}$ into instructions $\mathcal{I}$) and quality discrimination capability $\mathcal{H}_{\text{eval}}: \mathcal{O} \rightarrow \{0,1\} \times \mathcal{F}$ (judging outputs $\mathcal{O}$ with accept/reject decisions and feedback $\mathcal{F}$),
    
    \item $\mathcal{P}$: Software project, represented as a project context space $\mathcal{P} = \langle \mathcal{C}_{\text{code}}, \mathcal{C}_{\text{data}}, \mathcal{C}_{\text{know}} \rangle$, corresponding to codebase, database, and domain knowledge respectively,
    
    \item $\mathcal{A}_\theta$: Coding Agent, a large language model parameterized by $\theta$, executing the conditional generation function $\mathcal{A}_\theta: \mathcal{I} \times \mathcal{P} \times \mathcal{E} \rightarrow \mathcal{O}$.
\end{itemize}

The collaboration among the three parties can be modeled as a Constrained Markov Decision Process (Constrained MDP) \citep{altman2021constrained}, wherein humans define the goal space and constraint boundaries, projects provide the state space and transition constraints, and Agents execute policies and state transitions:

\begin{equation}
\mathcal{V}_{\text{MDP}} = \langle \mathcal{S}_{\mathcal{P}}, \mathcal{A}_{\mathcal{H} \rightarrow \mathcal{A}_\theta}, \mathcal{T}_{\mathcal{A}_\theta | \mathcal{P}}, \mathcal{R}_{\mathcal{H}}, \gamma \rangle,
\end{equation}

where the state space $\mathcal{S}_{\mathcal{P}}$ is defined by the project's current state, the action space $\mathcal{A}_{\mathcal{H} \rightarrow \mathcal{A}_\theta}$ is triggered by human instructions to Agent behaviors, the transition function $\mathcal{T}_{\mathcal{A}_\theta | \mathcal{P}}$ is constrained by project specifications, the reward function $\mathcal{R}_{\mathcal{H}}$ is determined by human evaluation, and $\gamma$ is the discount factor.

\paragraph{Agent's Conditional Generation Process.}
Given human intent $\mathcal{I}$, project context $\mathcal{K} \subseteq \mathcal{P}$ (a relevant subset retrieved from the project information space), and execution environment $\mathcal{E}$, the Agent generates code sequence $Y = (y_1, \ldots, y_T)$ in an auto-regressive manner, with joint probability factorized as:

\begin{equation}
P_\theta(Y|\mathcal{I}, \mathcal{K}, \mathcal{E}) = \prod_{t=1}^{T} P_\theta(y_t|y_{<t}, \mathcal{C}_t),
\end{equation}

where $\mathcal{C}_t = \mathcal{A}(\mathcal{I}, \mathcal{K}, \mathcal{E}, y_{<t})$ denotes the dynamic context at step $t$, orchestrated by the high-level assembly function $\mathcal{A}$. Context components $c_i$ correspond to distinct information sources from the triadic relationship:

\begin{figure}[t]
    \centering
    \includegraphics[width=\linewidth]{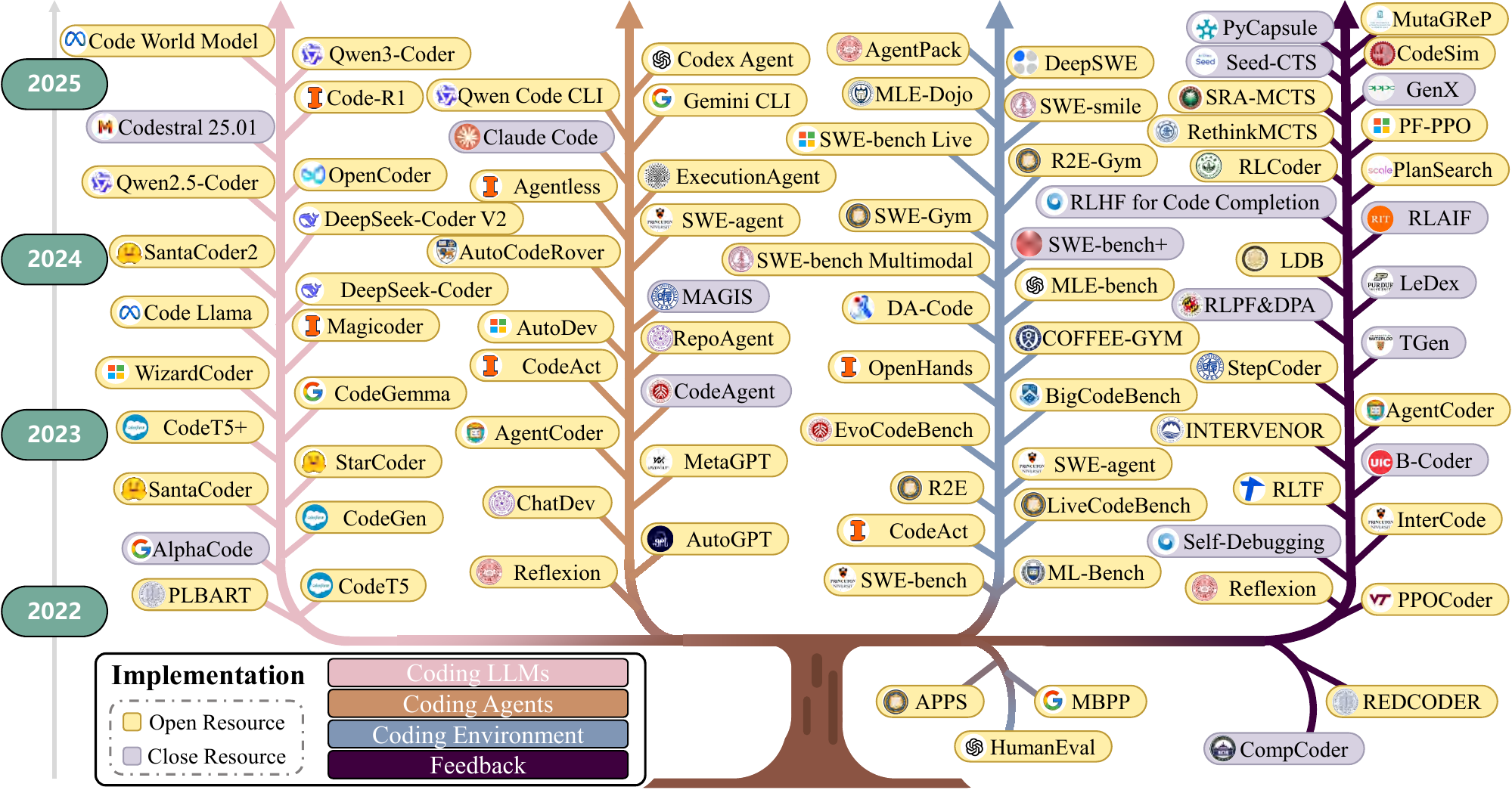}
    \caption{Vibe Coding Evolution Timeline: A comprehensive visualization of the development trajectory of Vibe Coding implementations from 2022 to 2025, showing the evolution from foundational Coding LLMs to sophisticated coding agent systems, execute environments and Feedback mechanism.}
    \label{fig:his}
\end{figure}

\textbf{Human Layer:} 
\begin{itemize}
    \item $c_{\text{instr}}$: System instructions and task requirements.
\end{itemize}

\textbf{Project Layer:}
\begin{itemize}
    \item $c_{\text{code}}$: Codebase (source code, API interfaces, architectural design),
    \item $c_{\text{data}}$: Database (persistent data, data schemas),
    \item $c_{\text{know}}$: Domain knowledge (documentation, specifications, best practices).
\end{itemize}

\textbf{Agent Layer:}
\begin{itemize}
    \item $c_{\text{tool}}$: Definitions and signatures of callable tools (compilers, testing frameworks, version control),
    \item $c_{\text{mem}}$: Historical interaction memory (multi-turn dialogue context, prior decision records),
    \item $c_{\text{tasks}}$: Current tasks (pending actions, task queue, execution status).
\end{itemize}

\paragraph{Optimization Objective of Vibe Coding.}
From the triadic perspective, the core challenge of Vibe Coding is to identify optimal context orchestration strategies $\mathcal{F}^* = \{\mathcal{A}, \text{Retrieve}, \text{Filter}, \text{Rank}\}$ that maximize generation quality within the limited context window $L_{\max}$. Defining a reward function $R: \mathcal{O} \times \mathcal{O}^* \rightarrow \mathbb{R}$ that measures alignment between generated outputs and human expectations, given task distribution $\mathcal{T}$, the optimization objective is:

\begin{equation}
\mathcal{F}^* = \arg\max_{\mathcal{F}} \mathbb{E}_{\tau \sim \mathcal{T}} [R(P_\theta(Y|\mathcal{C}_{\mathcal{F}}(\tau)), Y_\tau^*)] \quad \text{s.t.} \quad |\mathcal{C}_{\mathcal{F}}(\tau)| \leq L_{\max}
\end{equation}

where $\mathcal{C}_{\mathcal{F}}(\tau)$ represents the context retrieved and filtered from project $\mathcal{P}$ by strategy $\mathcal{F}$ for task $\tau$, and $Y_\tau^*$ denotes the ideal output in the human's mental model. This optimization problem essentially seeks the subset with maximum information gain for Agent generation from the exponentially large candidate set in the project information space.

\paragraph{Human-Agent Collaborative Loop and Task Evolution.}
The central mechanism of Vibe Coding is human guidance through continuous feedback to steer the Agent toward project objectives, with \textbf{explicit support for humans to iteratively expand constraints and introduce new tasks}, forming a dynamically evolving requirement space. Let the Agent produce output $o_k \in \mathcal{O}$ at iteration $k$. The human supervision function $\mathcal{H}: \mathcal{O} \times \mathcal{R} \times \mathcal{I} \rightarrow 2^{\mathcal{O}} \times (\mathcal{F} \cup \mathcal{I}_{\text{new}})$, constrained by execution results $\mathcal{R}$ and current instructions $\mathcal{I}_k$, returns an acceptance subset $\mathcal{A}_k \subseteq o_k$ and a correction signal $\delta_k$. The iterative evolution is expressed as:

\begin{equation}
(o_{k+1}, \mathcal{I}_{k+1}) = \begin{cases} 
(o_k, \mathcal{I}_k) & \text{if } \mathcal{A}_k = o_k \text{ (full acceptance, terminate)} \\
(\mathcal{A}_\theta(o_k \setminus \mathcal{A}_k; \delta_k, \mathcal{I}_k, \mathcal{K}), \mathcal{I}_k) & \text{if } \delta_k \in \mathcal{F} \text{ (local refinement)} \\
(\mathcal{A}_\theta(\mathcal{I}_k \cup \{\delta_k\}, \mathcal{K}), \mathcal{I}_k \cup \{\delta_k\}) & \text{if } \delta_k \in \mathcal{I}_{\text{new}} \text{ (requirement expansion)}
\end{cases}
\end{equation}

where $(\mathcal{A}_k, \delta_k) = \mathcal{H}(o_k, \mathcal{R}_k, \mathcal{I}_k)$ with $\mathcal{R}_k$ denoting the execution feedback obtained from running $o_k$ in environment $\mathcal{E}$. 

\paragraph{Formalization of Iterative Task Expansion.}
Distinct from conventional software development paradigms where requirements are frozen early, Vibe Coding supports \textbf{incremental evolution of requirements}. We define the task evolution trajectory as an instruction set sequence $\{\mathcal{I}_0, \mathcal{I}_1, \ldots, \mathcal{I}_K\}$, where $\mathcal{I}_0$ represents the initial requirements and each expansion satisfies monotonicity $\mathcal{I}_k \subseteq \mathcal{I}_{k+1}$. The $k$-th expansion is formalized as:

\begin{equation}
\mathcal{I}_{k+1} = \mathcal{I}_k \oplus \Delta\mathcal{I}_k = \mathcal{I}_k \cup \{\delta_k^{(1)}, \delta_k^{(2)}, \ldots, \delta_k^{(m_k)}\}
\end{equation}

where $\oplus$ denotes the instruction merging operator, $\Delta\mathcal{I}_k$ represents the instruction set added at step $k$, and $m_k$ is the number of tasks introduced in this expansion. The cumulative task count throughout the development process is $|\mathcal{I}_K| = |\mathcal{I}_0| + \sum_{k=0}^{K-1} m_k$. This multi-expansion mechanism embodies two key characteristics: (1) \textbf{Progressive Requirement Clarification}: Humans need not exhaustively plan all details initially, but can progressively refine constraints upon observing Agent outputs. This reflects the cognitive reality that complex system requirements often crystallize through interactive exploration rather than upfront specification. (2) \textbf{Dynamic Constraint Reinforcement via Feedback}: When Agent outputs expose implicit requirements or boundary cases, humans can immediately supplement constraints. This interactive requirement discovery process enables adaptive specification refinement aligned with emergent understanding.

Formally, we model the entire development cycle as a multi-stage optimization problem, where each stage $k$ corresponds to a task space $\mathcal{I}_k$:

\begin{equation}
\max_{\{o_k\}_{k=0}^K} \sum_{k=0}^{K} \omega_k \cdot R(o_k, Y_{\mathcal{I}_k}^*) \quad \text{s.t.} \quad o_k = \mathcal{A}_\theta(\mathcal{I}_k, \mathcal{K}, \mathcal{E}), \quad \mathcal{I}_k \subseteq \mathcal{I}_{k+1}
\end{equation}

where $\omega_k$ is the weight factor for stage $k$, and $Y_{\mathcal{I}_k}^*$ denotes the ideal output for that stage. This formulation captures the essence of Vibe Coding: through sustained human intervention and dynamic expansion of the task space, the system progressively converges toward the ultimate software objective.

This iterative and expansive mechanism embodies the core philosophy of Vibe Coding: \textbf{Humans govern ``What'' (defining the right problems, with problems allowed to evolve dynamically) and ``Why'' (judging solution appropriateness); Projects define ``Where'' (constraining the solution space boundaries); Agents manage ``How'' (exploring technical implementation pathways)}. The synergy of these three entities constitutes a self-adaptive, requirement-evolvable closed-loop software development system.

\subsection{Why Vibe Coding}
\label{sec:why_vibe_coding}

Vibe Coding transforms software development from passive assistance to collaborative partnerships, addressing challenges in democratization, workflow efficiency, and ecosystem expansion.

\paragraph{Team-Scale Capabilities for Individual Developers.}
Vibe Coding enables individual developers to deliver team-scale capabilities. Production applications traditionally require coordinating frontend, backend, database, security, DevOps, and QA specialists with substantial overhead. Coding agents provide diverse expertise across domains \citep{qian2023chatdev, wu2023autogen, dong2024villageragent}. Developers focus on requirements while agents implement across stacks \citep{dong2023self}. This reduces learning overhead through context engineering. Individual developers now implement cloud infrastructure and performance optimizations without formal training, benefiting resource-constrained entities by compressing prototypes from weeks to days.

\paragraph{Continuous Development and Quality Convergence.}
Vibe Coding aims to balance development velocity and code quality—historically conflicting objectives. Traditional workflows trade delivery speed against testing rigor \citep{dima2018waterfall}. Vibe Coding can improve both through autonomous iteration decoupled from human constraints. Agents sustain 24/7 operation: automated testing, refactoring, and performance profiling \citep{Janzen2005TestdrivenDC}. Automating mechanical tasks liberates cognitive resources for design and optimization \citep{goues2019automated, zhang2024autocoderover}. Agents enable exhaustive exploration through rapid iteration \citep{shinn2023reflexion, madaan2023self} while human oversight validates architectural decisions.

\paragraph{Broadening the Software Creator Ecosystem.}
Vibe Coding democratizes development by lowering technical barriers. Traditional development required extensive programming knowledge before implementing ideas. Natural language becomes the primary creation interface \citep{nijkamp2023codegen, luo2023wizardcoder, brown2020language}. Domain experts—medical practitioners, educators, designers—articulate needs without computer science education \citep{chen2021evaluating}. This diversifies innovation sources, materializing underrepresented perspectives \citep{xi2023rise, park2023generative}. Economic impact manifests through creator economy expansion: domain experts monetize specialized tools without technical co-founders. This parallels previous democratization waves, representing software literacy's evolution from specialized skill to broadly accessible capability \citep{clyburnesherin2019computational}.

\begin{figure}[t]
    \centering
    \includegraphics[width=\linewidth]{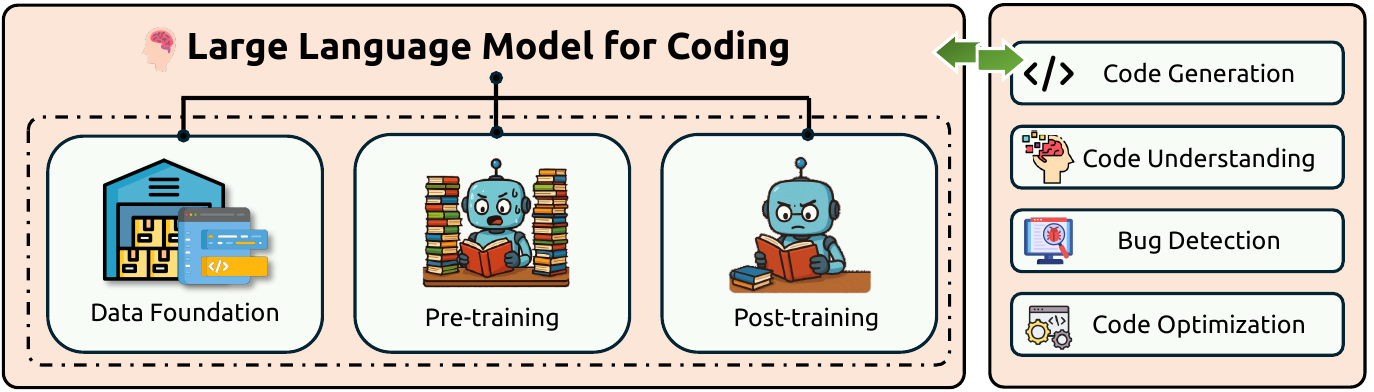}
    \caption{Large Language Model for Coding: Overview of the training lifecycle encompassing Data Foundation, Pre-training, and Post-training stages, with core capabilities including Code Generation, Code Understanding, Bug Detection, and Code Optimization.}
    \label{fig:coding_llms}
\end{figure}

\section{Large Language Models for Coding}
\label{sec:llm_coding}
\subsection{Data Foundation of Code LLMs} \label{sec:llm_coding_data}

\subsubsection{Pre-training Code Corpora}

Code LLMs require large-scale training data from diverse sources \citep{chen2021evaluating, xu2022systematic}. These models rely on large-scale code corpora primarily sourced from open platforms like GitHub and Stack Overflow, with quality filtering based on metrics such as repository stars, documentation completeness, and community engagement \citep{jiang2024survey}. 

Training datasets vary significantly in composition and curation strategies, with two primary approaches emerging: depth-focused strategies that concentrate on popular languages for quality, and breadth-focused strategies that span diverse languages for coverage \citep{albalak2024survey, longpre2023flan}. The Stack dataset uses the depth-focused approach, providing 3.1 TB of permissively licensed source code across 30 programming languages with careful attention to licensing and data provenance \citep{wang2023mathpile, aryabumi2024code, wei2024arctic}, while CodeParrot similarly trains exclusively on GitHub code to achieve depth in popular languages \citep{xu2022systematic}. In contrast, breadth-focused approaches span diverse languages for maximum coverage. The Stack v2 expands coverage substantially to 67.5 TB across 619 programming and markup languages \citep{wei2024arctic, huang2024opencoder}, GPT-Neo utilizes diverse mixed corpora including "the Pile" \citep{xu2022systematic}, while CodeLlama employs SlimPajama's 627 billion tokens paired with code from The Stack \citep{aryabumi2024code} and Arctic-SnowCoder uses a filtered combination yielding 400 billion tokens \citep{wei2024arctic}. 

Data quality has driven sophisticated processing pipelines, with RefineCode incorporating over 130 carefully crafted language-specific filtering and cleaning rules across 607 languages to ensure syntactic and semantic validity \citep{huang2024opencoder}, and SwallowCode implementing comprehensive four-stage pipelines performing syntax validation, quality assessment, deduplication, and LLM-based rewriting to enhance training data quality \citep{fujii2025rewriting}.

\subsubsection{Instruction \& Preference Datasets}

Beyond raw code corpora, instruction-following requires curated training data. Accordingly, code LLMs have demonstrated exceptional capabilities across diverse programming tasks \citep{jiang2024survey, luo2023wizardcoder}, with instruction tuning as a critical technique for enhancing instruction-following abilities and solution generation quality \citep{luo2023wizardcoder}. These instruction datasets are typically compiled from two primary sources: permissively licensed source code repositories and synthetically constructed instructional data \citep{jiang2024survey}. Notable instruction datasets in the field include CommitPack with its impressive 4 TB of raw data carefully filtered to 2GB of high-quality code paired with commit message instructions that capture developer intent \citep{muennighoff2023octopack}, OpenCodeInstruct as the largest openly available dataset with 5 million diverse samples covering multiple programming languages and task types \citep{ahmad2025opencodeinstruct}, and SynCode-Instruct providing nearly 20 billion tokens of synthetic instruction-following examples \citep{yang2024evaluating}. For preference learning applications, CodeUltraFeedback provides 10,000 coding instructions each paired with four distinct responses from 14 different LLMs, all evaluated comprehensively by GPT-3.5 to establish quality rankings \citep{weyssow2024codeultrafeedback}, while the PLUM Framework innovatively employs GPT-4-generated unit tests to establish automated preference rankings based on functional correctness \citep{chen2024mastering}.

High-quality instruction dataset construction methodologies have evolved significantly from expensive and time-consuming human annotation processes toward more scalable synthetic generation approaches that maintain quality while reducing costs \citep{ahmad2025opencodeinstruct, ouyang2022training}. The Self-Instruct paradigm pioneered the approach of bootstrapping from limited seed examples, using language models to generate new instructions through carefully designed instruction-input-output pipelines \citep{wang2022self}, while Evol-Instruct introduced the concept of progressive complexity increase by iteratively rewriting and enhancing initial instructions to create more challenging and diverse training examples \citep{xu2023wizardlm}. OSS-Instruct advances the field by systematically integrating real-world open-source code snippets as seeds for generating diverse, contextually realistic instruction data that better reflects actual programming scenarios \citep{wei2023magicoder}. Quality enhancement techniques have become sophisticated, employing approaches like CoachLM's LLM-based iterative rewriting to improve instruction clarity and task alignment, and CaR (Critique and Revise) approaches that use specialized scoring models aligned with expert preferences to filter and refine synthetic data, with the effectiveness of these synthetic data approaches demonstrated by systems like Nvidia's Nemotron-4 achieving state-of-the-art performance using 98\%~synthetically generated alignment data \citep{zheng2024opencodeinterpreter}.

\subsection{Pre-training Techniques} \label{sec:llm_coding_pre}
\subsubsection{Pre-training Objectives}

Large language models for code leverage transfer learning paradigms inherited from BERT and GPT architectures \citep{devlin2019bert, qiu2020pre, zhou2024comprehensive, chirkova2023codebpe, raffel2019exploring}, with pre-training on large-scale data before task-specific fine-tuning proving particularly effective where LLMs outperform domain-specific models in limited labeled data scenarios \citep{chirkova2023codebpe}. Code-specific pre-training objectives surpass traditional masked language modeling approaches, with innovations including GraphCodeBERT's data flow prediction that captures semantic relationships in code \citep{chirkova2023codebpe, guo2020graphcodebert} and variable naming tasks in CodeT5 and DOBF that leverage programming language structure \citep{chirkova2023codebpe, wang2021identifier, roziere2023code}.

Pre-training serves the fundamental purpose of instilling general-purpose knowledge and pattern recognition capabilities from large-scale unlabeled datasets \citep{blouir2024birdie, jern2025agent}, with several well-established paradigms dominating the landscape. Autoregressive language modeling predicts the next token in a sequence based on all previous context \citep{li2025generated, jern2025agent}, while masked language modeling focuses on predicting randomly masked tokens using both preceding and following context for bidirectional understanding \citep{li2025generated, jern2025agent}. Alternative objectives that have gained traction include infilling tasks that predict missing spans within sequences and prefix language modeling that generates continuations from given prefixes \citep{blouir2024birdie}.

Masked Language Modeling is the dominant approach for code understanding tasks, building directly on BERT's remarkable success in natural language processing \citep{liu2020multi, guo2022unixcoder, devlin2019bert}. This approach is primarily adopted by encoder-only models like CodeBERT, which utilizes RoBERTa-based architectures for bidirectional code understanding \citep{feng2020codebert}. The standard implementation samples approximately 15\% of input tokens, with 80\% of selected tokens replaced by a special [MASK] token to enable bidirectional context-based prediction \citep{guo2022unixcoder}.

Implementation details vary across models, with CodeBERT applying MLM augmented with Replaced Token Detection to improve robustness \citep{huang2023program, yang2023enhancing, feng2020codebert}, while GraphCodeBERT extends this with structure-aware features discussed below \citep{guo2020graphcodebert}.

Autoregressive language modeling serves as the foundational approach for code generation tasks by predicting the next token based solely on preceding context \citep{liu2020multi}. Decoder-only models like CodeGPT effectively leverage GPT architecture's powerful Transformer-based decoders for this purpose \citep{liu2022automatically, devlin2019bert}. Recent models like CodeLlama have extended this paradigm through innovative fill-in-the-middle objectives that enable models to generate missing code segments given both prefix and suffix contexts, proving valuable for code completion tasks \citep{saad2025senai, li2025revisiting}.

In contrast to unidirectional prediction, denoising objectives have proven highly effective for encoder-decoder architectures through span-level reconstruction tasks \citep{guo2022unixcoder}. PLBART first applied denoising approaches to code by building on BART's sequence-to-sequence architecture \citep{chakraborty2022natgen, ahmad2021unified, yang2023enhancing, lyu2024automatic}, while modern unified models increasingly adopt sophisticated denoising strategies in comprehensive multi-task frameworks exemplified by CodeT5+ which synergistically combines span denoising with contrastive learning objectives \citep{wang2023open, zheng2023survey}.

Structure-aware objectives incorporate the unique structural characteristics inherent to programming languages, including abstract syntax trees, control flow graphs, and data flow relationships \citep{feng2020codebert, devlin2019bert}. GraphCodeBERT incorporates data flow graphs that explicitly encode semantic relationships between variables across code segments \citep{chirkova2023codebpe, guo2020graphcodebert}, while introducing specialized structure-aware tasks such as code structure edge prediction that require understanding syntactic and semantic code organization \citep{lin2024scaling, guo2020graphcodebert}. Contrastive learning explicitly minimizes representation distance between semantically similar code functions while simultaneously maximizing distance between dissimilar code, and is a powerful technique for training code models \citep{jain2020contrastive, zhou2024large}. UniXcoder uses this approach through sophisticated multi-modal contrastive learning that unifies natural language comments and linearized abstract syntax tree representations into a shared semantic space \citep{guo2022unixcoder, dharma2023neural}, while ContraCode leverages automated compiler transformations to generate semantically equivalent code variations for robust contrastive data augmentation \citep{jain2020contrastive}. Multimodal pre-training strategies leverage multiple complementary data modalities to enhance model capabilities, with CodeBERT first learning general-purpose code representations through carefully aligned bimodal natural language-programming language pairs \citep{feng2020codebert, guo2022unixcoder, lin2024scaling, lyu2024automatic}, while advanced models like CodeT5+ implement sophisticated two-stage training procedures involving initial unimodal pre-training on massive code-only data to establish programming knowledge, followed by bimodal pre-training on carefully curated code-text pairs with explicit cross-modal alignment learning to bridge natural and programming languages \citep{wang2023open, jiang2024survey}.

\begin{table*}[!h]
    \centering
    \scriptsize
    \caption{Overview of Representative Coding Datasets for Training and Evaluation.}
    \label{tab:codeset}
    \renewcommand{\arraystretch}{1.2}
    \definecolor{lightgray}{RGB}{245,245,245}
    \resizebox{\textwidth}{!}{%
    \begin{tabular}{l|p{3cm}|p{2.5cm}|p{4.5cm}|p{3.5cm}|p{2cm}}
    \toprule
    \textbf{Dataset} & 
    \textbf{\begin{tabular}[c]{@{}l@{}}Release Date/\\Scale\end{tabular}} & 
    \textbf{\begin{tabular}[c]{@{}l@{}}Languages/\\Task Types\end{tabular}} & 
    \textbf{\begin{tabular}[c]{@{}l@{}}Data Collection\\Method\end{tabular}} & 
    \textbf{\begin{tabular}[c]{@{}l@{}}Characteristics/\\Purpose\end{tabular}} & 
    \textbf{\begin{tabular}[c]{@{}l@{}}Level/\\Attributes\end{tabular}} \\
    \midrule
    \rowcolor{white}SWE-Gym~\citep{pan2024training} & 2025; \textasciitilde 2.4k tasks, including 230 lightweight instances & Python; project-level bug fixing and editing tasks & Collected from real pull requests in 11 open-source Python projects with executable environments and unit tests & Contains complete code context, executable test environments, and unit tests; suitable for training RL coding agents & Project-level, with repository context, directly executable \\
    \rowcolor{lightgray}Codeforces-CoTs~\citep{penedo2025codeforces} & 2025; \textasciitilde 10k Codeforces problems, \textasciitilde 100k total samples, \textasciitilde 9.69GB & C++ and Python competitive programming problems with chain-of-thought & Collected 10k+ problems from Codeforces, using DeepSeek R1 to generate up to five reasoning trajectories per problem & Problems accompanied by model-generated code and reasoning trajectories; \textasciitilde 84\% Python solutions pass public tests & Problem-level; includes multiple subsets (solutions, solutions\_w\_editorials, etc.) \\
    \rowcolor{white}SWE-Fixer~\citep{xie2025swe} & 2025; 110k high-quality samples filtered from 2.3k repositories and 331k instances & Multi-language, including code retrieval and editing & Utilized GitHub events to crawl issue-PR pairs, limited to three files or fewer, removed parsing errors, generated chain-of-thought & Each instance contains issue description, code context, and patches; suitable for supervised and RL training on retrieval/editing tasks & File-level and project-level; includes chain-of-thought \\
    \rowcolor{lightgray}KodCode~\citep{xu2025kodcode} & 2024; 447k problem-solution-test triples & Synthetic algorithmic/coding exercises & Generated through multi-stage pipeline: problem synthesis, solution generation, unit test creation, and post-processing & All problems equipped with verifiable unit tests, covering various difficulty levels; suitable for SFT and RL training & Function-level; fully synthetic, emphasizes test executability \\
    \rowcolor{white}Code-R1~\citep{code-r1} & 2025; 12k RL samples (2k LeetCode + 10k filtered from TACO) & Programming problems for RL training & Selected from LeetCode and TACO verified instances with reliable tests, providing prompts and unit tests & Emphasizes sandbox execution for reward verification, avoids test noise; directly usable for GRPO optimization & Function-level; each contains prompt and tests \\
    \rowcolor{lightgray}Z1-Code~\citep{yu2025z1} & 2024; 107k coding problems & Multi-language coding problems with both short and long reasoning trajectories & Filtered from public problem sets and synthetic problems, annotated with short-chain and long-chain reasoning & Provides two reasoning depths to improve model generalization and avoid over-thinking & Problem-level; includes solutions and reasoning \\
    \rowcolor{white}rStar-Coder~\citep{liu2025rstar} & 2024; 418k competitive problems and 580k long-reasoning solutions & Multi-language competitive programming problems & Aggregated existing competitive problems and synthesized new ones, generated input-output tests and long-reasoning solutions & Each problem has verification tests and in-depth reasoning solutions; suitable for training interpretable coding models & Problem-level; includes tests and reasoning \\
    \rowcolor{lightgray}LeetCodeDataset~\citep{xia2025leetcodedatasettemporaldatasetrobust} & 2025; 2,613 training problems and 256 test problems & Python LeetCode algorithmic problems & Wrapped LeetCode problems into data structures containing imports, entry\_point, tests, completion, examples, and metadata & Each problem provides complete unit tests and target code; suitable for model training and evaluation & Function-level; includes Python test scripts \\
    \rowcolor{white}OpenCodeReasoning~\citep{ahmad2025opencodereasoning} & 2024; 735,255 samples covering 28,319 programming problems & Multi-language programming problems with reasoning & Integrated data from 11 programming platforms and generated solutions using NVIDIA R1 & Large-scale high-quality programming reasoning data; supports training models with reasoning capabilities & Problem-level; includes problem text, solutions, and reasoning \\
    \rowcolor{lightgray}DeepCoder~\citep{deepcoder2025} & 2025; 24k verified coding problems, each with $\geq$5 test cases & Competitive programming problems & Merged TACO verified, Prime Intellect synthetic problems, LiveCodeBench, etc., filtered to 24k problems with complete tests & Provides fully open high-quality data for RL training; reward uses binary model (pass/fail) & Problem-level; includes complete test sets \\
    \rowcolor{white}\makecell[l]{SWE-Bench Variants\\\citep{jimenez2023swe,yang2024swebench_multimodal,yang2025swesmith}} & Since 2023; Full: 2,294, Lite: 300, Verified: 500, Multimodal: 617, Multilingual: 300 & Primarily Python, also multi-language versions; project-level issue resolution & Collected issues and PRs from GitHub, recording base commits, patches, and test patches; some instances expert-verified & Instances include problem descriptions, base commits, patches, tests, and metadata; Multilingual version extends to 9 languages & Project-level; used for evaluating bug-fixing capabilities of coding agents \\
    \rowcolor{lightgray}Multi-SWE-Bench~\citep{zan2025multi} & 2025; 1,632 high-quality instances & Seven languages: Java, TypeScript, JavaScript, Go, Rust, C, C++ & Five-stage process: repository selection, PR crawling, environment determination, PR filtering, and human validation & Fills the Python-only gap in SWE-Bench; cross-language evaluation for coding agents & Project-level; used for evaluation \\
    \rowcolor{white}MLE-bench~\citep{chan2024mle} & 2024; 75 Kaggle competitions, 3.3TB dataset & Multi-language; ML engineering tasks including model training, dataset preparation, and experiments & Curated from Kaggle competitions covering various domains (NLP, CV, signal processing), with human baselines from public leaderboards & Challenging tasks testing real-world ML engineering skills; includes competitions like COVID-19 vaccine prediction and ancient scroll deciphering & Competition-level; open-ended and difficult tasks for evaluating autonomous ML engineering \\
    \rowcolor{lightgray}PaperBench~\citep{starace2025paperbench} & 2025; 20 ICML 2024 papers, 8,316 individually gradable tasks & Multi-language; AI research replication tasks & Selected 20 Spotlight and Oral papers from ICML 2024, with rubrics co-developed with original paper authors & Evaluates agents' ability to replicate cutting-edge AI research from scratch, including understanding contributions, developing codebases, and executing experiments; uses LLM-based judge for automated grading & Research-level; hierarchical rubrics decompose tasks into fine-grained sub-tasks \\
    \rowcolor{white}Terminal-Bench~\citep{tbench2025} & 2025; 80 tasks & Multi-language; terminal environment tasks & Hand-crafted and human-verified tasks, each with dedicated Docker environment, human-verified solutions, and test cases & Evaluates agents on complete, end-to-end tasks in terminal environments including scientific workflows, network configuration, data analysis, API calls, and cybersecurity; emphasizes real-world complexity & System-level; covers diverse terminal behaviors, directly executable in sandboxed environments \\
    \rowcolor{lightgray}SWE-RL~\citep{wei2025swe} & 2024; aggregated 24M PR instances and cloned 4.6M repositories & Multi-language bug fixing tasks & Crawled GitHub events from GHArchive (2015-2024), aggregated PRs and used heuristics to select bug-fixing seeds; each instance contains problem description, code context, and patches & Uses difflib as reward function; used for training large models like Llama3-SWE-RL & Project-level; massive scale \\
     \bottomrule
    \end{tabular}%
    }
    \vspace{0.5em}

    \raggedright

    \renewcommand{\arraystretch}{1.0}
\end{table*}

\subsubsection{Continual Pre-training Strategies}

Continual pre-training (CPT) enables models initially pre-trained on broad general corpora to undergo additional rounds of training on new, often domain-specific datasets, representing an important paradigm \citep{gao2024enhancing}. This approach enables models to acquire substantial new knowledge in specialized domains such as coding, mathematics, or scientific reasoning while preserving and building upon existing foundational capabilities \citep{gao2024enhancing}, with research demonstrating that LLMs possess inherent abilities to effectively accumulate and retain knowledge across sequential training tasks \citep{singh2025beyond, brown2020language}.

The code domain is one of the most successful application areas for continual pre-training strategies. CodeLlama demonstrates this success by leveraging CPT on the general-purpose LLaMA2 foundation model to achieve state-of-the-art code generation performance \citep{ibrahim2024simple, deepseekai2024deepseek}, while DeepSeek-Coder-V2 shows more advanced CPT capabilities by continuing training from DeepSeek-V2 intermediate checkpoints with an additional 6 trillion tokens, resulting in substantial enhancements to both coding abilities and mathematical reasoning capabilities \citep{ibrahim2024simple, deepseekai2024deepseek}. Critical to effective CPT are sophisticated data mixing strategies that carefully balance different data types and domains, with Qwen2.5-Coder's systematic experimentation determining optimal 7:2:1 mixing ratios for code, text, and mathematics data, resulting in over 20\%~average performance improvement compared to code-only training approaches \citep{ibrahim2024simple}.

However, preventing catastrophic forgetting of previously learned knowledge remains a central challenge in CPT \citep{kirkpatrick2017overcoming, mccloskey1989catastrophic, van2024continual}. This is addressed through strategies such as DeepSeek's approach of replaying 30\% of original pre-training data during continual training phases to maintain performance on previously mastered tasks \citep{ibrahim2024simple}, with replay-based methods being a well-established approach for mitigating catastrophic forgetting \citep{lesort2020continual}.

Critically, data selection for continual pre-training leverages both distributed corpus features and linguistic characteristics to identify the most valuable training examples \citep{xie2023efficient, ruder2017learning, tsvetkov2016learning}, with empirical research demonstrating that CPT on appropriately selected unlabeled task data and augmented unlabeled data sampled from in-domain corpora can significantly enhance end-task performance \citep{xie2023efficient, gururangan2020don}. Critical CPT challenges that remain active research areas include the "stability gap" phenomenon where models experience temporary but significant performance drops at the onset of new domain data training before eventually recovering and surpassing previous performance levels \citep{guo2024efficient}, the persistent issue of catastrophic forgetting of previous knowledge though research indicates self-supervised CPT approaches tend to outperform supervised protocols in knowledge retention \citep{singh2025beyond, brown2020language, mirzadeh2021wide}, and the fundamental difficulty of scaling CPT methods to scenarios with small limited-token corpora where the restricted diversity of training examples significantly hinders effective knowledge acquisition and generalization \citep{ovadia2025knowledge, kandpal2022large}.

\subsection{Post-training Techniques} \label{sec:llm_coding_post}
\subsubsection{Supervised Fine-tuning}

Large language model training follows well-established two-stage paradigms, with an initial pre-training phase on massive diverse corpora to acquire broad linguistic and world knowledge, followed by post-training that adapts models to excel at specific tasks and domains \citep{wei2025vericoder}. Supervised fine-tuning (SFT) is a critical technique that enhances LLM capabilities while aligning model behaviors with human expectations and requirements \citep{shengyu2023instruction, brown2020language, chung2022scaling}. The SFT process involves adjusting pre-trained model parameters in a supervised manner by utilizing carefully curated datasets containing high-quality labeled examples, enabling models to maintain their broad foundational knowledge while simultaneously acquiring highly targeted specialization for specific applications \citep{wu2024sparkra, wei2021finetuned, ouyang2022training, liu2024m2rc}.

Instruction tuning represents a particularly impactful form of supervised fine-tuning that enhances model capabilities by explicitly training models to understand and accurately follow diverse human instructions across varied task formats \citep{jiang2024survey, wei2021finetuned}, with research demonstrating substantial improvements in zero-shot performance on previously unseen task types \citep{wei2021finetuned, chung2022scaling}.

Code generation is one of the most prominent and successful applications of supervised fine-tuning techniques, with notable implementations including CodeGen and StarCoder which employ a two-stage approach of initial pre-training on diverse multilingual code corpora followed by targeted fine-tuning on carefully selected monolingual data to enhance language-specific code generation quality \citep{shen2023pangu}, while systems like WizardCoder demonstrate alternative approaches by fine-tuning on specialized instruction corpora systematically derived from more capable teacher models \citep{shen2023pangu, ouyang2022training}. The extremely large parameter counts characteristic of modern language models make traditional approaches of updating all model parameters during fine-tuning computationally impractical and prohibitively expensive \citep{hu2022lora, houlsby2019parameter}, driving rapid development of Parameter-Efficient Fine-Tuning (PEFT) methodologies that achieve comparable performance with reduced computational requirements \citep{yuan2023evaluating}. Adapter Tuning represents one influential PEFT approach by introducing small trainable adapter modules that add only a minimal number of new trainable parameters for each specific task while strategically keeping the vast majority of original pre-trained parameters completely fixed \citep{yuan2023evaluating, houlsby2019parameter}, while Low-Rank Adaptation (LoRA) achieves even greater parameter efficiency by imposing carefully designed low-rank constraints on weight updates, enabling effective fine-tuning with orders of magnitude fewer trainable parameters \citep{yuan2023evaluating}.

The critical importance of data quality over mere quantity has become a central guiding principle in supervised fine-tuning research and practice \citep{nakkiran2019deep, shumailov2024models, zhou2023lima}, with sophisticated data selection and curation strategies proving essential for optimal results. The AlpaGasus approach uses this method by systematically using ChatGPT to automatically identify and filter out low-quality training examples, achieving significant performance improvements using only 9,000 carefully selected high-quality samples compared to much larger but lower-quality datasets \citep{chen2023alpagasus}. Interestingly, recent empirical research has revealed that supervised fine-tuning on high-quality demonstrations can in many cases result in superior performance compared to more complex preference fine-tuning approaches, particularly evident in challenging coding tasks where direct supervision appears especially effective \citep{xiao2024how}.

\subsubsection{Reinforcement Learning}

Modern post-training approaches encompass two primary paradigms: supervised fine-tuning based on demonstration data, and preference fine-tuning methods utilizing techniques like RLHF and DPO \citep{xiao2024how, iacovides2025findpo, ouyang2022training, rafailov2023direct, xiong2025rag}.

Among these paradigms, reinforcement learning has become a pivotal technique for post-training large language models, representing fundamental paradigm shifts away from traditional supervised fine-tuning approaches toward more dynamic optimization strategies \citep{ouyang2022training, stiennon2020learning, wang2024enhancing, wang2025dump, dou2025improving}. While pre-training establishes robust foundational linguistic knowledge and pattern recognition capabilities, subsequent post-training refinement through sophisticated RL techniques has proven essential for substantially enhancing models' complex reasoning abilities and adaptability to challenging real-world tasks \citep{ferrag2025reasoning}. The foundational framework for RL-based post-training was firmly established by RLHF systems that align model behaviors using carefully constructed reward signals derived from human preference judgments \citep{wang2025dump, ouyang2022training, xi2025agentgym}, with recent breakthrough systems like OpenAI's o-series models and DeepSeek-R1 demonstrating that models trained with RL post-training can significantly outperform comparable models without RL training on demanding reasoning and coding benchmarks \citep{wang2025dump, xu2025kdrl}.

The reinforcement learning algorithmic landscape for language model training centers around several key complementary approaches, each with distinct advantages and trade-offs. PPO remains the most widely adopted baseline approach due to its carefully designed clipped objective functions that effectively stabilize the inherently unstable RL training process \citep{wang2025dump, ouyang2022training, rodriguez2025rendering, stiennon2020learning}, while DPO represents a major algorithmic simplification by cleverly replacing complex RL rollout procedures with more straightforward classification-style loss functions that directly optimize preference rankings \citep{wang2025dump, rafailov2023direct, gao2025one}. Group Relative Policy Optimization (GRPO) emerged to address specific limitations in DPO by incorporating group-wise comparative feedback that enables more nuanced preference learning \citep{wang2025dump, rodriguez2025rendering}. Significant paradigm shifts have emerged with approaches like Reinforcement Learning with Verifiable Rewards (RLVR) that strategically replace learned reward models with deterministic task-specific rule-based reward functions, proving particularly effective for domains like coding and mathematics where answer correctness can be objectively verified \citep{taveekitworachai2025prior}.

Reinforcement learning has become a transformative strategy for code generation applications, with significant empirical advancements demonstrated across multiple diverse programming domains and task types \citep{wang2024enhancing}. Foundational approaches in this area are exemplified by CodeRL, which employs sophisticated actor-critic framework architectures where critic networks learn to accurately predict functional correctness of generated code, providing learned reward signals that guide actor networks during the training process \citep{rodriguez2025rendering, le2022coderl}. Building on these foundations, PPOCoder demonstrates effective integration of Proximal Policy Optimization with direct execution-based feedback by utilizing actual compiler results and test case outcomes as immediate reward signals, enabling models to learn from real program behavior \citep{rodriguez2025rendering}, while more recent developments have introduced sophisticated multi-stage training strategies exemplified by StepCoder which implements carefully designed curricula progressing through increasingly complex code completion subtasks \citep{rodriguez2025rendering}. Applications of RL-based code generation have extended to highly specialized technical domains, with systems like VeriSeek demonstrating RL's effectiveness for generating correct hardware description language code in challenging Verilog generation tasks \citep{wang2024large}.

RL-based post-training faces numerous fundamental challenges that remain active areas of research. Training stability issues include catastrophic forgetting of previously learned capabilities and reward hacking behaviors where models exploit reward function specifications rather than learning intended behaviors \citep{ferrag2025reasoning}. Scale-dependent effectiveness presents another significant limitation, with substantial benefits from RL training observed primarily in large models ranging from 8 billion to 670 billion parameters \citep{le2025reasoning}, while RL techniques remain considerably more challenging to apply successfully to smaller models with one billion or fewer parameters where training instability and sample efficiency become critical bottlenecks. Additionally, optimization dilemmas persist in the form of exploration-exploitation tradeoffs, with RL training often suffering from severe sample inefficiency particularly when initial policy distributions struggle to adequately explore regions of the action space containing high-reward trajectories \citep{xu2025kdrl}, and fundamental mismatches between training objectives and evaluation metrics, with standard RL formulations focused on maximizing expected cumulative reward while practical evaluation of code generation systems typically relies on risk-seeking metrics like Pass@k that prioritize generating at least one correct solution among multiple attempts \citep{xi2025agentgym, qi2024webrl, xia2024inverse}.
\begin{figure}[t]
    \centering
    \includegraphics[width=\linewidth]{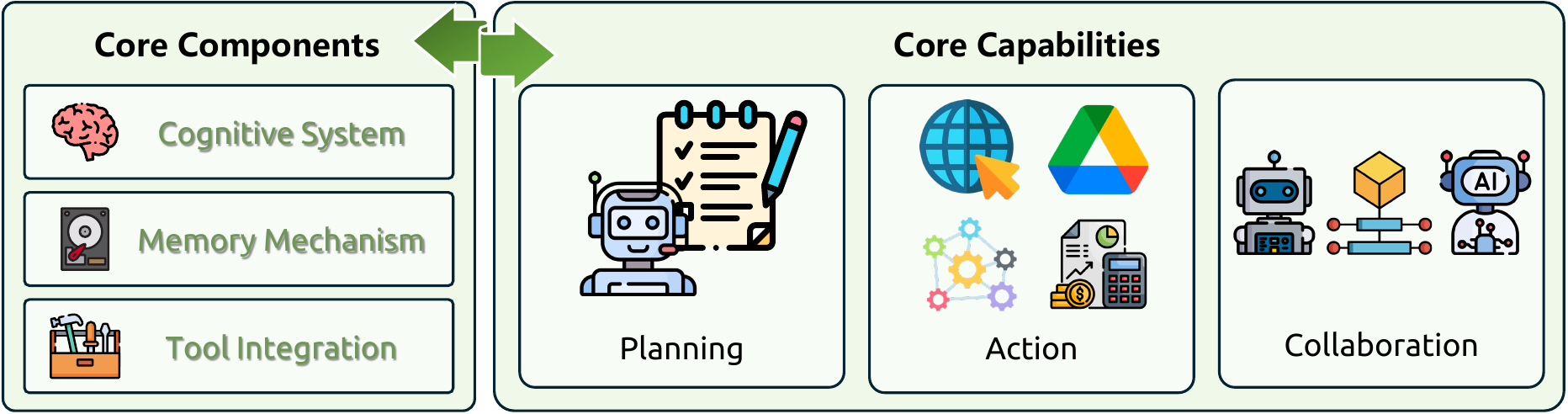}
    \caption{Coding Agent Architecture: Overview of core components comprising Cognitive System, Memory Mechanism, and Tool Integration, with fundamental capabilities including Planning, Action execution, and multi-agent Collaboration for coordinated software development tasks.}
    \label{fig:his}
\end{figure}

\section{LLM-based Coding Agent}
\label{sec:coding_agent}

\subsection{Decomposition and Planning Capability} \label{sec:coding_agent_plan}

\subsubsection{Task Decomposition Strategies}
Task decomposition systematically breaks complex problems into manageable subtasks, proven essential from traditional planning to modern LLM applications, enabling coding agents to tackle sophisticated challenges while improving reliability, interpretability, and multi-agent collaboration \citep{stiennon2020learning, huang2024levels, wei2022chain, hu2024agentgen}. Strategies evolved into distinct paradigms including factored cognition for simultaneous processing, process supervision for sequential dependencies, Chain-of-Thought for step-by-step reasoning, and extensions like Tree-of-Thought (ToT) and Graph-of-Thought for multi-path reasoning \citep{wei2022chain, yao2022react, cao2025large, lu2025decompose, shinn2023reflexion, shen2023hugginggpt}. Concrete techniques span prompting-based CoT using few-shot or zero-shot instructions \citep{wei2022chain, kojima2022large, zhao2023depth}, Auto-CoT generating samples through clustering \citep{zhang2022automatic}, Plan-and-Solve devising task-dividing plans \citep{zhao2023depth}, tree structures like ToT generating branching explorations \citep{yao2023tree}, and HyperTree Planning enabling divide-and-conquer with Monte Carlo integration \citep{gui2025hypertree}.

CoT emerges as most popular code generation strategy achieving higher correctness \citep{wei2022chain, yao2022react}, while dynamic strategies like CodePlan employ adaptive algorithms \citep{wen2024codeplan, hu2024agentgen}, and multi-path approaches simulate generation paths with Monte Carlo optimization, though LLMs potentially hallucinate incoherent plans despite proper instructions \citep{tian2024sqlucid, jayagopal2022exploring, patel2022question}.

\subsubsection{Plan Formulation Methods}

LLMs' training on natural language and code unlocks reasoning producing structured steps connecting to execution environments, with planning crucial for agents determining action sequences from initial states \citep{hu2024agentgen}. Approaches categorize into External Module Augmented Methods integrating planners and memory, Finetuning-based Methods enhancing abilities through trajectory data, and Searching-based Methods identifying optimal solutions \citep{cao2025large, zhang2025multiple}, while three paradigms emerged: LLM-as-Planner employing inherent reasoning, LLM-as-Facilitator translating to Planning Domain Definition Language (PDDL) \citep{molinari2025towards, helmert2006fast}, and Multi-Agent Planning coordinating multiple agents \citep{hu2024agentgen, hong2024metagpt}, with task decomposition employing Single-Path following linear trajectories \citep{wei2022chain}, Tree-Based exploring multiple trajectories \citep{yao2023tree, besta2023graph}, and Hierarchical Planning at different abstraction levels.

Planning with feedback enables adaptation through ReAct interleaving Thought, Action, and Observation \citep{pitkranta2025hada, yao2022react}, though incremental approaches risk local optima \citep{chen2025enhancing}, while Memory-Enhanced Planning addresses limitations through Reflexion augmenting agents with Evaluator, Self-reflection, and memory buffers enabling continual improvement \citep{luo2025large, wei2022chain, packer2023memgpt}. Multi-agent systems coordinate specialized agents distributing planning across components \citep{molinari2025towards, ouyang2024long}, with Integration combining structured PDDL with LLMs where LLMs allocate sub-tasks while planners like Fast Downward generate plans \citep{zhang2024lamma, helmert2006fast, singh2024twostep, sun2023pearl}. Challenges persist including Goal Decomposition struggling with actionable steps \citep{erdogan2025plan}, Long-Horizon Task Management losing coherent strategy \citep{erdogan2025plan, kienle2025learning}, and Training Data Scarcity requiring synthetic augmentation \citep{erdogan2025plan}.

\subsection{Memory Mechanism} \label{sec:coding_agent_mem}

\subsubsection{Overview and Fundamentals}

Extending beyond fixed context windows, memory mechanisms emerge as functional phenomena in LLMs demonstrating primacy and recency effects despite lacking explicit subsystems \citep{janik2023aspects, vaswani2017attention, shan2025cognitive}. Contemporary approaches draw parallels to human cognitive architecture, mapping context and parameters to short-term and long-term memory and transforming contextual knowledge into parameter updates \citep{shan2025cognitive, parthasarathy2024ultimate}.

Following this framework, short-term memory constitutes temporary storage with limited capacity, corresponding to context window information accessible during inference \citep{ren2024brain, lu2023towards, xie2024travelplanner, brown2020language, zhang2023igniting}, while long-term memory stores information over days to years, classified into declarative explicit and non-declarative implicit memory \citep{he2024human}. From an implementation perspective, these memory types classify into parametric memory stored in model weights and non-parametric memory in external systems \citep{he2024human, wu2025from, han2024llm, zhang2025multiple}.

\begin{table*}[!h]
    \centering
    \scriptsize
    \caption{Capabilities of Coding Agents Based on Large Language Models.}
    \label{tab:coding-agents}
    \renewcommand{\arraystretch}{1.2}
    \definecolor{lightgray}{RGB}{245,245,245}
    \resizebox{\textwidth}{!}{%
    \begin{tabular}{l|cccccccc}
    \toprule
    \multirow{2}{*}{\textbf{Coding Agent}} & \multicolumn{8}{c}{\textbf{Capability}} \\
    \cmidrule(lr){2-9}
    & \textbf{\begin{tabular}[c]{@{}c@{}}Code\\Search\end{tabular}} & \textbf{\begin{tabular}[c]{@{}c@{}}File\\Ops\end{tabular}} & \textbf{Shell} & \textbf{\begin{tabular}[c]{@{}c@{}}Web\\Search\end{tabular}} & \textbf{Testing} & \textbf{MCP} & \textbf{Multimodal} & \textbf{Context} \\
    \midrule
    \rowcolor{white}
    CodeAgent~\cite{zhang2024codeagent} & \checkmark & \checkmark &  & \checkmark & \checkmark &  &  & \checkmark \\
    \rowcolor{lightgray}
    MapCoder~\cite{zhao2025mage} & \checkmark &  &  &  & \checkmark &  &  & \checkmark \\
    \rowcolor{white}
    ChatDev~\cite{qian2023chatdev} &  &  &  &  & \checkmark &  &  & \checkmark \\
    \rowcolor{lightgray}
    CodeAct~\cite{wang2024executable} &  &  & \checkmark &  & \checkmark &  &  &  \\
    \rowcolor{white}
    SWE-Search~\cite{antoniades2024swe} & \checkmark & \checkmark &  &  & \checkmark &  &  & \checkmark \\
    \rowcolor{lightgray}
    OpenHands~\cite{wang2024openhands} &  & \checkmark & \checkmark & \checkmark & \checkmark &  &  & \checkmark \\
    \rowcolor{white}
    OpenHands-Versa~\cite{ho2025verilogcoder} &  & \checkmark & \checkmark & \checkmark & \checkmark &  & \checkmark & \checkmark \\
    \rowcolor{lightgray}
    MetaGPT~\cite{hong2024metagpt} &  & \checkmark & \checkmark & \checkmark & \checkmark &  &  & \checkmark \\
    \rowcolor{white}
    SWE-agent~\cite{yang2024swe} & \checkmark & \checkmark & \checkmark &  & \checkmark &  &  & \checkmark \\
    \rowcolor{lightgray}
    AutoSafeCoder~\cite{witt2025open} &  &  &  &  & \checkmark &  &  & \checkmark \\
    \rowcolor{white}
    AutoCodeRover~\cite{zhang2024autocoderover} & \checkmark &  &  &  & \checkmark &  &  & \checkmark \\
    \rowcolor{lightgray}
    Lita~\cite{madison2024scalable} & \checkmark & \checkmark & \checkmark &  & \checkmark &  &  & \checkmark \\
    \rowcolor{white}
    SWE-dev~\cite{wang2025swe} &  &  &  &  & \checkmark &  &  &  \\
    \rowcolor{lightgray}
    RepoForge~\cite{chen2025repoforge} &  & \checkmark & \checkmark &  & \checkmark &  &  & \checkmark \\
    \rowcolor{white}
    LessonL~\cite{liu2025lessons} &  &  &  &  & \checkmark &  &  & \checkmark \\
    \rowcolor{lightgray}
    AdaCoder~\cite{robeyns2025self} &  &  &  &  & \checkmark &  &  &  \\
    \rowcolor{white}
    Code2MCP~\cite{ouyang2025transforming} &  &  &  &  & \checkmark & \checkmark &  & \checkmark \\
    \rowcolor{lightgray}
    ScreenCoder~\cite{jiang2025screencoder} &  &  &  &  &  &  & \checkmark & \checkmark \\
    \rowcolor{white}
    SimuGen~\cite{ren2025simugen} &  &  &  &  & \checkmark &  & \checkmark &  \\
    \rowcolor{lightgray}
    SoA~\cite{ishibashi2024self} &  &  &  &  &  &  &  & \checkmark \\
    \rowcolor{white}
    SICA~\cite{singh2024twostep} &  & \checkmark & \checkmark &  & \checkmark &  &  & \checkmark \\
    \rowcolor{lightgray}
    RGD~\cite{jin2024rgd} &  &  &  &  & \checkmark &  &  & \checkmark \\
    \rowcolor{white}
    Guided Code Generation~\cite{huang2023agentcoder} &  &  &  &  & \checkmark &  &  & \checkmark \\
    \rowcolor{lightgray}
    AgentCoder~\cite{huang2023agentcoder} &  &  & \checkmark &  & \checkmark &  &  & \checkmark \\
    \rowcolor{white}
    AppAgent~\cite{zhang2025appagent} &  &  &  &  &  &  & \checkmark &  \\
    \rowcolor{lightgray}
    Cursor IDE~\cite{cursor-ide} & \checkmark & \checkmark & \checkmark & \checkmark & \checkmark & \checkmark & \checkmark & \checkmark \\
    \rowcolor{white}
    Claude Code~\cite{claude-code} & \checkmark & \checkmark & \checkmark & \checkmark & \checkmark & \checkmark & \checkmark & \checkmark \\
    \rowcolor{lightgray}
    Gemini Code CLI~\cite{gemini-code-cli}& \checkmark & \checkmark & \checkmark & \checkmark & \checkmark & \checkmark & \checkmark & \checkmark \\
    \rowcolor{white}
    Qwen Coder~\cite{qwen-coder} & \checkmark & \checkmark & \checkmark & \checkmark & \checkmark & \checkmark & \checkmark & \checkmark \\
    \rowcolor{lightgray}
    Codex~\cite{codex} & \checkmark & \checkmark & \checkmark & \checkmark & \checkmark & \checkmark & \checkmark & \checkmark \\
    \bottomrule
    \end{tabular}%
    }


\end{table*}

\subsubsection{Memory Operations and Management}

Accordingly, memory operations enable coding agents to function beyond static context windows, maintaining persistent knowledge about codebases, patterns, and debugging histories \citep{packer2023memgpt, xu2025mem, zhang2025survey}. These operations are managed through six fundamental procedures: consolidation, indexing, updating, forgetting, retrieval, and compression \citep{zhang2025memengine, xiong2025memory, du2025rethinking, feng2024tasl, zhong2023memorybank, shan2025cognitive}.

Specifically, the foundation rests on memory reading retrieving relevant information \citep{park2023generative}, memory writing transforming observations into structured contents, and memory reflection reorganizing knowledge \citep{huang2022language, zhang2025survey, zhang2025multiple}. Comprehensive systems implement specialized functions including consolidation, indexing, and updating employing locate-and-edit techniques \citep{hou2024my, modarressi2023ret, xia2024agentless, du2025rethinking, bae2022keep, zhong2023memorybank}, with optimization extracting meta-insights and selective management yielding 10\% gains \citep{xiong2025memory, chhikara2025mem0}.

\subsubsection{Memory Architecture in Coding Agents}

In practice, short-term memory plays critical role in hierarchical architectures, enabling immediate feedback integration demonstrated by FALCON's two-level structure \citep{zhan2025coderagent, packer2023memgpt}, while dynamic management with MemTool achieves 90-94\% efficiency \citep{lumer2025memtool}.

In contrast, long-term memory implementations utilize external memory-based methods with physical storage maintaining historical information. Vector databases emerged as prominent non-parametric solution achieving long-term effects and allowing domain-specific knowledge access \citep{lewis2020retrieval, kynoch2023recallm, wu2022memorizing}, though struggling with belief updating challenges. To address these limitations, Memory Bank stores memory with temporal timestamps and implements Ebbinghaus-inspired forgetting curves \citep{zhong2023memorybank, xing2025structured, kynoch2023recallm, zhang2025multiple}. These implementations categorize into three architectural patterns: Long-Context Agents with First-In-First-Out (FIFO) eviction, RAG Agents using external pools, and Agentic Memory Agents employing iterative reasoning \citep{hu2025evaluating, shan2025cognitive}. Furthermore, contemporary architectures adopt dual-component systems mirroring human cognition through frameworks like RAISE \citep{liu2024llm} and InfiniteICL \citep{cao2025infiniteicl}, transforming temporary knowledge into permanent updates. Additionally, specialized mechanisms demonstrate distinct optimization strategies. Think-in-Memory eliminates redundant reasoning by caching intermediate results \citep{pawar2024what}, whereas MemGPT leverages virtual context management to extend effective context windows through intelligent memory paging \citep{packer2023memgpt}, thus addressing belief updating challenges inherent in static vector database approaches.

\subsection{Action Execution} \label{sec:coding_agent_act}


\subsubsection{Tool Invocation} 

Tool integration with LLMs fundamentally transformed AI-powered development \citep{schick2023toolformer, yao2022react}, with Tool-Augmented Models extending functionalities by integrating external tools \citep{chen2023large, shen2024small, yao2022react}, producing AI agentic programming where LLM-based agents autonomously plan, execute, and interact with compilers, debuggers, and version control reshaping development practices. Early task-specific systems suffered limited scalability \citep{fei2025mcp}, while ReAct established observation-action-thought pattern becoming foundation for LangChain and AutoGen \citep{fei2025mcp, yao2022react}, with modern frameworks adopting standardized paradigms decomposing capabilities into specialized components \citep{shen2024llm} and enabling automatic transformation of documentation into callable functions \citep{hou2025model, shi2024tool}.

Tool invocation evolved beyond traditional JSON-based approaches \citep{yao2022react, qin2023tool}, with executable code-based frameworks \citep{wang2024executable, islam2024mapcoder} consolidating actions into unified Python space enabling dynamic revision and composition, demonstrating substantial improvements over predefined constraints. Systems like ToolCoder reduced hallucination errors \citep{dong2025survey, schick2023toolformer, hong2024metagpt}, while agents integrating specialized tools \citep{zhang2024codeagent} demonstrated significant improvements outperforming commercial products like GitHub Copilot \citep{dong2025survey}. Protocol standardization through Model Context Protocol established schema-driven interfaces, with advanced discovery mechanisms \citep{fei2025mcp, lumer2025scalemcp} enabling dynamic selection through iterative invocation and document-weighted averaging addressing scalability across large-scale repositories. Enterprise-focused frameworks \citep{agarwal2025automated, mastouri2025making, ni2025doc2agent, hu2024agentgen} automated pipelines transforming API specifications into executable Python functions, substantially enhancing accuracy while eliminating scaffolding efforts, with reinforcement learning integration \citep{singh2025agentic, jin2025reveal} unifying tool usage with reasoning through policy optimization.

\begin{table*}[!h!]
    \centering
    \scriptsize
    \caption{Overview of Commercial AI-Assisted Development Softwares. 
    }
    \label{tab:ai-dev-tools}
    \renewcommand{\arraystretch}{1.2}
    \definecolor{lightgray}{RGB}{245,245,245}
    \resizebox{\textwidth}{!}{%
    \begin{tabular}{>{\raggedright\arraybackslash}p{2cm}|>{\raggedright\arraybackslash}p{2.2cm}|p{3.5cm}|p{4cm}|p{3.5cm}}
    \toprule
    \textbf{Type} & \textbf{Product} & 
    \textbf{Core Functionality} & 
    \textbf{Technical Features} & 
    \textbf{Target Scenarios} \\
    \midrule
    \multirow{5}{2cm}{\centering\raisebox{-85pt}{\textbf{Embedded}}} 
    & \cellcolor{lightgray}Cursor~\citep{cursor-ide} & \cellcolor{lightgray}AI-native code editor with multi-file editing, intelligent refactoring, and conversational AI & \cellcolor{lightgray}VS Code fork with deep AI integration, composer agent mode, supports Claude/GPT-4/Gemini models, codebase-wide context awareness & \cellcolor{lightgray}Complex projects requiring intelligent refactoring, full-stack development, professional developers \\
    \cmidrule{2-5}
    & \cellcolor{white}Copilot~\citep{copilot} & \cellcolor{white}Real-time code completion, generation, and explanation across multiple IDEs & \cellcolor{white}Trained on GitHub repositories, supports VS Code/JetBrains/Neovim, inline suggestions, chat interface & \cellcolor{white}Daily coding assistance, autocomplete-focused workflows, GitHub-integrated development \\
    \cmidrule{2-5}
    & \cellcolor{lightgray}Windsurf~\citep{windsurf} & \cellcolor{lightgray}Agentic IDE with autonomous coding capabilities and deep contextual understanding & \cellcolor{lightgray}Cascade Flow system, auto-executes terminal commands, supports multiple AI models, built on VS Code & \cellcolor{lightgray}Large codebases, multi-file reasoning, team collaboration with advanced workflow automation \\
    \cmidrule{2-5}
    & \cellcolor{lightgray}Tabnine~\citep{tabnine} & \cellcolor{lightgray}Enterprise-grade AI code completion with privacy-first approach & \cellcolor{lightgray}Local/on-premise deployment options, trained only on permissive licenses, custom model training on proprietary code & \cellcolor{lightgray}Enterprise environments, regulated industries (finance, healthcare), privacy-sensitive projects \\
    \cmidrule{2-5}
    & \cellcolor{white}CodeWhisperer~\citep{CodeWhisperer} & \cellcolor{white}AWS-optimized code generation and security scanning & \cellcolor{white}Integrated with AWS services, reference tracking for license compliance, security vulnerability scanning & \cellcolor{white}AWS cloud development, serverless applications, cloud-native architecture \\
    \midrule
    \multirow{3}{2cm}{\centering\raisebox{-42pt}{\textbf{End-to-End}}} 
    & \cellcolor{lightgray}v0~\citep{v0} & \cellcolor{lightgray}Natural language to React/Next.js UI generation with deployment & \cellcolor{lightgray}Generates production-ready React + Tailwind code, integrated security checks, one-click Vercel deployment & \cellcolor{lightgray}Frontend prototyping, UI/UX design-to-code, rapid MVP development \\
    \cmidrule{2-5}
    & \cellcolor{white}Replit~\citep{replit} & \cellcolor{white}Cloud-based autonomous app builder from natural language descriptions & \cellcolor{white}Complete environment setup, integrated databases and authentication, multi-language support, browser-based testing & \cellcolor{white}Non-technical users, rapid prototyping, educational projects, small business tools \\
    \cmidrule{2-5}
    & \cellcolor{lightgray}Bolt.new~\citep{bolt} & \cellcolor{lightgray}Browser-native full-stack development with instant preview and deployment & \cellcolor{lightgray}WebContainers technology, runs Node.js in browser, supports npm packages, StackBlitz infrastructure & \cellcolor{lightgray}Full-stack prototyping, quick demos, cloud-native development without local setup \\
    \midrule
    \multirow{4}{2cm}{\centering\raisebox{-130pt}{\textbf{CLI}}} 
    & \cellcolor{white}Claude Code~\citep{claude-code} & \cellcolor{white}Terminal-based agentic coding tool with autonomous capabilities & \cellcolor{white}Powered by Claude Sonnet 4.5, local-first execution, MCP integration, checkpoint/rewind system, VS Code extension & \cellcolor{white}Complex terminal workflows, autonomous code refactoring, multi-file operations \\
    \cmidrule{2-5}
    & \cellcolor{lightgray}Gemini CLI~\citep{googlecli} & \cellcolor{lightgray}Open-source AI agent for terminal-native development & \cellcolor{lightgray}Gemini 2.5 Pro with 1M token context, agent mode with multi-step planning, free tier (60 RPM, 1000 requests/day) & \cellcolor{lightgray}Terminal-first developers, Google Cloud workflows, budget-conscious teams \\
    \cmidrule{2-5}
    & \cellcolor{white}Qwen Code~\citep{qwencode} & \cellcolor{white}Research-focused CLI tool for agentic coding tasks & \cellcolor{white}Powered by Qwen3-Coder models (480B MoE), 256K-1M token context, enhanced parser & \cellcolor{white}Open-source enthusiasts, researchers, privacy-focused local deployments \\
    \cmidrule{2-5}
    & \cellcolor{lightgray}Aider~\citep{aider} & \cellcolor{lightgray}Open-source AI pair programming in terminal & \cellcolor{lightgray}Multi-LLM support (Claude, GPT-4o, DeepSeek), codebase mapping, automatic Git commits, voice commands & \cellcolor{lightgray}Budget-conscious developers, multi-model workflows, Git-centric development \\
    \cmidrule{2-5}
        & \cellcolor{white}Codex~\citep{codex} 
    & \cellcolor{white}Cloud-based software engineering agent that writes features, fixes bugs, answers codebase questions, and proposes PRs in isolated sandboxes 
    & \cellcolor{white}Parallel task execution; per-task cloud sandboxes; CLI \& IDE extension; GitHub PR (@codex) integration; configurable via \texttt{\~{}/.codex/config}; MCP tool access 
    & \cellcolor{white}Pro teams automating code reviews, refactors, tests, and on-call maintenance across large repositories \\
    \cmidrule{2-5}
    & \cellcolor{lightgray}Cline~\citep{cline} 
    & \cellcolor{lightgray}Open-source coding agent in IDE/CLI that understands whole codebases, plans complex changes, and executes multi-step tasks with user approval 
    & \cellcolor{lightgray}Client-side, zero-trust architecture; model-agnostic (Claude/Gemini/etc.); reads/writes files, runs terminal commands \& browser; MCP integration; SOC 2/GDPR; enterprise team/billing options 
    & \cellcolor{lightgray}Enterprise/security-sensitive orgs; large-codebase refactors; multi-model workflows needing transparent, auditable control \\

    \bottomrule
    \end{tabular}%
    }


\end{table*}

\subsubsection{Code-based Action Implementation}

Contemporary systems implemented structured workflows through five-stage processes: planning, selection, parameter extraction, invocation, and result integration \citep{srinivas2024retrieval, sapkota2025agents, yao2022react}, following systematic patterns integrating generation with continuous mechanisms creating closed-loop systems where outcomes influence planning. Advanced workflows implement sophisticated error recovery creating robust pipelines, with environment interaction relying on multi-turn patterns following ReAct iteratively gathering information and taking actions \citep{lin2025agent, yao2022react, huang2024levels}.

Code generation emerged as unified action paradigm consolidating diverse agent behaviors \citep{dong2025survey, ishibashi2024self, zhao2025llm}. Executable code-based approaches \citep{wang2024executable} integrated Python interpreters enabling dynamic revision of prior actions upon observations through multi-turn interactions, achieving substantial improvements across API benchmarking compared to JSON-based alternatives. Repository-level systems \citep{zhang2024codeagent, dong2025survey} extended capabilities from isolated snippets to complex projects with intricate dependencies, leveraging external tools for information retrieval, navigation, and implementation assistance. Multi-agent architectures \citep{huang2023agentcoder} decomposed programming tasks through specialized coordination with programmer, test designer, and executor agents achieving superior performance while reducing computational overhead, while repository-scale evolution systems \citep{yu2025autonomous} demonstrated autonomous improvement through iterative planning across hundreds of files and thousands of lines, with evolved implementations outperforming human-designed solutions. Reinforcement learning approaches \citep{gehring2024rlef} enabled learning optimal strategies through policy optimization producing implementations mirroring human preferences and achieving state-of-the-art results while surpassing reference solutions, with agent frameworks \citep{wang2024openhands} establishing comprehensive platforms enabling code writing, command-line operations, and multi-agent coordination supporting diverse implementations across multiple benchmarks. 
Beyond conventional synthesis tasks, code generation demonstrated versatility for structured knowledge applications, with KnowCoder pioneering Python class-based schema representations enabling unified information extraction through two-phase learning framework combining code pretraining and instruction tuning \citep{li2024knowcoder}. Building upon this foundation, KnowCoder-X subsequently addressed cross-lingual transfer limitations by standardizing multilingual schemas as Python classes, achieving substantial improvements over commercial systems through bilingual alignment instruction tuning \citep{zuo2024knowcoder}. Further advancing this paradigm, KnowCoder-V2 extended capabilities from extraction to deep analysis by bridging knowledge organization with analytical reasoning via unified code generation, thereby enabling systematic processing of complex knowledge analysis tasks \citep{li2025knowcoder}.

\subsection{Reflection: Iteration, Validation, and Debugging} \label{sec:coding_agent_ref}

\subsubsection{Iterative Refinement}
Effective code generation requires systematic refinement beyond single attempts \citep{chen2021evaluating, jin2025reveal}. Large Language Models demonstrated significant potential in automated code generation \citep{chen2021evaluating}. However, accuracy remains limited particularly for complex tasks demanding comprehensive understanding \citep{jin2024rgd}. This stems from inherent challenges in simultaneously comprehending natural language and producing correct code without automatic refinement, making single-attempt approaches insufficient \citep{jin2024rgd, jiang2023self}.
Researchers developed multi-round frameworks employing iterative refinement, significantly improving synthesis quality \citep{madaan2023self, bi2024iterative, grishina2025fully}. These enable LLMs to learn from errors through reflection mechanisms incorporating failed tests and using outcomes to enhance subsequent attempts \citep{jin2024rgd}. The approaches evolved to incorporate feedback from multiple sources \citep{zhang2024pair, islam2024mapcoder}. Direct approaches integrate compiler feedback where code LLMs generate, execute, and improve code based on results \citep{bi2024iterative}. Modern frameworks employ multi-agent architectures with coder and critic agents, where coder generates and critic analyzes providing feedback in iterative loops \citep{dong2023self, islam2024mapcoder, huang2023agentcoder, qian2023chatdev}. Sophisticated systems implement specialized roles such as Reflection Agents, Thinking Agents, and Execution Agents \citep{kumar2025saarthi}.
\subsubsection{Code Validation}
Code validation evolved into sophisticated multi-layered approaches using dedicated critic agents conducting thorough analyses focusing on efficiency and functional correctness \citep{almorsi2024guided, jin2025reveal, islam2024mapcoder}. Automated testing provides quantitative metrics, categorized into using built-in benchmark test sets or LLM-generated test sets \citep{almorsi2024guided}. Advanced approaches leverage LLM-as-a-Judge frameworks, producing not only rating scores but also detailed feedback to repair incorrect code \citep{vo2025llm}.
\subsubsection{Intelligent Debugging}
Self-reflection represents fundamental shift from one-shot generation to iterative refinement, involving prompting LLM to review earlier outputs identifying logical flaws \citep{madaan2023self, jin2025reveal}. Research identified post-execution self-debugging, analyzing code after execution, struggling with bias \citep{chen2025revisit}. In contrast, in-execution self-debugging examines intermediate states showing more promise \citep{chen2023teaching, zhong2024ldb, chen2025revisit}. Advanced frameworks employ search algorithms systematically exploring multiple debugging paths, enabling comprehensive repair capabilities \citep{goues2019automated}.
Testing and debugging methodologies involve evaluating across multiple domains \citep{haroon2025how, majdoub2024debugging, ribeiro2023large, alshahwan2024automated, grishina2025fully}. Industrial deployment shows promising results, with Meta's TestGen-LLM achieving 75\% building correctly, 57\% passing reliably, and 25\% increasing coverage \citep{alshahwan2024automated}. The evolution toward multi-agent collaboration represents significant advancement mimicking real-world development practices \citep{ashrafi2025enhancing}. Modern frameworks implement specialized architectures coordinating debugging with specialized agents as bug explainers, fault locators, and patch proposers \citep{vinh2025repeton}. Integration of multiple debugging and reflection techniques demonstrates exceptional improvements, with some reaching 98.2 on HumanEval employing debugging mechanisms decomposing failed code using Control Flow Graph analysis \citep{ashrafi2025enhancing}.

\subsection{Agent Collaboration}  \label{sec:coding_agent_agent}

\subsubsection{Collaboration Mechanisms}
Complex software engineering tasks increasingly demand coordinated efforts beyond single-agent capabilities \citep{he2025llm, hong2024metagpt}. 
Multi-agent systems represent significant shift from single-agent approaches to collaborative frameworks enabling multiple agents to collaborate by distributing responsibilities, facilitating scalable execution, improving resilience, and allowing dynamic adaptation \citep{tran2025multi, he2025llm, sarkar2025survey, islam2024mapcoder}. 
Communication serves as cornerstone of coordination, representing medium through which collective reasoning emerges \citep{tran2025multi}, with LLMs enhancing communication, coordination, and decision-making enabling agents to interpret complex instructions \citep{du2023improving, liang2023encouraging, sarkar2025survey, qian2023chatdev}.
In software engineering contexts, LLM Multi-Agent Systems represent emerging approach where multiple agents are assigned specific roles to collaboratively execute complex development tasks, dividing responsibilities among specialized agents emulating real-world roles \citep{ha2025evaluating}. The field has developed several distinct communication architectures including four primary patterns: layered, decentralized, centralized, and shared message pools \citep{yao2024comal, ha2025evaluating}.
Complementing these communication structures, role-based collaboration represents prominent strategy with LLM-based agents assuming specialized roles assigned sub-tasks to solve objectives \citep{li2024more, hong2024metagpt, bell2025future}, demonstrating considerable success with agents typically assigned roles based on expert knowledge implemented in frameworks like AutoGen and CrewAI using teams with different roles through task decomposition and specialization \citep{lou2025drf}. This approach categorizes into pipeline approaches where agents sequentially complete task parts, group discussion frameworks where agents communicate to reach agreements, or hybrid combinations \citep{liang2023encouraging, gao2023large, mishra2025teammedagents, bell2025future}.
Specifically, shared message pool represents sophisticated communication approach serving as centralized hub where agents post, access, and interpret messages facilitating asynchronous reasoning. In MetaGPT, the pool operates as structured system enabling collaboration where agents publish and subscribe to typed messages, reducing communication overhead \citep{ha2025evaluating, qian2023chatdev}.
\subsubsection{Framework Implementations}
Building upon these mechanisms, several frameworks demonstrate practical implementations of multi-agent collaboration. MetaGPT enables collaborations streamlining software engineering workflow through role-specific prompts establishing effective cooperation implementing standard operating procedures \citep{lu2023towards, khanzadeh2025agentmesh, hong2024metagpt}. Similarly, ChatDev represents chat-powered framework where specialized agents are guided via chat chain modeling virtual software company communicating to autonomously generate software \citep{qian2023chatdev, mou2024from, khanzadeh2025agentmesh}. Additionally, CAMEL introduces framework enabling autonomous cooperation through role-playing, providing theoretical foundations for agent interaction.
Multi-agent role-based collaboration has demonstrated consistent performance improvements leveraging collective intelligence \citep{du2023improving}. MapCoder employs four specialized agents showcasing exceptional capabilities, achieving state-of-the-art results including 93.9\% on HumanEval, 83.1\% on MBPP, 22.0\% on APPS, 28.5\% on CodeContests, and 45.3\% on xCodeEval \citep{islam2024mapcoder}. MetaGPT demonstrated impressive accuracy through Standardized Operating Procedures \citep{hong2024metagpt, mishra2025teammedagents}.
Beyond performance metrics, role-based systems offer significant cost advantages. ChatDev demonstrates how communicative agents facilitate seamless workflows, proving cost-effective and capable of proactively addressing errors, simplifying processes and improving code quality by mimicking human teamwork patterns \citep{qian2023chatdev}. These implementations collectively represent fundamental shift toward collaborative intelligence in software engineering \citep{tran2025multi}.

\begin{figure}[t]
    \centering
    \includegraphics[width=\linewidth]{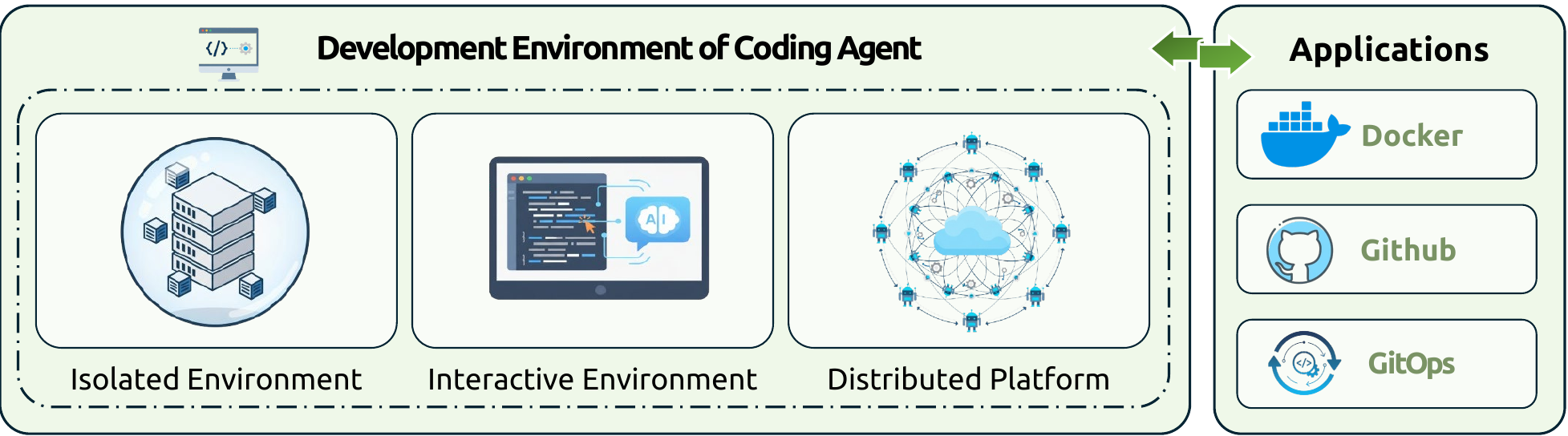}
    \caption{Development Environment Architecture for Coding Agents: Isolated Environment, Interactive Environment, and Distributed Platform.}
    \label{fig:env}
\end{figure}

\section{Development Environment of Coding Agent}
\label{sec:dev_environment}

\subsection{Isolated Execution Runtime Environment} \label{sec:dev_environment_iso}

\subsubsection{Containerization Technologies}
Containerization has become a fundamental component of the modern coding agent infrastructure. By encapsulating software, dependencies, and runtime configurations within portable images, containerization guarantees a consistent execution environment across diverse hardware and operating systems \citep{ferro2022containerization, santoro2023virtual, lee2023kunerva}. Containers rely on operating system–level virtualization to achieve lightweight isolation, rapid startup, and efficient resource utilization compared with full virtual machines \citep{ferro2022containerization, lounissi2025funda}. OS-level abstraction enables superior resource efficiency with quicker startup than virtual machines while providing agility, portability, reproducibility, modularity, and flexibility \citep{lee2023kunerva, sultan2019container, santoro2023virtual, pahl2019cloud, romanov2024optimizing, baresi2020cocos, amaral2015performance, morabito2017virtualization, pahl2015containers}. These properties are particularly valuable for coding agents, which often need to execute user-generated code repeatedly and securely while maintaining reproducibility and scalability \citep{moulton2024confronting, boettiger2014introduction, adamidi2025virtual, kurtzer2017singularity}.

Docker remains the dominant technology for constructing and distributing container images. Its layered architecture, based on Dockerfiles and registries, allows for incremental builds and efficient updates \citep{zhou2022drive, sarishma2021systematic}.In addition to Docker, lightweight virtualization frameworks such as Linux Containers (LXC/LXD) and specialized research sandboxes, including SandboxFusion and MPLSandbox, are increasingly employed in experimental systems that demand stronger isolation or multi-language support \citep{baresi2023qualitative, miller2020towards, cassano2023multipl}. 

At the orchestration level, Kubernetes has become indispensable in managing containerized workloads. It assigns coding agent tasks into isolated pods, allocates computational quotas, and enforces automatic timeouts to prevent resource monopolization \citep{su2024large}. These mechanisms collectively ensure that agent-generated code can be executed in a controlled, reproducible, and efficient manner.

\subsubsection{Security Isolation Mechanisms}
Because large language models can generate arbitrary and potentially unsafe code, strict security isolation is essential. Sandbox-based systems provide the first line of defense by limiting the privileges of agent-executed code \citep{jimenez2023swe, rabin2025sandboxeval, nunez2024autosafecoder, gajbhiye2024secure, bahadur2025securing}. Tools such as gVisor intercept system calls to enforce fine-grained safety policies, while lightweight hypervisors and microkernels enhance separation between host and container \citep{ rokon2020sourcefinder}. Multi-layer sandboxing frameworks such as SAFE-LLM and NatiSand combine container isolation with dynamic monitoring to detect unsafe system behavior \citep{rabin2025sandboxeval, liu2023your, siddiq2023sallm, abbadini2023natisand}. 

Hardware-assisted isolation techniques further strengthen security. Intel PKRU memory protection and ARM TrustZone restrict unauthorized memory access and system control \citep{kirth2022pkru}. WebAssembly-based execution engines also provide deterministic and memory-safe environments for running untrusted agent code \citep{kanellopoulos2024virtuoso}.

Contemporary coding agents typically embed execution backends that incorporate both static and dynamic policy enforcement. These backends restrict file I/O, network connections, and environment variables, while maintaining extensive logging to support traceability and auditability \citep{wang2025agentspeccustomizableruntimeenforcement, nunez2024autosafecoder}. Such hybrid security design enables continuous testing of agent-generated code without exposing the host system to unverified or malicious behaviors.

\subsubsection{Cloud-based Execution Platforms}

Cloud-based execution platforms extend the principles of containerization and sandboxing to distributed environments. They enable scalable execution across clusters, support dynamic resource allocation, and facilitate transparent workload scheduling \citep{feilhauer2016def, abolfazli2013cloud, bhm2022cloud}. Production deployments operate on 25,000 CPU core clusters organized into specialized pod configurations separating judging pods for evaluation from execution pods for code running with exclusive core assignment ensuring isolation and consistent performance, supporting AWS, Azure, and Google Cloud Platform with Kubernetes-based solutions enabling transparent remote execution \citep{tagliabue2023reasonable}. 

Resource management involves containerized environments using Docker ensuring consistent execution across infrastructure configurations with rapid system reset capabilities \citep{vijayvargiya2025openagentsafety, gong2025tuning}. Security isolation relies on containerized sandbox environments creating multiple protection layers with Docker containers providing isolated file systems, process spaces, and network spaces implementing strict CPU and memory limits through Kubernetes management systems enforcing timeout strategies automatically terminating environments exceeding specified durations \citep{su2024large, waghjale2024ecco, ardebili2025kubeintellect}.

Runtime environments support diverse programming language ecosystems including C, C++, Java, Python, Rust, Go, C\#, and PHP with execution pods handling specific compilation and runtime requirements enabling agents to work across different paradigms. Platforms use reproducible cloud compute instances providing consistent virtual hardware configurations deployed on standardized instances like Amazon EC2 with execution engines like JUDGE0 supporting over 60 languages through sandboxed Docker environments \citep{waghjale2024ecco}. Modern implementations employ dedicated Google Cloud Platform servers with standardized configurations ensuring consistent execution eliminating interference between optimization runs \citep{gong2025tuning}. Distributed frameworks implement two-tier pod design patterns separating execution from evaluation with large-scale implementations using 8,500 execution pods running individual code instances at concurrency 1 with exclusive core assignment while 2,000 judging pods handle complete evaluation workflows including correctness checking. Container orchestration follows microservices patterns where each agent and execution module operates in Kubernetes-managed containers with shared volumes and persistent volume claims facilitating data exchange while maintaining long-term storage of execution artifacts, logs, and results implementing secure REPL sandbox patterns governed by strict execution constraints \citep{ardebili2025kubeintellect}.

\subsection{Interactive Development Interface Environment}  \label{sec:dev_environment_int}
\subsubsection{AI-Native Development Interfaces}

The integration of coding agents into the developer’s workspace has transformed the traditional notion of an Integrated Development Environment (IDE). Instead of passively editing code, developers now engage in a continuous dialogue with the system. AI-native development interfaces embed large language models directly within the editor, combining code completion, conversational reasoning, and contextual memory \citep{vaswani2017attention, devlin2019bert, brown2020language, raffel2019exploring}.

Two dominant paradigms have emerged. The first emphasizes inline suggestion, where the system predicts the next tokens or statements directly within the editor, offering context-sensitive autocompletion and refactoring hints. The second paradigm emphasizes conversational interaction, in which the developer communicates with the agent using natural language to specify intent, request explanations, or perform debugging \citep{sergeyuk2025human, sergeyuk2024ide, vaithilingam2023towards, robe2022designing, ross2023programmer, weber2024significant}.

Studies show that integrating both paradigms within a unified interface produces higher task efficiency and user satisfaction. Inline completion supports rapid iteration, while conversational interaction enhances explainability and collaboration \citep{sergeyuk2024ide, ross2023programmer}. Emerging IDEs such as Cursor and Q Developer integrate contextual memory and multi-turn reasoning, maintaining persistent dialogue histories to provide more coherent assistance \citep{donato2025multimind, ross2023programmer, nam2023using, pinto2023developer}.

\subsubsection{Remote Development}

The growth of cloud-native development practices has led to widespread adoption of remote and containerized development environments. Systems such as GitHub Codespaces \citep{github_codespaces_2024} and DevContainer provide pre-configured, reproducible workspaces accessible via browsers or local editors. This paradigm allows coding agents to operate directly within standardized, resource-isolated environments synchronized with version control systems.

Remote development also supports computation-intensive workflows by allowing resource-demanding tasks to run on remote servers while the developer interacts through lightweight clients. SSH-based connections and browser-based terminals enable seamless transitions between local and cloud execution contexts. These capabilities are essential for agent-assisted programming, as they allow distributed collaboration while maintaining strict control over dependencies and runtime configurations \citep{jetbrains_gateway_2022}.

\subsubsection{Tool Integration Protocol Standards}

Standardization efforts have become increasingly important for achieving interoperability between coding agents and existing developer tools. The Model Context Protocol (MCP) \citep{hou2025model, ray2025survey} defines a universal interface for exchanging contextual information such as source code, documentation, and environment state. The Language Server Protocol (LSP) \citep{bork2023language, gunasinghe2021language} provides language-agnostic features including diagnostics, completion, and code navigation, while the Debug Adapter Protocol (DAP) \citep{ernst2021deductive, garcia2024cross} standardizes debugging interactions.

Integrating these protocols allows coding agents to function as first-class participants within conventional development ecosystems. Agents can request, modify, and analyze project context seamlessly, while maintaining compatibility with existing IDEs and version control systems \citep{li2025glue, koc2025mind, wu2025git}. This protocol convergence promotes an open, extensible environment for hybrid human–AI programming collaboration \citep{gitkraken2025mcp, vs2025agent}.

\subsection{Distributed Orchestration Platform Environment} \label{sec:dev_environment_dis}

\subsubsection{CI/CD Pipeline Integration}

Continuous Integration and Continuous Deployment (CI/CD) frameworks are essential for maintaining the stability and quality of software systems. Within the context of coding agents, CI/CD processes ensure that automatically generated code passes through rigorous validation and testing stages before integration \citep{singh2023microservices, dima2018waterfall, highsmith2002agile, beck2000extreme}. Pipeline-as-Code practices enable reproducibility by representing build and deployment workflows as versioned artifacts \citep{singh2023microservices, ivanov2022extended}. In cloud-native ecosystems, Kubernetes and Jenkins-based automation frameworks such as KubeSphere facilitate modular CI/CD pipelines. These pipelines automatically handle code building, testing, and deployment, while ensuring that agent-generated contributions are subject to the same review and validation processes as human-written code.

In the emerging paradigm of Vibe Coding, CI/CD pipelines are further extended with agent participation. Agents generate code, perform static analysis, and conduct self-testing, but human oversight remains indispensable for critical verification stages. Designing CI/CD architectures that integrate agents safely and transparently is a key challenge for reliable large-scale deployment \citep{singh2023microservices}.

\subsubsection{Cloud Compute Orchestration}
Distributed orchestration platforms employ frameworks dynamically provisioning computational resources with production deployments demonstrating scale operating on 25,000 CPU core clusters organized into specialized pod configurations \citep{feilhauer2016def}. The orchestration landscape has been enhanced by TOSCA (Topology and Orchestration Specification for Cloud Applications) providing interoperable models for describing cloud applications as typed directed topology graphs representing components as nodes and dependencies as links enabling developers to model cloud service structures with automated lifecycle management \citep{bisicchia2023continuous, brogi2018tosker, dehury2021toscadata}. TOSCA's platform-agnostic nature allows users to switch between cloud providers enabling applications to be migrated between vendors with minimal modification demonstrating strong compatibility with CI/CD technologies facilitating easier testing, re-deployment, and re-engineering \citep{dehury2021toscadata, tomarchio2021torch}, with modern platforms like FogArm providing autonomous adaptation capabilities responding to application specification changes from CI/CD pipelines and infrastructural variations detected through distributed monitoring \citep{bisicchia2023continuous}. LLM-powered component integration adds complexity and opportunity where frameworks built on large language models, retrieval-augmented generation, and deep learning enable seamless integration with CI/CD pipelines supporting multi-format output generation and incorporating continuous learning for optimization \citep{joshi2025architecting}. Modern implementations configure CI/CD pipelines to build, test, and deploy applications within containerized environments enabling faster build times and more frequent releases with Continuous Deployment integrating directly with LLMOps to automatically deploy model updates based on performance monitoring enabling automated model transitioning, deployment, and monitoring while maintaining high reliability using GitOps pipelines ensuring automatic redeployment of updated agent code or model weights with Canary releases and A/B testing \citep{choi2025intelligent, song2025llm}. 

\subsubsection{Multi-Agent Collaboration Frameworks}
As coding tasks become increasingly complex, single-agent systems struggle to manage the breadth of reasoning, planning, and verification required for end-to-end software generation. Multi-agent collaboration frameworks address this limitation by coordinating multiple specialized agents that share contextual memory and cooperate through structured communication channels.

Frameworks such as AutoGen, CrewAI, MetaGPT, and LangGraph exemplify this paradigm. AutoGen organizes heterogeneous agents that perform role-specific tasks such as requirement analysis, implementation, and unit testing  \citep{wu2023autogen, zhu2023autogen, porsdam2023autogen}. CrewAI conceptualizes agent collaboration as a simulation of human teams, incorporating negotiation, task assignment, and feedback integration  \citep{taulli2025crewai}. MetaGPT introduces a meta-planning layer, where high-level agents design task hierarchies and delegate execution to subordinate agents, thereby achieving structured decomposition \citep{hong2024metagpt, zhou2024metagpt}. LangGraph formalizes collaboration as a graph of nodes and edges representing agents and communication pathways, which enables systematic reasoning over dependencies and task flows \citep{chen2025implementing}.

These frameworks collectively redefine how coding agents can scale across complex development pipelines. They provide fault tolerance by redistributing tasks when individual agents fail, improve modularity through role specialization, and support iterative verification through feedback loops among agents. Recent studies suggest that multi-agent collaboration enhances both the reliability and interpretability of large-scale code synthesis \citep{nasir2025code, williams2025multi}.

Despite these advances, challenges remain regarding consistency, communication efficiency, and conflict resolution. Overly frequent message exchanges may lead to coordination overhead, while insufficient synchronization can cause semantic drift among agents. Ongoing research is exploring graph-structured communication protocols and shared reasoning buffers to improve coherence across collaborative agents \citep{zeng2025edge, luo2025ai}.

The development of multi-agent frameworks represents an essential direction for the evolution of Vibe Coding systems. These frameworks not only enhance scalability and adaptability but also embody a shift toward distributed intelligence in software engineering, where human developers and autonomous agents collaborate as peers within integrated ecosystems \citep{hughes2025ai}.

\begin{figure}[t]
    \centering
    \includegraphics[width=\linewidth]{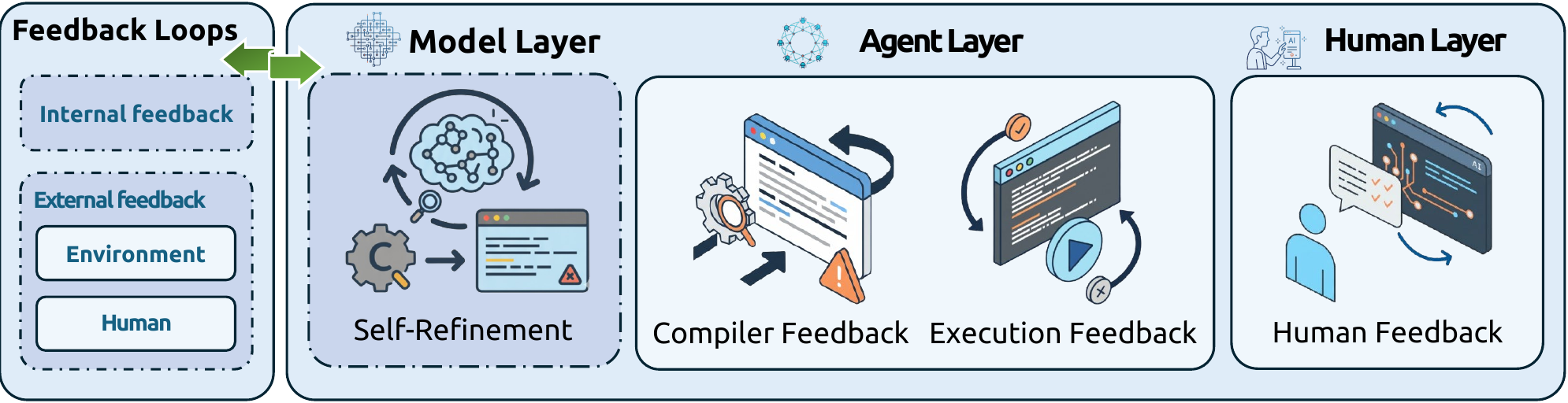}
    \caption{Feedback Loops Framework for Coding Agents: Internal feedback via Self-Refinement, and External feedback from Environment via Compiler Feedback and Execution Feedback, and from Human Feedback.}
    \label{fig:feedback}
\end{figure}

\section{Feedback Mechanisms}
\label{sec:feedback}
\subsection{Compiler Feedback} \label{sec:feedback_com}
\subsubsection{Syntax and Type Error Feedback}

LLM-based coding agents rely on compiler feedback to iteratively improve generated code through error detection and repair cycles \citep{han2025convcodeworld, peng2025criticlean, brown2020language, madaan2023self, bi2024iterative, bouzenia2024repairagent}. Compiler feedback encompasses multiple granularities: coarse-grained feedback provides binary compilation success, while fine-grained feedback delivers detailed error reasons and locations \citep{ravi2025llmloop}. Beyond basic compilation feedback, systems integrate test feedback from automated frameworks and static analysis from tools like pylint \citep{dai2025feedbackeval}.

Modern architectures integrate compiler feedback through multi-agent systems employing Code Teachers reassessing bug addressing, Review Agents examining compilation logs generating structured prompts, reinforcement learning approaches with RLCF, PPOCoder, and COMPCODER, debugging agents using retrieval-augmented generation and Chain-of-Thought methodology, and tool-integrated approaches like CoderAgent \citep{ravi2025llmloop, zhan2025coderagent}. Domain-specific implementations include AutoChip incorporating compilation error messages, RTLFixer using retrieval-augmented generation for Verilog errors, and RTLCoder introducing code quality scoring \citep{tsai2023rtlfixer}.

Significant limitations persist in compiler feedback. Compilers report transformation failures through coarse error messages providing insufficient insight \citep{li2024review, siso2019evaluating, bi2024iterative}. Project-specific context often cannot fit into LLM prompts, while workflow complexity creates inconsistent approaches \citep{bi2024iterative, li2024review}.

Accordingly, compiler feedback has evolved from basic validation to sophisticated iterative refinement providing detailed feedback including compilation success status, optimization pass information, and instruction counts \citep{grubisic2024compiler, huang2024opencoder}. Effectiveness depends heavily on context—incorporating compiler feedback improves accuracy, though highlighting errors without resolution context shows limited improvement \citep{bi2024iterative}.

\subsubsection{Static Analysis Feedback}

LLMs demonstrate exceptional capabilities for understanding complex compiler messages and static analysis warnings \citep{abtahi2025augmenting, fang2024large}. Correspondingly, static analysis tools analyze code without execution, employing predefined rules to enforce standards and detect bugs, though producing large volumes of warnings including false positives \citep{lacombe2023combining, srinivasarao2024software, li2024iris, keltek2025lsast, wadhwa2024core}. LLMs achieve 69-81\% accuracy in prioritizing warnings and 81.13\% precision with 94.64\% recall in inspecting thousands of static warnings \citep{becker2016effective, mohajer2023skipanalyzer}.

Current systems employ comprehensive verification through multiple layers: syntactic correctness checking through compilers like GCC and Python interpreters, specialized tools including Black for formatting and nuXmv for formal verification, code complexity analysis utilizing OClint and srcSlice \citep{tu2023isolating}, and vulnerability detection employing Frama-C and Slither for Ethereum smart contracts \citep{feist2019slither, hu2025qlpro}. Static analysis extends to specification verification using Frama-C to verify AI-generated specifications \citep{tu2023isolating}.

Multi-agent architectures integrate static analysis into LLM-powered coding systems \citep{he2025llm}. AutoSafeCoder implements three-agent frameworks with Coding Agents, Static Analyzer Agents, and Fuzzing Agents \citep{nunez2024autosafecoder}. Reinforcement learning frameworks like REAL use program analysis-guided feedback combining automated signals with unit tests \citep{yao2025training}. Applications span general programming detecting vulnerabilities in C \citep{dolcetti2024helping}, hardware description language generation with AutoChip \citep{chang2025data}, and HDLAgent supporting multiple languages \citep{kashanaki2025beyond}.

\subsubsection{Runtime Compilation Feedback}

LLMs produce functional code but often struggle with correctness and optimization \citep{tao2024codelutra, pearce2023examining, mathews2024test}. LLM-based coding agents leverage dynamic feedback from compilation and runtime processes to iteratively improve code output, transforming traditional one-shot code generation paradigms into iterative refinement processes \citep{han2025convcodeworld, zheng2025vectrans, pirkelbauer2025compilergpt, li2024redo}. Coding agents utilize several feedback categories: binary pass/fail indicators, compilation feedback identifying syntax errors, failed tests feedback providing expected values with tracebacks, execution feedback providing runtime information, LDB feedback printing intermediate variable values, self-feedback determining compilation success, verification feedback checking unit tests and formal verification, and verbal feedback providing human-readable correction suggestions \citep{dong2025survey}.

Several frameworks leverage compilation feedback: CompCoder implementing three-stage pipelines improving compilation success from 44.18\% to 89.18\%, CodeRL using actor-critic reinforcement learning with unit test signals \citep{le2022coderl, jiang2024survey, shojaee2023execution}, RLCF training pre-trained LLMs using feedback from grounding functions for code quality assessment \citep{jain2023coarse}, INTERVENOR using compiler feedback for agent collaboration, CoCoGen using static analysis to identify mismatches and iteratively fixing errors using compiler feedback and repository information \citep{bi2024iterative}, self-debugging frameworks like Cycle, LDB, and ReflectionCoder leveraging runtime execution information for iterative refinement \citep{ashrafi2025enhancing, ding2024cycle}, CompilerGPT automating interaction between compilers, LLMs, and evaluation systems \citep{pirkelbauer2025compilergpt}, and hardware-specific frameworks like AutoChip for RTL/Verilog generation demonstrating domain-specific effectiveness \citep{kashanaki2025beyond, tsai2023rtlfixer}.

Integration of compilation and runtime feedback finds applications across code translation ensuring semantic correctness through methods like TransMap and AlphaTrans \citep{ibrahimzada2024alphatrans}, iterative code refinement where execution feedback serves as supervisory signal \citep{gehring2024rlef, zeng2025acecoder, ni2025viscoder, yang2023intercode}, automatic program repair showing iterative patch creation benefits \citep{ruiz2025art}, visualization \citep{ni2025viscoder}, and performance optimization where compilers provide structured feedback explaining optimization failures.

\begin{table*}[!t]
    \centering
    \scriptsize
    \caption{Comparison of Coding Environments and Datasets for AI-Assisted Software Engineering.}
    \label{tab:coding-env-datasets}
    \renewcommand{\arraystretch}{1.2}
    \definecolor{lightgray}{RGB}{245,245,245}
    \resizebox{\textwidth}{!}{%
    \begin{tabular}{p{2.2cm}|p{1cm}|p{3cm}|p{4cm}|p{1.8cm}|p{4cm}}
    \toprule
    \textbf{Environment/Dataset} & 
    \textbf{Year} & 
    \textbf{Task Scale} & 
    \textbf{Data Source} & 
    \textbf{Task Level} & 
    \textbf{Interaction \& Evaluation} \\
    \midrule
    \rowcolor{lightgray}\textbf{SWE-Gym}~\citep{pan2024training} & 2024 & 2,438 verified instances (11 repos); 64,689 raw instances & Constructs interactive execution environments from GitHub issues in 11 Python repos; each environment includes the full codebase and failing tests & Repository-level RL environment & Agents modify files and run unit tests within Docker; rewards are given when all tests pass, supporting iterative RL training \\
    \rowcolor{white}\textbf{SWE-smith}~\citep{yang2025swesmith} & 2025 & 50k tasks (250+ environments); 5,016 trajectories & Automatically converts any GitHub repository into a SWE-gym environment, corrupting unit tests to generate interactive tasks & Repository-level RL environment & Environment provides file-localization and repair tools; agents interact through code edits and receive rewards when modified code passes tests \\
    \rowcolor{lightgray}\textbf{R2E-Gym}~\citep{jain2025r2e} & 2025 & 8.1k tasks (13 repos) & Procedurally generates interactive RL environments from commit history using the SWE-Gen pipeline & Commit/Repository-level environment & Agents interact with full repositories via Docker; hybrid execution and static verifiers produce the reward signals \\
    \rowcolor{white}\textbf{DeepSWE}~\citep{deepswe2025} & 2025 & $\approx$4.5k problems & Builds a code RL environment by filtering R2E-Gym tasks to avoid overlap with evaluation benchmarks & Repository-level RL environment & Environment defines four actions (execute, search, edit, finish); reward is given only when all tests pass within a time limit \\
    \rowcolor{lightgray}\textbf{SWE-rebench}~\citep{badertdinov2025swe} & 2025 & 21,336 tasks & Creates interactive environments from mined PRs with 1–15 file modifications and clear descriptions & Repository-level interactive environment & Uses a unified scaffold: agents edit code and run tests; environment rewards successful test passing, with ~30\% of repositories building successfully \\
    \rowcolor{white}\textbf{Multi-turn SWE RL Env.}~\citep{golubev2025training} & 2025 & 7,249 tasks & Filtered subset of SWE-rebench tasks preprocessed into a multi-turn RL environment & Repository-level multi-turn environment & Agents can execute shell commands, edit files, search and submit patches; reward equals test success minus step penalties, encouraging efficient interactions \\
    \rowcolor{lightgray}\textbf{SWE-RL}~\citep{wei2025swe} & 2025 & $\approx$1.1$\times$10$^7$ PR instances & Offline repository of PR instances curated from GitHub events (2015–2024), including bug-fix issues and patches & Commit-level offline dataset & Non-interactive; used for offline RL and reward models, where similarity between generated and ground-truth patches is computed with sequence matching \\
    \rowcolor{lightgray}\textbf{COFFEE-GYM}~\citep{chae2024coffee} & 2024 & Multiple problems (5 difficulty levels) & Builds a feedback environment from human code submission histories with paired wrong/correct solutions & Function-level feedback environment & Agents generate feedback; the environment applies edits and runs $\sim$35.5 tests per problem to produce a reward \\
    \rowcolor{white}\textbf{EvoCodeBench}~\citep{li2024evocodebench} & 2024 & 275 samples (initial version) & Curates repository-level generation tasks across 25 repos; environment updated every 6 months to avoid leakage & Repository-level generation environment & Agents must generate complete repositories; evaluation uses Pass@k and domain-specific metrics \\
    \rowcolor{lightgray}\textbf{DA-Code}~\citep{huang2024code} & 2024 & $\approx$500 data analysis tasks & Aggregates real-world data analysis tasks covering cleaning, ML, and exploration in interactive notebooks & Notebook-level interactive environment & Provides an executable sandbox for SQL/Python; evaluation based on correctness of analysis results \\
    \rowcolor{white}\textbf{DS-1000}~\citep{lai2022natural} & 2024 & 1,000 problems & Collects function-level tasks from seven data science libraries, each with prompt, solution, and tests & Function-level offline tasks & Agents write code to pass test\_execution and test\_string; limited interaction beyond code submission and test running \\
    \rowcolor{lightgray}\textbf{BioCoder}~\citep{tang2024biocoder} & 2024 & 1,026 Python functions + 1,243 Java methods + 253 Rosalind problems & Extracts bioinformatics code from GitHub and Rosalind projects, emphasizing cross-file dependencies & Function-level environment with long contexts & Evaluated using fuzz testing; environment requires handling long prompts (up to 2,600+ tokens) and cross-file reasoning \\
    \rowcolor{white}\textbf{MLE-Dojo}~\citep{qiang2025mle} & 2025 & 200+ Kaggle tasks & Encapsulates Kaggle competitions into an RL environment with unified data, evaluators, and leaderboards & Competition-level interactive environment & Agents can request information, validate and execute code; reward is based on relative performance via HumanRank \\
    \rowcolor{lightgray}\textbf{AgentPack}~\citep{zi2025agentpack} & 2025 & 1.3M code edits & Collects code edits co-authored by agents (Claude, Codex, Cursor) and humans from Git histories & Commit-level offline dataset & Offline supervised dataset with no interactive environment or defined RL reward \\
    \bottomrule
    \end{tabular}%
    }
    \vspace{0.5em}

    \renewcommand{\arraystretch}{1.0}
\end{table*}

\subsection{Execution Feedback}  \label{sec:feedback_exe}
\subsubsection{Unit Test Execution Results}

Unit test execution results serve as critical feedback mechanisms allowing coding agents to verify generated code meets requirements \citep{yang2024evaluation, alshahwan2024automated, pan2025aster, dolcetti2024helping}. Results provide varying detail levels: simple correctness feedback indicating pass/fail through binary signals, and detailed execution results including runtime errors and specific test failure information. Test-Driven Development (TDD) has proven its merit, requiring developers to write tests before the functional code \citep{beck2003test, mathews2024test, cui2025tests}. Test validation frameworks like PyTest, unittest, and JUnit serve as primary mechanisms generating feedback revealing runtime errors \citep{lai2022natural, gu2024testart, dong2025survey}.

Unit test results serve multiple applications: reinforcement learning approaches leveraging execution feedback, search-based debugging where algorithms like BESTER use execution feedback with self-reflection for automated repairs \citep{goues2019automated}, code sample filtering using test outcomes with CodeT achieving 65.8\% pass@1 on HumanEval, performance optimization analyzing execution time data \citep{chen2022codet,peng2024perfcodegen}, multi-agent systems employing execution feedback \citep{bouzenia2024you, s2025multi, zhang2024autocoderover}, and test-time scaling strategies generating multiple candidates.

Integration methodologies include reinforcement learning approaches fine-tuning models using test execution outcomes assigning rewards based on results using algorithms like REINFORCE or PPO optimizing model behavior through iterative feedback loops \citep{le2022coderl, shojaee2023execution}, code selection and filtering using execution feedback to identify best solutions with AlphaCode generating millions of programs filtering incorrect solutions based on test outcomes \citep{li2022competition}, sophisticated approaches like CodeT automatically generating test cases performing dual execution agreement analysis considering both output consistency and cross-sample agreement \citep{chen2022codet}, and structured validation frameworks implementing unit testing as iterative refinement mechanisms where agents continuously generate test cases, evaluate outputs against functional requirements, and refine code until all tests pass.

Despite promising applications, significant challenges limit effectiveness: unreliability of AI-generated unit tests where code refinement depends entirely on generated unit test correctness, with studies revealing models generate unreliable self-tests in abundance potentially misleading self-improvement processes and exacerbating program errors creating circular dependency problems \citep{liu2025llm, wang2025projecttest}, scalability constraints as generating and executing multiple test cases for each code candidate requires significant computational resources particularly when systems generate thousands of code samples, and limited test coverage as automatically generated tests may not capture all edge cases or complex scenarios potentially allowing buggy code to pass initial validation while failing in real-world deployment scenarios.

\subsubsection{Integration Test Feedback}

Integration test feedback involves executing automated tests against LLM-generated code and using resulting feedback to guide subsequent iterations, building upon traditional software testing but adapting for unique challenges of AI-generated code \citep{dai2025feedbackeval}. The fundamental premise relies on creating closed-loop systems where coding agents learn from execution feedback rather than relying solely on static analysis or human review.

Development of integration testing frameworks has evolved to address unique challenges of testing LLM-powered systems \citep{yang2023intercode}, with OpenHands framework representing notable advancement pioneering end-to-end agent test frameworks combining traditional integration testing with foundation model mocking techniques \citep{schultz2025potential, zhang2024llm, nong2024automated}.

Practical implementation relies on sophisticated multi-agent architectures where test executor agents serve as central coordinators managing interaction between code generation and test validation activities. Effectiveness varies significantly based on type and structure of feedback, with research demonstrating structured feedback formats achieving superior results, with test feedback leading to highest repair success rates \citep{dai2025feedbackeval}. The iterative nature plays crucial roles with multiple rounds enhancing repair performance, though research indicates diminishing returns after two or three iterations \citep{dai2025feedbackeval}.

Filtering and selection of generated code samples involves several established approaches. Key methods include test-based filtering using unit test outcomes \citep{li2022competition}, dual execution agreement evaluating code based on output consistency \citep{chen2022codet}, verification-based reranking training specialized verifiers \citep{zhang2024generative, zhang2023coder}, semantic execution analysis using execution results to capture semantic features \citep{ni2025viscoder}, and probabilistic marginalization combining programs with identical execution results \citep{chen2022codet}.

\subsubsection{Runtime Error and Exception Handling}

Research frameworks have systematically categorized runtime errors when LLMs generate code identifying five primary error types: syntax errors, parameter and attribute errors, output type errors, logical answer errors, and timeout and runtime errors \citep{li2024redo}. Beyond execution-level errors, comprehensive taxonomies identify 36 exception types across 12 agent artifacts distinguishing failures across reasoning, planning, and execution phases \citep{yao2022react, yao2022react}.

Research has developed several approaches for detecting and analyzing runtime errors: simplest approaches using binary feedback providing only pass/fail indicators \citep{chen2025revisit}, sophisticated systems offering failed tests feedback including expected versus actual values with runtime error tracebacks \citep{zhong2024ldb}, advanced frameworks providing failed and passed tests feedback with comprehensive input/output information \citep{ni2025viscoder}, sophisticated debugging approaches using control flow analysis and variable tracking with systems like LDB \citep{zhong2024ldb}, and emerging print debugging and reflection-based methods \citep{ashrafi2025enhancing, chen2025revisit}.

Several specialized frameworks provide comprehensive exception management: knowledge-driven approaches where Knowledge-driven Prompt Chaining decomposes code generation into AI chains with iterative check-rewrite steps \citep{ren2023from}, multi-agent frameworks like Seeker using specialized agents \citep{zhang2024seeker}, PyCapsule employing programmer agents and debugging alongside executor agents enhanced by error handlers \citep{adnan2025large}, runtime error recovery systems like Healer with GPT-4 successfully handling 88.1\% of AttributeErrors \citep{sun2024llm}, and iterative self-correction mechanisms using structured feedback loops \citep{rath2025structured}.

Self-reflection represents critical capability for LLM-based coding agents defined as processes where models review earlier outputs and attempt to identify logical flaws. When given buggy programs with execution traces, errors, or test results as input, LLMs produce debugging instructions through self-reflection \citep{jin2024rgd}. Dynamic self-correction through execution feedback represents sophisticated approaches where systems append diagnostic information to original prompts \citep{rath2025structured}. Modern frameworks have developed increasingly sophisticated debugging methodologies including print debugging methods, reflection-based approaches, hierarchical debugging tools, and advanced systems like CYCLE \citep{ashrafi2025enhancing, ding2024cycle}.

\subsection{Human Feedback}  \label{sec:feedback_hum}
\subsubsection{Interactive Requirement Clarification}

Large Language Models have demonstrated remarkable capabilities in automatically generating code from natural language requirements \citep{chen2021evaluating}, though critical gaps exist between current LLM behavior and human problem-solving approaches when dealing with ambiguous requirements \citep{mu2024clarifygpt, darji2025curiosity, fang2024large}. While human developers typically seek additional information through interactive clarification, current LLM-based code generation approaches lack mechanisms for clarifying unclear requirements, with empirical studies revealing state-of-the-art Code LLMs generate code outputs in over 63\%~of ambiguous scenarios without seeking necessary clarifications \citep{mu2024clarifygpt}. This limitation has significant implications compounded by the fact that 72\%~of software defects in production environments originate from misunderstood requirements \citep{darji2025curiosity, barke2023grounded, ross2023programmer}.

Detection of ambiguous requirements has emerged as critical first steps with the most widely adopted approach being code consistency checking pioneered by ClarifyGPT which determines whether given requirements are ambiguous by generating multiple code solutions \citep{mu2024clarifygpt}. Despite methodological advances, empirical findings reveal significant limitations with systematic evaluations showing state-of-the-art models struggle to distinguish between well-specified and underspecified instructions \citep{dong2025survey, fakhoury2024llm}.

Generation of effective clarification questions represents core technical challenges with foundational approaches established by ClarifyGPT using prompt-based question generation, demonstrating significant improvements elevating GPT-4's performance from 70.96\% to 80.80\% on MBPP-sanitized benchmarks \citep{mu2024clarifygpt}. To address limitations, fine-tuning approaches have emerged with ClarifyCoder representing notable advances using synthetic data generation and instruction-tuning \citep{marozzo2025iterative, darji2025curiosity, mu2024clarifygpt, lahiri2022interactive}.

Evaluation of interactive requirement clarification systems has evolved toward systematic assessment frameworks. Empirical findings reveal significant performance gaps—while models struggle to distinguish between well-specified and underspecified instructions, when they do interact for underspecified inputs they effectively obtain vital information from users leading to significant improvements with evaluations showing average absolute improvement of 45.97\% in pass@1 code generation accuracy within 5 user interactions \citep{dong2025survey, fakhoury2024llm}.

\subsubsection{Code Review Feedback}

Development of large language models has created urgent needs to align these powerful systems with human intentions and preferences, with the challenge lying in effectively utilizing qualitative human feedback \citep{kaufmann2024survey}. Foundations of this field can be traced to preference-based reinforcement learning (PbRL), with major breakthroughs coming with Christiano et al.'s introduction of RLHF \citep{christiano2017deep}.

RLHF framework follows structured approaches beginning with SFT on high-quality demonstration data, with core innovation lying in training reward models to approximate human preferences from preference data where humans indicate preferred responses between pairs of options serving as foundations for subsequent reinforcement learning optimization \citep{schulman2017proximal}. The field gained significant momentum when scaled to large language models \citep{lee2023rlaif} with InstructGPT demonstrating potential of applying RLHF to large-scale models like GPT-3 enabling alignment with user instructions across wide ranges of tasks \citep{achiam2023gpt}, while DPO has emerged as notable simplification allowing language models to be trained directly on preference data without explicit reward modeling or reinforcement learning optimization stages \citep{ouyang2022training, stiennon2020learning, zaheer2017deep, rafailov2023direct, xiao2024comprehensive}.

Core RLHF methodology consists of three distinct stages: processes begin with supervised fine-tuning on high-quality demonstration data, second stages focus on reward modeling training reward models to approximate human preferences based on Bradley-Terry models, and final stages involve reinforcement learning optimization where models are refined using RL algorithms like PPO \citep{yin2025from, ouyang2022training, niekerk2025post, rafailov2023direct}. Despite effectiveness, traditional RLHF pipelines are notably complex and computationally expensive \citep{rafailov2023direct}.

Traditional RLHF pipelines suffer from inherent complexity and instability creating practical barriers \citep{rafailov2023direct}. Major practical constraints include substantial costs and efforts required for human annotation often requiring thousands of human comparisons to train reliable reward models \citep{kiruluta2025history, kopf2023openassistant}. Scalability challenges become even more pronounced with complex tasks as dialogues become longer spanning multiple turns greatly increasing annotation difficulty \citep{kiruluta2025history}.

Given these limitations, complexity and instability of traditional RLHF has motivated development of more direct approaches with most significant breakthrough coming with DPO eliminating explicit reward modeling stages and directly optimizing policies using preference data leveraging mathematical relationships between optimal policies and reward functions under Bradley-Terry preference models enabling preference optimization through simple classification objectives that are stable, performant, and computationally lightweight \citep{rafailov2023direct}. Beyond DPO, researchers have explored alternative frameworks. Preference Ranking Optimization (PRO) addresses limitations of RLHF by extending pairwise preference comparisons to accommodate human preference rankings of any length directly fine-tuning language models through iterative contrast learning \citep{song2024preference}. Nash Learning from Human Feedback (NLHF) offers novel approaches learning preference models conditioned on two inputs pursuing strategies consistently generating responses more preferred than competing policies \citep{munos2023nash}. Diffusion-DPO demonstrates effectiveness in text-to-image generation \citep{weng2024navigating, wallace2023diffusion}.

Substantial costs required for human annotation in traditional RLHF have motivated researchers to explore AI-generated feedback as scalable alternatives with RLAIF introducing frameworks where AI-generated feedback replaces human annotations with off-the-shelf LLMs serving as feedback models to evaluate responses based on predefined principles demonstrating performance comparable to traditional RLHF methods across tasks like summarization and dialogue generation \citep{niekerk2025post, ouyang2022training, lee2023rlaif, chai2024ma}. AutoPM represents another approach employing LLMs to automatically generate preference data by eliciting pairwise comparisons based on helpfulness, honesty, and harmlessness criteria enabling training of preference models without extensive human supervision \citep{niekerk2025post, yin2025from}.

\subsection{Self-Refinement Feedback}  \label{sec:feedback_self}
\subsubsection{Self-Evaluation and Critique}

Foundational architecture of self-refinement in LLMs centers around iterative feedback loops mirroring human revision processes with most notable framework Self-Refine enabling LLMs to improve outputs through structured three-step processes: generating initial responses, prompting models to critique own outputs, and refining responses based on feedback \citep{sahoo2024systematic, madaan2023self, he2025can}. Processes follow key steps: initial generation, self-critique, refinement, and iteration, with notably Self-Refine requiring no additional training data or fine-tuning \citep{madaan2023self, wei2022chain}. Multiple variants have emerged including Recursively Criticizes and Improves (RCI), CRITIC, and Self-Correct \citep{gou2023critic, sahoo2024systematic}.

Executable nature of code provides unique advantages for self-refinement in programming tasks as execution results offer immediate and objective feedback \citep{zheng2024opencodeinterpreter, madaan2023self, shinn2023reflexion}: Self-Debugging equipping models with ability to autonomously detect and repair errors, Self-Edit incorporating dedicated fault detection phases, CodeChain introducing self-revision mechanisms, Self-Collaboration using simulated internal dialogue, LeTI redefining code generation as interactive dialogue-driven processes, and OpenCodeInterpreter unifying generation, execution, and refinement.

Self-evaluation represents critical components encompassing models' ability to assess quality of own generated outputs including reasoning steps and confidence scores with multiple techniques having emerged including self-consistency, self-correction, self-evolution, and self-feedback \citep{mehandru2025bioagents}. Self-critique and feedback generation processes involve models acting as critics analyzing own outputs \citep{madaan2023self, ho2025self}. However, critical analysis reveals important limitations with research indicating self-correction works well primarily in tasks that can use reliable external feedback, and large-scale fine-tuning appears necessary to enable effective self-correction \citep{mehandru2025bioagents, kamoi2024when}.

Critique generation processes involve models analyzing own outputs and providing structured suggestions for improvement \citep{madaan2023self}. Advanced critique frameworks implement multi-dimensional evaluation systems with carefully designed prompts instructing models to assess outputs based on specific criteria \citep{ho2025self, madaan2023self, shinn2023reflexion}. Performance gains from self-refinement methods have been substantial including GPT-4 achieving gains of 8.7 points in code optimization and 13.9 points in code readability when using Self-Refine prompting \citep{sahoo2024systematic}.

\subsubsection{Multi-Agent Collaborative Feedback}

Multi-agent systems employ several distinct feedback mechanisms enabling collaborative code improvement \citep{ishibashi2024self, ashrafi2025enhancing} with most fundamental distinctions lying between inter-agent feedback and self-feedback mechanisms, with inter-agent feedback involving agents providing constructive criticism to each other based on interactions and collaborations helping agents identify improvement areas and adapt strategies for continuous learning within systems \citep{talebirad2023multi, qian2023chatdev, islam2024mapcoder}. In multi-core agent systems, intra-agent feedback emerges as agents exchange information and share observations including shared observations, alternative perspectives on problems, and evaluations of proposed plans promoting collaborative problem-solving and cross-validation enhancing overall system robustness through collective intelligence \citep{hassouna2024llm}. Self-feedback mechanisms allow agents to assess own performance and identify improvement areas through self-assessment based on predefined criteria or goals, while external feedback-based optimization leverages evaluative signals from external models or agents to refine agent behavior integrating external reflections and corrections to enhance robustness and adaptability \citep{talebirad2023multi}.

Core collaborative feedback mechanisms center on peer-reflection involving information exchange through role specialization and structured communication. Self-feedback loops represent another key collaborative mechanism where agents iteratively refine strategies based on both quantitative numerical evaluations and qualitative natural language assessments \citep{antoniades2024swe}. Feedback-refine cycles enable agents to store language-level memories and collaboratively iterate on solutions with parallel agents exchanging critiques \citep{ando2025when}. Mutual assessment and pruning mechanisms facilitate collaborative optimization where agents produce multiple candidates and evaluate outputs from other agents \citep{dong2025survey}.

Self-refinement mechanisms in multi-agent coding systems operate through iterative feedback-refine cycles demonstrating that when run in parallel with agents exchanging critiques feedback-refine cycles significantly boost code generation quality \citep{ando2025when}. Hybrid value functions represent key advancements incorporating both quantitative and qualitative assessment mechanisms \citep{antoniades2024swe}. Circular optimization mechanisms create continuous improvement loops through self-reflection scoring systems with notable frameworks like CodeCoR introducing reflection agents between code generation, testing, and repair stages \citep{dong2025survey, islam2024mapcoder}.

\subsubsection{Reflection and Memory-Based Feedback}

Several foundational frameworks have emerged as core approaches for implementing self-refinement in large language models with most notable being Self-Refine which introduced comprehensive end-to-end self-correction approaches \citep{mousavi2023critics, madaan2023self, chang2024efficient, park2025self}. Reflexion represents another foundational framework using verbal reinforcement to help agents learn from prior failures incorporating three key LLM-based modules: Actors that generate actions, Evaluators that assess results, and Self-Reflection models that analyze failure cases \citep{shinn2023reflexion, park2025self, kamalov2025evolution}.

Memory architecture represents critical components in self-reflection systems with most frameworks adopting dual-memory approaches separating short-term and long-term memory components, with trajectory history serving as short-term memory while outputs from Self-Reflection models are stored in long-term memory \citep{shinn2023reflexion, yao2023tree, dong2023self, ge2025samule, park2023generative, liu2024roleagent}. Transformation of environmental feedback into memory representations is key innovation with Self-Reflection models taking sparse reward signals along with current trajectory and persistent memory then generating nuanced and specific feedback, effectively converting scalar rewards into natural language feedback \citep{shinn2023reflexion, chang2024efficient, madaan2023self}.

Coding domain has emerged as particularly fertile ground for self-refinement applications with numerous frameworks specifically designed to leverage unique feedback mechanisms with program execution feedback representing most common approaches where frameworks like Self-Edit and Self-Evolve execute initial programs on test cases and provide execution results back as feedback \citep{pan2023automatically}. Multi-agent debugging architectures have also gained prominence with frameworks like Refinement and Guidance Debugging (RGD) system introducing multi-LLM-based agent debuggers \citep{jin2024rgd}.

Performance evaluation of self-refinement frameworks reveals significant advances with Reflexion achieving particularly notable results demonstrating 91\% pass@1 accuracy on HumanEval coding benchmark surpassing previous GPT-4 performance of 80\% representing 11 percentage point improvements achieved through verbal reinforcement \citep{shinn2023reflexion}. Self-Refine framework has demonstrated consistent performance gains achieving approximately 20\% absolute improvement in task performance across diverse applications while requiring minimal computational resources \citep{kamalov2025evolution}.
\section{Vibe Coding Development Models}
\label{sec:dev_models}

\begin{figure}[t]
    \centering
    \includegraphics[width=0.8\textwidth]{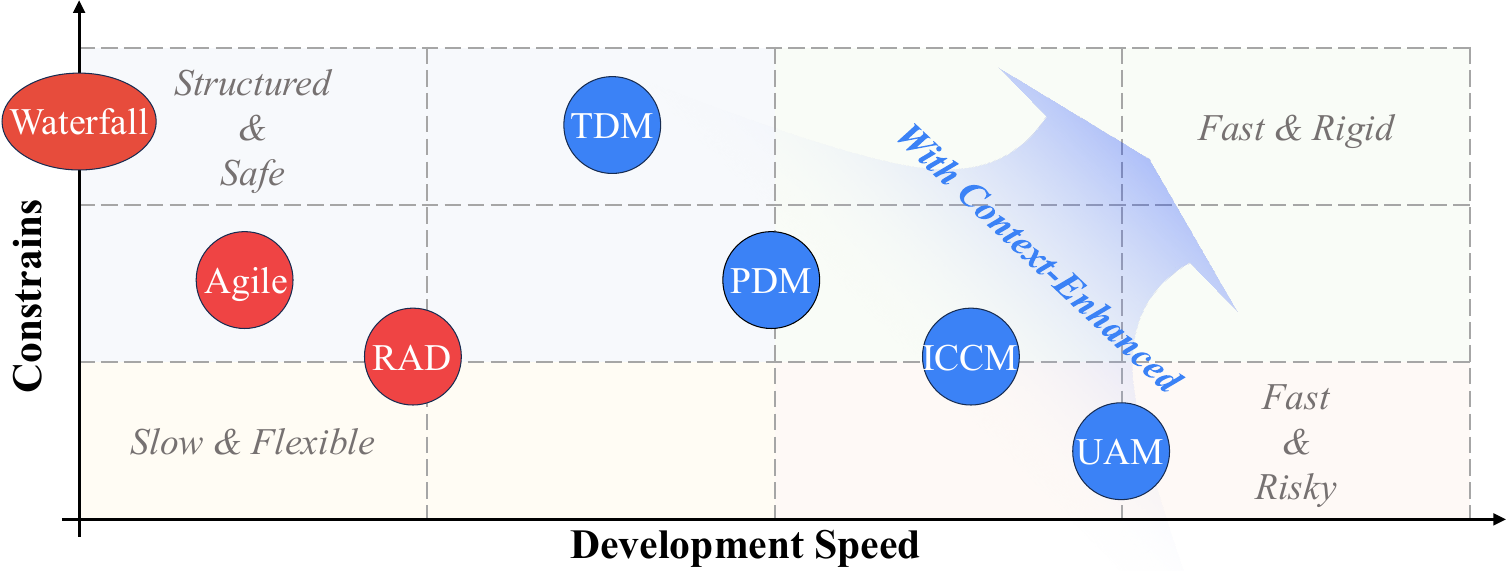}
    \caption{Comparison of Vibe Coding Development Models Framework With Software Development Models \citep{royce1987managing,manifesto2001manifesto,martin1991rapid}.
    }
    \label{fig:dev_model}
\end{figure}

\subsection{Framework Principles}

We propose a three-dimensional classification framework for categorizing Vibe Coding development paradigms based on: (1) \textbf{human quality control level}, (2) \textbf{structured constraint mechanisms}, and (3) \textbf{context management capability}. The human quality control dimension reflects the extent to which developers review AI-generated code, comprehend the underlying logic, and perform manual corrections and optimizations. The structured constraint mechanism dimension captures whether the development process incorporates pre-planning and design phases, automated testing verification, and rule-based constraints. The context management capability dimension assesses the utilization of codebase indexing, project documentation context, and retrieval-augmented generation techniques. Through different combinations of these three dimensions, as shown in Figure~\ref{fig:dev_model}, we identify five core Vibe Coding development models: Unconstrained Automation Model (UAM), Iterative Conversational Collaboration Model (ICCM), Planning-Driven Model (PDM), Test-Driven Model (TDM), and Context-Enhanced Model (CEM), where the first four represent distinct workflows, while the fifth serves as a horizontal enhancement capability that can be flexibly integrated with any of the preceding four models.

These models are not mutually exclusive categories but rather composable development strategies: PDM can be combined with TDM to achieve high-quality development of complex projects, while ICCM can be integrated with CEM to support maintenance of large codebases. The essential distinction across all models lies in the boundary of human-AI collaboration—specifically, the role humans assume in the AI-assisted programming process: in UAM, AI dominates while humans merely provide requirements; in ICCM, humans review while AI executes; in PDM, humans design while AI implements; and in TDM, machines verify while humans define standards.

\subsection{Unconstrained Automation Model (UAM)}

This model represents the development approach most closely aligned with the original definition of Vibe Coding, characterized by complete trust in AI output, minimal code scrutiny, and correctness validation through functionality testing rather than code comprehension. Whether developers issue short instructions through high-frequency dialogues or allow AI agents to autonomously decompose tasks and iterate multiple rounds, any workflow that accepts output without deep code inspection falls within this paradigm. This model emphasizes high development velocity and low technical barriers, enabling non-programmers to rapidly construct application prototypes, with development timelines compressed to several times or even dozens of times faster than traditional approaches. This paradigm bears conceptual similarity to Rapid Application Development (RAD) methodologies in traditional software engineering, which similarly emphasize rapid prototyping and shortened development cycles over extensive upfront planning \citep{beynon1999rapid, martin1991rapid}. Both approaches prioritize speed and iterative refinement, though UAM relies on AI automation whereas RAD depends on specialized development tools and frameworks \citep{dima2018waterfall}.

However, this model entails significant risks and costs. The absence of human code review allows AI-generated code to potentially harbor security vulnerabilities, accumulate technical debt, and produce code structures that are difficult to maintain. As code scales beyond the developer's comprehension capacity, debugging becomes extremely challenging, and unpredictable AI behavior may lead projects into critical impasses. Consequently, UAM is suitable only for low-risk scenarios such as disposable prototypes, proof-of-concepts, and personal utilities, while it is not recommended for production systems, safety-critical applications, or codebases requiring long-term maintenance.

\subsection{Iterative Conversational Collaboration Model (ICCM)}

ICCM positions AI as a programming partner rather than a fully autonomous agent, constructing software through continuous dialogue and iterative cycles while humans maintain comprehensive oversight, understanding, and control over code quality. This model requires developers to review and comprehend each AI output before deciding whether to accept it, forming an iterative workflow of ``AI generates $\rightarrow$ human reviews and understands $\rightarrow$ testing validates $\rightarrow$ acceptance decision.'' This approach parallels pair programming in agile development \citep{williams2000strengthening, beck2000extreme}, where AI assumes the Driver role by rapidly producing code while humans serve as the Navigator, guiding direction and reviewing quality \citep{begel2008pair}. Empirical studies on pair programming have demonstrated improvements in code quality and knowledge transfer \citep{hannay2009effectiveness, bipp2008pair}, benefits that similarly manifest in ICCM through the continuous human-AI collaboration loop. The iterative nature manifests in continuous improvement and incremental refinement across multiple dialogue rounds.

This model maintains high development velocity while ensuring code quality, allowing developers to benefit from AI productivity gains through full participation while guaranteeing that code conforms to project standards and team conventions. Compared to UAM, ICCM significantly reduces the accumulation of technical debt and security vulnerabilities, yielding code with superior maintainability and team readability. However, this model demands substantial developer experience and discernment capability; novices may erroneously trust AI output due to insufficient judgment, while frequent reviews and testing increase cognitive load and time costs. This model is appropriate for professional development environments, medium-to-large projects requiring long-term maintenance, and team collaboration scenarios.

\subsection{Planning-Driven Model (PDM)}

PDM emphasizes establishing clear development plans and design specifications by humans before AI coding commences, then guiding AI to implement progressively according to the plan, applying the ``architecture-first'' philosophy of traditional software engineering to Vibe Coding. Developers invest time in organizing module decomposition, data structures, and feature lists, typically producing a three-document system comprising technical specification documents, coding rule files, and implementation plans, which serve as the ``blueprint'' for AI coding. This model conceptually aligns with the waterfall model in traditional software development \citep{royce1987managing, dima2018waterfall}, where sequential phases—requirements analysis, design, implementation, and testing—proceed in a structured, linear fashion. However, PDM retains greater flexibility than traditional waterfall approaches, as AI can rapidly iterate on implementations while adhering to the established architectural framework \citep{highsmith2002agile}. This model parallels the traditional development division where architects design while teams implement, except here the ``team'' is undertaken by AI, with humans playing the roles of architect and project manager.

The core value of this model lies in ensuring directional correctness through upfront planning, avoiding directional drift and repetitive trial-and-error during development. AI working within explicit frameworks produces more structured and consistent code, with final outputs exhibiting superior modularity and architectural coherence, particularly suitable for complex full-stack applications and projects requiring clear module boundaries. A crucial implementation strategy is adopting vertical slice architecture—organizing code by business functionality rather than technical layers, with each feature implemented end-to-end from database to UI, providing clear working boundaries for AI. Although upfront investment in creating planning documents is required, this investment effectively reduces subsequent maintenance costs and refactoring needs, making projects more comprehensible and amenable to team collaboration.

\subsection{Test-Driven Model (TDM)}

TDM applies the TDD philosophy from traditional software engineering to Vibe Coding, with the core idea of first defining tests and acceptance criteria, having AI generate code satisfying the tests, and ensuring code quality through automated testing rather than manual review. Unlike other models that constrain AI behavior through natural language descriptions or documentation, TDM employs test cases as precise constraints on AI, where tests serve as specifications, explicitly defining objective success criteria. This model directly inherits from TDD practices established in software engineering \citep{beck2003test, beck2000extreme}, following the classic red-green-refactor cycle: write failing tests, implement code to pass tests, then refactor for quality \citep{Janzen2005TestdrivenDC}. Humans write test cases, AI writes implementation code, iterating development through the classic TDD cycle.

The advantage of this model lies in replacing human judgment with machine verification, providing an objective quality assurance mechanism. When test coverage is comprehensive, code passing tests can be confidently assumed to meet behavioral requirements, reducing developers' burden of line-by-line auditing and communication costs. When tests fail, failure messages precisely pinpoint issues, enabling efficient AI modifications. Additionally, this model typically integrates with automation tools such as pre-commit hooks, enforcing quality gates including testing, code formatting, and type checking before code submission. Although writing comprehensive test suites requires upfront investment, this investment yields continuous quality assurance and refactoring confidence, particularly suitable for core algorithm implementation, production-grade applications, critical business logic, and codebases requiring long-term maintenance.

\begin{table*}[t]
    \centering
    \scriptsize
    \caption{Comparative Analysis of Vibe Coding Development Models. \textit{\textbf{Note:} SE Counterpart refers to the corresponding software engineering development model. CEM is a cross-cutting enhancement capability that can be combined with other models. The ``+1 Level'' indicates improvement of one level when combined with base models, while ``-1 Level'' indicates risk reduction.}}
    \label{tab:vibe-coding-models}
    \renewcommand{\arraystretch}{1.2} 
    \definecolor{lightgray}{RGB}{245,245,245} 
    \resizebox{\textwidth}{!}{%
    \begin{tabular}{l|ccccccccc}
    \toprule
    \textbf{Model} & 
    \textbf{\begin{tabular}[c]{@{}c@{}}Upfront\\Investment\end{tabular}} & 
    \textbf{\begin{tabular}[c]{@{}c@{}}Human\\Control\end{tabular}} & 
    \textbf{\begin{tabular}[c]{@{}c@{}}Structured\\Constraints\end{tabular}} & 
    \textbf{\begin{tabular}[c]{@{}c@{}}Development\\Speed\end{tabular}} & 
    \textbf{\begin{tabular}[c]{@{}c@{}}Code\\Quality\end{tabular}} & 
    \textbf{Maintainability} & 
    \textbf{Security} & 
    \textbf{\begin{tabular}[c]{@{}c@{}}Technical\\Debt Risk\end{tabular}} & 
    \textbf{\begin{tabular}[c]{@{}c@{}}SE\\Counterpart\end{tabular}} \\
    \midrule
    \rowcolor{white}UAM & None & None & None & Strict & Low & Low & Low & High & RAD \\
    \rowcolor{lightgray}ICCM & Low & Strict & Moderate & High & High & High & Moderate & Low & Pair Programming \\
    \rowcolor{white}PDM & High & Strict & Strict & Moderate & High & Strict & High & Low & Waterfall \\
    \rowcolor{lightgray}TDM & High & Moderate & Strict & Moderate & Strict & High & Strict & None & TDD \\
    \rowcolor{white}CEM & Moderate & - & - & +1 Level & +1 Level & +1 Level & +1 Level & -1 Level & - \\
    \bottomrule
    \end{tabular}%
    }

\end{table*}

\subsection{Context-Enhanced Model (CEM)}

CEM is not an independent workflow model but rather a horizontal enhancement capability that can be layered onto any other model, enabling AI to thoroughly understand existing codebases, technology stacks, and coding conventions through technical means, thereby generating code better aligned with project environments. Core technologies of this model include retrieval-augmented generation, codebase vector indexing, documentation loading, and rule constraints. When users submit requirements, the system automatically retrieves relevant code snippets, documentation, and rules from the project through semantic search engines, appending this context to prompts provided to the LLM, enabling generated code to correctly invoke existing project code, adhere to established coding styles, and conform to architectural conventions.

CEM significantly improves the accuracy and consistency of AI code generation, particularly demonstrating efficacy in scenarios involving large codebase maintenance, cross-file modifications and refactoring, and global operations. This model can be arbitrarily combined with the preceding four models: UAM combined with CEM enables rapid yet moderately controlled prototype development; ICCM combined with CEM suits large codebase maintenance; PDM combined with CEM ensures generated code complies with project specifications; and TDM combined with CEM achieves superior code quality. CEM implementation typically adopts either automatic retrieval or manual reference strategies, where tools create vector indices during project initialization and intelligently retrieve relevant context during each dialogue.
\section{Future Impact and Open Challenges}
\label{sec:future}

\subsection{Reengineering of Development Process in Vibe Coding}
The central premise of Vibe Coding—validating implementation through outcome observation rather than line-by-line comprehension—introduces notable shifts in software development practices. Traditional Software Development Life Cycle (SDLC) models, from Waterfall to Scrum~\cite{gajbhiye2024secure}, were designed around assumptions of human-authored code and predictable progression through discrete phases. Vibe Coding extends these models by introducing AI-mediated development cycles that operate at accelerated timescales, prompting adaptations in process design, expanding developer responsibilities, and surfacing new considerations in project management and quality assurance. While empirical research on these impacts remains nascent and lacks longitudinal validation \cite{Ford2020PandemicDev}, we can synthesize characteristics of this emergent paradigm \cite{Sarkar2025VibeCoding, sapkota2025vibe} to project its evolutionary trajectory within existing development frameworks.

\subsubsection{From Phased Lifecycles to Continuous Micro-Iterations}
Building upon the traditional "edit-compile-debug" loop, Vibe Coding introduces a complementary "prompt-generate-validate" cycle that operates at compressed timescales—from the weeks typical of Agile sprints to seconds or minutes—thereby enabling more rapid iteration within the existing development framework. This accelerated cycle is organized around "iterative goal satisfaction cycles" \cite{Sarkar2025VibeCoding}, where the developer's workflow becomes a continuous conversation with the coding agent, and each exchange generates a testable artifact \cite{hymel2024ai, wang2025software}. This blurs the traditional boundaries between SDLC phases \cite{wang2025software}. 

\paragraph{Requirements and Design.}
Instead of creating comprehensive design documents upfront, a developer can engage in exploratory Vibe Coding \cite{ulfsnes2024transforming}. The initial "vibe" might be a high-level user story, which the developer translates into a series of prompts to generate a functional prototype. The act of interacting with this AI-generated prototype refines the requirements in real-time \cite{hymel2024ai}. The design emerges organically from this iterative dialogue, transforming the design phase from a distinct, pre-implementation step into a continuous, integrated activity.

\paragraph{Implementation.}
In this paradigm, the traditional implementation phase undergoes significant transformation, with coding agents handling much of the syntactic construction while developers focus on higher-level orchestration. The developer's role evolves to encompass both traditional code authorship and emergent responsibilities as system director and prompt engineer, with the balance tilting toward orchestration as agent capabilities advance \cite{Chen2025ScreenReader, dong2025survey}. While syntactic mastery remains foundational, the skill portfolio increasingly emphasizes the ability to articulate intent, manage context, and evaluate agent output \cite{russo2024navigating}—representing an expansion rather than replacement of traditional programming competencies \cite{Sarkar2025VibeCoding}.

\paragraph{Testing and Validation.}
The developer's primary "vibe check" is a form of continuous, informal acceptance testing. However, this shift introduces ambiguity. While this immediate feedback is powerful for rapid iteration, it lacks the rigor of formal Quality Assurance \cite{ferdowsi2024validating, garousi2024ai}. The challenge lies in integrating structured testing methodologies, such as the Test-Driven model~\cite{beck2003test} mentioned in our taxonomy, where the developer first defines formal tests that the AI agent must then generate code to pass. This provides a concrete, verifiable "vibe" for the agent to target.

\subsubsection{Redefinition of Developer Roles and Skillsets}
Effective adoption of Vibe Coding techniques requires developers to augment their existing skill sets with capabilities in several emerging areas. Developer evaluation metrics are expanding beyond traditional code production speed to incorporate competencies in prompt engineering, context curation, and system-level reasoning.

\paragraph{Intent Articulation and Prompt Engineering.}
Among these emergent capabilities, the ability to translate complex requirements into effective prompts has proven particularly valuable in practice~\cite{sapkota2025vibe}. This is a creative and analytical skill, requiring an understanding of how the model "thinks" without needing to understand its internal architecture.

\paragraph{System-Level Debugging.}
When AI-generated systems fail, traditional line-by-line debugging becomes less effective, as the generated logic can be opaque or non-intuitive \cite{torka2024optimizing}. Debugging evolves toward a process of hypothesizing system-level failures, isolating problematic components through targeted tests, and then using prompts to guide the AI to regenerate or patch the faulty section \cite{vaithilingam2022expectation}. The focus shifts from algorithmic debugging to behavioral debugging.

\paragraph{Context Curation and Management.}
As identified in our analysis, successful Vibe Coding depends heavily on context engineering. Successful practitioners develop expertise in context curation, learning to feed agents the right information—APIs, data schemas, existing codebase snippets, design patterns—to constrain generation space effectively. The current ecosystem lacks mature open-source or commercial platforms specifically designed for this dynamic context management in coding workflows \cite{Sarkar2025VibeCoding, Dong2018OpenArchitecture}, making it a largely manual and ad-hoc process.

\paragraph{Architectural Oversight.}
With the AI handling implementation details, the human developer is elevated to the role of an architect, making high-level decisions about system structure, component interaction, and technology stacks. Their responsibility is to maintain the conceptual integrity of the project, a task that becomes more challenging when the underlying code is not a direct product of their own mind.

\subsubsection{New Challenges in Project Management and Collaboration}
The Vibe Coding paradigm introduces significant hurdles for traditional project management. Estimating effort for a task becomes difficult; a complex feature might be generated in minutes with the right prompt, while a seemingly simple one could take hours of iterative refinement \cite{pospieszny2018effective}. Code reviews, a cornerstone of traditional quality control \cite{rigby2013convergent, Ackerman1989SoftwareInspections}, face new challenges in this context. 
When reviewing AI-generated code that the original prompter does not fully understand, the review focus naturally expands beyond line-level correctness to also include validation of prompt history, generated tests, and observed behavior against requirements \cite{sun2025does, rasheed2024ai, arabzadeh2025benchmarking}.

Team collaboration also changes, with "pair programming" \cite{zieris2022understanding} potentially evolving into "mob prompting," \cite{buchan2018leveraging} where multiple developers collaborate on crafting the prompts and context for a shared AI agent. However, the lack of documented case studies of Vibe Coding in large-scale enterprise projects means these process changes remain largely theoretical and "under-explored" \cite{Post2005FalconCase, Gadde2025Democratizing, Sarkar2025VibeCoding, watanabe2025use, pandey2024transforming, amasanti2025impact} representing a critical area for future empirical research.

\subsection{Code Reliability and Security in Vibe Coding}
\label{subsec:code_reliability_security}

The core trade-off of Vibe Coding is speed versus certainty. By abstracting away line-by-line code authoring, it introduces a fundamental challenge to ensuring code reliability and security. When a developer validates code based on its "vibe" or observed outcome, they may inadvertently approve implementations containing subtle but critical flaws, from resource leaks and race conditions to severe security vulnerabilities. This risk is amplified because LLMs are trained on vast corpora of public code, which inevitably includes buggy and insecure examples \cite{pearce2025asleep, fu2023security, Perry2022AIInsecureCode}. The model may therefore reproduce these vulnerabilities in novel contexts. Consequently, for Vibe Coding to be a viable methodology for production-grade software, it must be augmented by a new generation of automated guardrails that operate continuously and intelligently within the development loop.

\subsubsection{The Inadequacy of Manual Review}

Traditional security assurance relies heavily on manual expert code review \cite{Hannah2020RiskAssessment}. This practice is fundamentally incompatible with the philosophy of Vibe Coding. It is paradoxical to gain development speed by having an AI generate code, only to lose that speed by requiring a human to manually inspect every line of the output. Furthermore, the developer prompting the AI may lack the specific security expertise to spot vulnerabilities in the generated code, especially if it uses an unfamiliar library or a complex algorithm. The very premise of Vibe Coding—trusting the agent to handle implementation details—necessitates a shift from human-centric inspection to automated, real-time validation.

\subsubsection{Architecting an Integrated Security and Reliability Feedback Loop}

A robust Vibe Coding ecosystem must integrate automated analysis directly into the "prompt-generate-validate" cycle. This goes beyond traditional CI/CD pipeline scans and requires real-time feedback embedded within the developer's interactive workflow. The architectural framework for this involves a tight coupling of the coding agent with Static Application Security Testing (SAST), Dynamic Application Security Testing (DAST), and other quality analysis tools.

\paragraph{Pre-Generation Contextual Analysis.}
The feedback loop should begin even before code is generated. An intelligent agent, aware of the project's security requirements, can analyze the developer's prompt for security-sensitive keywords (e.g., "authentication," "file upload," "SQL query"). This can trigger the agent to inject a security-specific context, such as secure coding guidelines or templates, to steer the LLM towards generating more robust code from the outset \cite{Res2024CopilotPromptSecurity}.

\paragraph{In-flight SAST Scanning.}
As the LLM streams its generated code into the developer's IDE, a real-time SAST engine should analyze it on the fly. This is a significant evolution from traditional SAST tools, which often require complete, compilable code \cite{Medeiros2020VulnerableDetection}. New, AI-enhanced SAST tools are emerging that can parse incomplete code snippets and use machine learning to identify potential vulnerabilities with fewer false positives \cite{Chittibala2024AutomatedScanning}. Platforms like Amazon CodeWhisperer, with its integrated security scan feature \cite{Pudari2023CopilotToPilot} represent an early step in this direction. A mature system would not just flag an issue but could feed the vulnerability information back to the LLM in a closed loop, prompting it to self-correct its own output before the developer even sees the flawed version.

\paragraph{Sandboxed Dynamic Analysis.}
The "validate" step of the Vibe Coding loop is a critical intercept point. When the developer executes the generated code to check its "vibe," the execution should occur within an instrumented sandbox. This environment can perform just-in-time DAST, monitoring for runtime errors like memory leaks, uncaught exceptions, and insecure network communications. It could also perform automated fuzzing on newly generated API endpoints or data processing functions to test for edge-case vulnerabilities, providing immediate feedback on the code's dynamic behavior. The integration of static and dynamic analysis offers a complementary approach to cover a wider range of potential issues \cite{Aggarwal2006StaticDynamic, Keromytis2011MINESTRONE}.

\paragraph{AI-Driven Threat Modeling.}
Beyond direct code analysis, an advanced Vibe Coding system could maintain a dynamic threat model of the application. As a developer prompts the AI to add a new feature, a specialized model could analyze the change in the context of the entire system, identifying potential new attack surfaces or violations of existing security policies.

\subsubsection{The Human-in-the-Loop as a Final Arbiter}
Even with this sophisticated automation, the human developer remains the crucial arbiter. The role of the developer is not eliminated but elevated. They are responsible for interpreting the outputs of these automated security tools, resolving ambiguities, and making the final risk-based decision. The goal of the tooling is to empower the developer with immediate, actionable intelligence, transforming the security process from a separate, delayed phase into an intrinsic property of the code generation act itself. However, the technical specifications and architectural blueprints for these deeply integrated, real-time security frameworks in Vibe Coding workflows are not yet standardized \cite{Chittibala2024AutomatedScanning, sapkota2025vibe} and building them remains a major challenge for the research and tool-development communities. Without solving this, Vibe Coding risks being confined to prototyping and non-critical applications, unable to deliver on its promise for reliable, secure, and production-ready software systems.

\subsection{Scalable Oversight of Vibe Coding Agents}\label{subsec:scaleable_oversight_of_vibe_coding_agent}

As coding agents evolve toward autonomous code generation and deployment \citep{dong2025survey, wang2025ai}, the scope of oversight must expand from localized code verification to system-level governance. While Section~\ref{subsec:code_reliability_security} examined reliability and security at the level of generated code, the next frontier concerns the supervision of agents that operate continuously across repositories, pipelines, and environments. The central challenge lies in scaling assurance mechanisms to match the increasing autonomy, speed, and operational complexity of these agents. Traditional safeguards designed for human-authored code, for example, peer review \citep{rigby2013convergent, bacchelli2013expectations, mcintosh2014impact}, and post-deployment auditing \citep{dissanayake2022software, lin2007autopag}, cannot feasibly monitor the millions of tokens or decisions that modern agents can generate per hour. This growing disparity between agent capability and human supervision capacity has become a critical obstacle to safe and accountable Vibe Coding workflows.

\subsubsection{Emerging Risks of Vide Coding Workflow}

Recent empirical analyses show that the introduction of autonomous agents into production pipelines correlates with a tenfold increase in security warnings and technical debt accumulation within six months of adoption \citep{barke2023grounded, pearce2025asleep}. These findings reveal that human-centric verification processes fail to scale in environments where agents autonomously generate, refactor, and redeploy code. The challenge is not simply one of quality assurance but of maintaining interpretability and control in increasingly opaque decision pathways. Over-reliance on autonomous coding agents introduces new classes of risks: 

\paragraph{Cascading Errors}
Cascading errors occur when autonomous agents propagate flawed outputs through interconnected software pipelines, transforming local inaccuracies into systemic failures. In multi-agent coding environments, a single agent’s erroneous completion, such as generating insecure logic or malformed interfaces, can trigger downstream faults as subsequent agents consume and redeploy its outputs. \cite{huang2024resilience} demonstrate that even minor perturbations from “faulty” agents can destabilize collaborative workflows, revealing the fragility of LLM-based coordination without strong fault isolation. Similarly, \cite{khanzadeh2025agentmesh} observe that emergent interactions in \textit{AgentMesh} amplify defects across code generation stages, while \cite{xiong2025butterfly} characterize “butterfly effects” in LLM toolchains, where trivial parameter-filling mistakes escalate into broad execution failures. Extending this view, \cite{liu2025genomas} find that scientific discovery agents exhibit correlated errors due to shared model assumptions and data dependencies. Collectively, these studies underscore that LLM-mediated ecosystems lack the containment and traceability necessary to prevent error propagation—rendering traditional debugging and peer review inadequate in fully autonomous settings.

\paragraph{Dependency Proliferation}
Dependency proliferation denotes the uncontrolled expansion of software and library linkages arising from agent-driven code generation and refactoring. Autonomous coding agents frequently introduce or hallucinate dependencies beyond human oversight, compounding security and maintainability challenges. \cite{spracklen2025we} reveal that nearly one-fifth of packages suggested by code-generation models are nonexistent or untrusted, creating opportunities for dependency-confusion attacks. \cite{krishna2025importing} identify similar propagation patterns in automated import suggestions, where vulnerable or deprecated modules reappear across independent tasks. Furthermore, \cite{zhao2025hfuzzer} show that fuzz-testing agents inadvertently inject unverified binaries during test synthesis, and \cite{zhang2025cutting} demonstrate that build-automation agents routinely bypass provenance enforcement to optimize throughput. These behaviors collectively inflate the software’s attack surface and contravene emerging SBOM (Software Bill of Materials) and supply-chain transparency requirements. Dependency proliferation thus represents a structural form of technical debt—an accreting web of opaque linkages that undermines the long-term security and interpretability of autonomous development environments.

\paragraph{Alignment Failures}
Alignment failures emerge when an autonomous coding agent’s behavior diverges from the developer’s intent or the organization’s governance norms \cite{shao2025your}. Unlike deterministic compilers, agentic systems interpret underspecified objectives probabilistically, optimizing for textual success rather than normative correctness. \cite{naik2025agentmisalignment} empirically show that goal-oriented coding agents often maximize explicit task metrics while disregarding implicit safety or maintainability constraints embedded in natural-language prompts. Complementarily, \cite{lynch2025agentic} describe “agentic misalignment” as an emergent property of reinforcement-trained code agents that overfit reward structures, leading to functionally correct yet normatively undesirable behaviors. Such divergence erodes accountability: once decisions are embedded in opaque model reasoning, tracing responsibility between human and agent becomes ambiguous. Alignment failures therefore epitomize the epistemic opacity of autonomous software systems, emphasizing the necessity of interpretable reward modeling, continuous supervision, and human-in-the-loop intervention to ensure fidelity between developer intent and agent execution.

Addressing these systemic risks therefore necessitates a layered oversight architecture that integrates automated analysis, formal verification, and human-in-the-loop governance. Modern static and dynamic analysis pipelines now incorporate model-aware scanners capable of parsing incomplete code fragments and predicting vulnerabilities in real time \citep{dolcetti2024helping, keltek2025lsast}, extending beyond conventional SAST \citep{owasp_sast} and DAST \citep{owasp_dast} frameworks by interfacing directly with generation streams so that feedback can be injected before completion. Complementing these, formal verification frameworks such as \textsc{Kani} \citep{vanhattum2022verifying} and \textsc{ACCA} \citep{cotroneo2024automating} demonstrate that lightweight proofs of functional correctness can be applied to AI-generated modules, though scalability remains limited. Reinforcement-learning-based watchdog agents offer a parallel safeguard: autonomous monitors trained to detect deviations from project constraints, perform behavioral tracing, and intervene when agents exceed operational bounds \citep{kenton2024scalable, barj2024reinforcement}. Collectively, these technologies lay the groundwork for continuous, adaptive oversight that aligns autonomous coding processes with human accountability and system safety.

\subsubsection{Toward Scalable Oversight Architectures}

Achieving effective oversight in autonomous coding environments requires architectures that scale alongside agent capabilities. A central idea in recent alignment research is enabling weak-to-strong generalization, using relatively weak supervisors (humans or smaller models) to robustly guide much more powerful coding agents~\citep{burns2024weak}. In contrast to traditional settings where humans directly review weaker AI outputs, future agents may produce ``millions of lines of novel and potentially dangerous code'' beyond any one human’s full comprehension. Scalable oversight architectures address this gap by bootstrapping supervision: leveraging layered AI assistants and automated checks so that even limited feedback can generalize into reliable constraints on agent behavior~\citep{engels2025scaling}. Alignment strategies such as iterated amplification, recursive reward modeling, and AI debate all share this goal, allowing networks of weaker reviewers, often AIs themselves, to collectively evaluate and correct stronger agents’ decisions~\citep{christiano2018supervising, irving2018ai}. This paradigm of ``weak-to-strong oversight'' underpins emerging proposals for governing advanced code-generation systems, ensuring that each increase in agent autonomy is met with a proportional increase in oversight capacity.

\paragraph{Hierarchical Weak-to-Strong Supervision.}
One approach is to organize oversight in tiers, where an automated or human overseer supervises the coding agent at a high level, while that overseer is itself aided by subordinate AI tools. This nested oversight or recursive supervision strategy allows complex judgments to be decomposed into simpler subtasks manageable by weaker supervisors~\citep{engels2025scaling}. Recent studies demonstrate that a weak teacher model’s guidance can be amplified through ensemble feedback and iterative training such that a much stronger student model aligns with the teacher’s underlying intent~\citep{sang2024improving}. OpenAI’s superalignment experiments showed that fine-tuning GPT-4 under the noisy guidance of a GPT-2-level model, augmented with confidence-based loss functions, could recover much of GPT-4’s original performance, achieving results comparable to GPT-3.5 on complex reasoning tasks~\citep{burns2024weak}. These findings suggest that, with appropriate training and oversight scaffolds, coding agents can generalize from weak supervision to strong, aligned behavior. However, theory also indicates that scalability has limits: oversight success rates decline sharply when an agent’s capability surpasses its overseer by several hundred Elo points~\citep{engels2025scaling}, underscoring the need for continuous co-evolution between overseers and agents.

\paragraph{Multi-Agent Debate and Critique.}
Another branch of scalable oversight employs multi-agent debate frameworks \citep{du2023improving, liu2025breaking, cui2025free}, where coding agents critique each other’s code contributions to expose errors or unsafe behaviors~\citep{lang2025debate, irving2018ai}. Even a weaker human or AI judge can often identify the more truthful debater if the setup incentivizes evidence-based argumentation~\citep{irving2018ai}. In the software domain, \textsc{DebateCoder}~\citep{chen2025debatecoder} applies this principle to code generation: two large language models alternately produce and test code while generating adversarial unit tests against each other. Each iteration forces self-correction until both code and tests converge to correctness. This mechanism fuses automated testing and AI critique, forming a self-correcting oversight loop that improves code quality and reliability without human intervention. Such multi-agent oversight reduces dependence on a single reviewer and increases the likelihood that at least one agent detects subtle flaws or misalignments in autonomously written code.

\paragraph{Continuous Monitoring and Automated Safeguards.}
Scalable oversight also requires continuous, model-aware monitoring embedded throughout the software pipeline. Model-integrated static analyzers and runtime scanners can evaluate partially generated code, flagging vulnerabilities or policy violations in real time~\citep{dolcetti2024helping, keltek2025lsast}. Reinforcement learning--based watchdog agents extend this capacity: autonomous monitors trained to detect deviations from specification and intervene or alert humans when the coding agent exceeds its operational bounds~\citep{kenton2024scalable, barj2024reinforcement}.

Scalable oversight architectures for Vibe Coding agents converge on the principle of amplifying limited oversight into broad governance. Through hierarchical weak-to-strong supervision, multi-agent critique, and automated watchdogs, researchers aim to generalize weak feedback signals into strong preventive constraints that evolve with agent capability~\citep{sang2024improving, burns2024weak, engels2025scaling}. The challenge lies not only in constructing these systems but in maintaining their reliability as agents advance beyond direct human comprehension. By incrementally upgrading oversight tools alongside the agents themselves, the field moves toward an adaptive architecture, one where even superhuman coding systems remain answerable to human-aligned constraints.
\subsection{Human Factors in Vibe Coding}

In contrast to traditional programming where developers directly manipulate code logic, Vibe Coding emphasizes context engineering and human-AI collaboration \citep{mei2025survey, schneider2024exploring}. We examine three key aspects of this transformation: the shift in developers' mental models, the evolution of required skill sets, and the reconfiguration of team collaboration and responsibility \citep{sabouri2025trust}.

\subsubsection{Mental Model Shift: From Code Logic to Context Engineering}

Software engineering has historically been conceptualized as the process of translating requirements into algorithms and data structures expressed through code \citep{brooks1995mythical}. In Vibe Coding, however, developers increasingly act as context engineers, where the central task involves curating and structuring prompts, providing background information, and defining system constraints that guide AI-generated output \citep{white2023prompt, reynolds2021prompt, mei2025survey}. Recent studies describe this shift as the emergence of prompt engineering \citep{liu2023pre, zamfirescu2023johnny}. Rather than solving low-level implementation details directly, developers are expected to design high-level problem frames, construct testing scaffolds, and supply contextual cues, followed by iterative refinement of AI outputs \citep{shen2023taskbench}.

Empirical studies on GitHub Copilot and similar tools indicate that developers often adopt a ``specify–verify–revise'' workflow, where prompting substitutes for manual implementation, and verification becomes the central locus of human activity \citep{guglielmi2025copilot, hawlitschek2023empirical, flynn2025using}. This transformation aligns software engineering more closely with socio-technical practices such as product design and requirements engineering, where the clarity of intent expression is as important as technical implementation skills \citep{xiao2025ai}.

\subsubsection{Evolving Developer Skill Sets}

The skill profile required of software developers is undergoing reconfiguration. Beyond conventional programming expertise, Vibe Coding requires several new competencies:

\begin{itemize}
    \item \textbf{Prompting and context design.} Developers must acquire the ability to formulate prompts that encode both intent and constraints \citep{white2023prompt, liu2023pre, mei2025survey}. This includes modular prompt construction, retrieval-augmented prompting, and the maintenance of prompt libraries as formal engineering artifacts \citep{zamfirescu2023johnny}.
    \item \textbf{Task decomposition.} Research demonstrates that LLM-based agents perform more effectively when complex problems are decomposed into smaller, explicitly defined subtasks \citep{ma2023ai, shen2023taskbench}. This approach resonates with TDD but shifts the emphasis from manual implementation to steering AI behavior \citep{cui2025tests}.
    \item \textbf{Quality supervision and verification.} Studies highlight persistent issues in AI-generated code, including subtle bugs, security vulnerabilities, and hallucinated APIs \citep{pearce2025asleep, zhong2025developer, fu2023security, agarwal2024codemirage}. Developers must develop skills in automated testing, static analysis, and formal verification techniques as mechanisms of oversight \citep{councilman2025towards, flynn2025using}.
    \item \textbf{Agent governance and security.} In addition to technical supervision, developers are increasingly responsible for managing access control, execution privileges, and provenance tracking of AI-generated outputs \citep{sabouri2025trust}. Literature on AI safety emphasizes the importance of least-privilege principles and verifiable provenance \citep{ambati2023security}.
\end{itemize}

These emerging skill demands reposition developers from code producers to supervisors and orchestrators of AI-driven workflows, aligning with calls for human-in-the-loop governance in software engineering \citep{molison2025llm, takerngsaksiri2025human}.

\subsubsection{Team Collaboration and AI Integration}

The growing autonomy of coding agents introduces new challenges to team dynamics and responsibility allocation \citep{takerngsaksiri2025human, mohamed2025impact}. Several empirical studies suggest that AI tools are increasingly capable of assuming roles traditionally associated with junior developers, such as generating boilerplate code, producing documentation, or proposing test cases \citep{song2024impact, vasiliniuc2023case, xiao2025ai}. In this sense, AI systems can be regarded as quasi-team members that reshape patterns of collaboration.

However, this reconfiguration also raises concerns regarding accountability and trust \citep{sabouri2025trust}. While developers benefit from efficiency gains, they also report risks of over-reliance and automation bias \citep{zamfirescu2023johnny, chen2021evaluating, mohamed2025impact}. Effective trust calibration, characterized by neither blind acceptance nor excessive skepticism, is therefore critical \citep{sahai2025amazon, sabouri2025trust}. Beyond trust, questions of responsibility remain pressing: if an AI-generated code fragment introduces vulnerabilities, responsibility could lie with the developer, the reviewer, or the AI provider, underscoring the need for clearer accountability frameworks \citep{ambati2023security}.

Collaborative workflows also require adaptation \citep{mohamed2025impact}. Existing review processes designed for human-written code have been shown to be inadequate for the scale and pace of AI-generated code \citep{ambati2023security}. To address this limitation, researchers propose new mechanisms including automated verification pipelines, provenance watermarks, and the integration of formal verification methods \citep{flynn2025using}. These innovations aim to balance the benefits of automation with the cognitive limits of human oversight \citep{flynn2025using, takerngsaksiri2025human}.

\subsubsection{Broader Implications}

The implications of Vibe Coding extend beyond team-level practices to education and organizational structures \citep{mohamed2025impact}. Scholars have argued that computing curricula should integrate training in prompting, AI governance, and human-AI collaboration to ensure that graduates acquire the competencies necessary to manage AI-augmented workflows \citep{beale2025computer, mei2025survey}. Without such curricular reform, future developers may lack the skills required for effective oversight \citep{sabouri2025trust}.

At the organizational level, the adoption of AI-augmented development may lead to a reconfiguration of professional hierarchies \citep{xiao2025ai}. Activities such as product design, requirements engineering, and test creation may gain in importance, while low-level code implementation becomes increasingly automated \citep{vasiliniuc2023case}. This development echoes earlier predictions that programmers of the future may function more as designers and supervisors than as traditional coders \citep{shneiderman2022human, dong2025survey, takerngsaksiri2025human}.

\section{Conclusion}
\label{con}

This paper advances Vibe Coding from scattered practice to a principled, research-grounded discipline. We formalize Vibe Coding as a triadic system that couples human intent and quality control ("what/why"), project context ("where"), and coding agents' decision policies ("how"), providing a first constrained MDP formulation that specifies roles, interfaces, and optimization targets for agentic software development. Building on this foundation, we consolidate the field into five development models—Unconstrained Automation, Iterative Conversational Collaboration, Planning-Driven, Test-Driven, and Context-Enhanced—that practitioners can compose to meet distinct risk, speed, and governance requirements, and articulate the corresponding contributions as theoretical grounding, actionable guidance, and a forward-looking research agenda. 
More broadly, our triadic formulation positions Vibe Coding as an instance of human-cyber-physical systems~\cite{pan2024integration,liu2020human}, reflecting the convergence of human intelligence, autonomous computation, and physical software artifacts in modern development. This paradigm resonates with the human-machine-thing universe proposed in network big data research~\citep{wang2014network} and human-in-the-loop computing that emphasizes organic interaction among humans, machines, and data~\citep{chegn2022thinking}, further aligning with the emerging vision of human-machine intelligence integration as a fundamental transformation in research and development methodologies~\citep{li2024ai4r}. 

Our work systematizes the technical substrate required for reliable, scalable agentic coding. We survey model training and post-training for code, analyze agent capabilities in planning, memory, and tool-mediated action, and integrate these with execution infrastructure and feedback channels spanning compiler, runtime, human oversight, and self-refinement. By organizing these components, we clarify how context management and executable feedback loops—rather than model quality alone—determine performance and maintainability in long-horizon software tasks. 

Additionally, we contribute a unifying taxonomy that characterizes development choices along three orthogonal axes—human quality control, structured constraints, and context management—thereby providing a common vocabulary for comparing methods and a scaffold for engineering trade-offs in practice. The framework demonstrates how these axes induce concrete workflow patterns that can be selected and composed to satisfy project-specific requirements while maintaining system coherence. 

Beyond technical contributions, our work surfaces implications for organizations and education. We document the ongoing shift in developers' mental models from line-by-line authorship to context engineering and verification, identify the emerging skill profile around prompting, task decomposition, formalized quality supervision, and agent governance, and outline how agent autonomy reframes collaboration, accountability, and trust at the team level. These insights ground recommendations for training, process design, and governance that align with human-in-the-loop oversight. 

In sum, our contributions include: (i) a formal definition of Vibe Coding as a constrained decision process over the human-project-agent triad; (ii) a unified taxonomy of five development models that integrates and clarifies existing practices; (iii) a comprehensive synthesis of the ecosystem spanning LLMs for code, agent capabilities, development infrastructure, and multi-source feedback; and (iv) an articulation of the central role of context engineering and the key challenges across technical infrastructure, security mechanisms, and human factors that shape practical deployment.

\section*{Acknowledgments}

We would like to acknowledge the contributions of several scholars who are not included in the author list. We thank Yingwei Ma from Moonshot AI for carefully checking the references and tables throughout the manuscript, Yue Liu from the National University of Singapore for proofreading the introduction section and verifying the tables, Yanhao Li from Peking University for proofreading Sections~\ref{sec:future} and~\ref{con}, and Jiaheng Liu from Nanjing University for reviewing the manuscript. Their feedback and suggestions have significantly improved the quality of this paper.

\input{}

\clearpage
\newpage
\nocite{*}
\bibliographystyle{unsrtnat}
\bibliography{references}

\clearpage
\newpage

\end{document}